\documentclass[sigconf, nonacm]{acmart}

\newcommand\vldbdoi{XX.XX/XXX.XX}

\newcommand\vldbavailabilityurl{URL_TO_YOUR_ARTIFACTS}
\usepackage{tikz}
\usepackage{amsmath}
\usepackage{bm}
\usepackage{enumitem}
\usepackage{multirow}
\usepackage{amsthm}
\usepackage[noend]{algorithmic}
\usepackage[noend,lined,boxed,vlined,ruled]{algorithm2e}
\usepackage{graphicx}
\usepackage{etoolbox}
\usepackage{float}
\usepackage{caption, subcaption}
\usepackage{array}
\usepackage{makecell}
\usepackage{tabularx}
\usepackage[normalem]{ulem}
\usepackage[capitalise]{cleveref}
\usepackage{listings}
\SetKwInOut{Input}{input}
\SetKwInOut{Output}{output}

\newcommand{\sys}{\textsc{Table-Specialist}\xspace}

\newcommand{\stitle}[1]{\vspace{1ex}\noindent{\bf #1}}

\newtheorem{df}{Definition}
\newtheorem{ex}{Example}
\newtheorem{pr}{Proposition}
\newtheorem{lemm}{Lemma}
\newtheorem{lem}{Theorem}

\newenvironment{example}{\begin{ex} \nopagebreak
\begin{rm}}{{\hfill$\Box$}\end{rm}\end{ex}} 

\newenvironment{definition}{\begin{df} \nopagebreak
\begin{rm}}{{\hfill$\Box$}\end{rm}\end{df}} 

\newenvironment{proposition}{\begin{pr} \nopagebreak
\begin{rm}}{{\hfill$\Box$}\end{rm}\end{pr}}

\usepackage{fontawesome}

\newcommand{\codeq}[1]{{\tt {\small ``#1''}}}

\newcommand{\code}[1]{\texttt{\small #1}}

\newtoggle{full}
\toggletrue{full}
\begin{document}
\pagestyle{plain} % removed header with this

\title{Table-Specialist: Language Model Specialists for Tables using Iterative Generator-Validator Fine-tuning}

\author{Junjie Xing}
\authornote{Work done while at Microsoft.}
\affiliation{%
  \institution{University of Michigan}
}
\email{jjxing@umich.edu}

\author{Yeye He, Mengyu Zhou, Haoyu Dong, Shi Han, Dongmei Zhang, Surajit Chaudhuri}
\affiliation{%
  \institution{Microsoft Corporation}
}
\email{yeyehe@microsoft.com}

% \author{Mengyu Zhou}
% \affiliation{%
%   \institution{Microsoft Research}
% }
% \email{mezho@microsoft.com}

% \author{Haoyu Dong}
% \affiliation{%
%   \institution{Microsoft Research}
% }
% \email{hadong@microsoft.com}

% \author{Shi Han}
% \affiliation{%
%   \institution{Microsoft Research}
% }
% \email{shihan@microsoft.com}

% \author{Dongmei Zhang}
% \affiliation{%
%   \institution{Microsoft Research}
% }
% \email{dongmeiz@microsoft.com}

% \author{Surajit Chaudhuri}
% \affiliation{%
%   \institution{Microsoft Research}
% }
% \email{surajitc@microsoft.com}

\begin{abstract}
% Outline:

% vanilla model (1) perform well on table tasks highly-available on the web (NL-to-SQL) (2) can be supervised fine-tuned with training data.

% However, (1) doesn't work well on less available tasks (2) supervised fine-tuning on highly available tasks faces domain-shift and over-fit problem (3) annotating training data for specific task or use case is expensive.

% Propose \sys, where we use language models like gpt-35-turbo to generate and validate training data for specific table tasks with "text-book" generation instructions, and iteratively fine-tune the model. The experiment result demonstrates: (1) improved performance on various table tasks and datasets (2) comparable performance with GPT-4, but cheaper and faster (3) generalizability, benchmark-agnostic (4) labeling free(no human)

% -------------------------------------------------

Language models such as GPT and Llama have shown remarkable ability on diverse natural language tasks, yet their performance on complex table tasks (e.g., NL-to-Code, data cleaning, etc.) continues to be suboptimal. 
To improve their performance, task-specific fine-tuning is often needed, which, however, require expensive human labeling and is prone to over-fitting.

In this work, we propose \sys, a self-trained fine-tuning paradigm specifically designed for table tasks. Our insight is that for each table task, there often exist two dual versions of the same task, one \emph{generative} and one \emph{classification} in nature. Leveraging their duality, we propose a \emph{Generator-Validator} paradigm to iteratively generate-then-validate training data from language models, to fine-tune stronger \sys models that can specialize in a given task, without using manually-labeled data.

Extensive evaluations of \sys on Llama, GPT-3.5 and GPT-4 suggest that our \sys has (1) \textit{strong performance} on diverse tasks over vanilla language-models -- for example, \sys fine-tuned on  GPT-3.5 not only outperforms vanilla GPT-3.5, but can often surpass GPT-4 level quality, (2) \textit{lower cost} to deploy, because when \sys fine-tuned on GPT-3.5 achieve GPT-4 level quality, it becomes possible to deploy smaller models with lower latency/cost at comparable quality, and (3) \textit{better generalizability} when evaluated across multiple benchmarks, since \sys is fine-tuned on a broad range of training data systematically generated from diverse real tables. 

Our code is available at \href{https://github.com/microsoft/Table-Specialist}{\faGithub~microsoft/Table-Specialist}.
%and a corresponding technical report can be found at \href{https://arxiv.org/abs/2410.12164}{arXiv}.
Specialist models fine-tuned using \sys have been integrated into Microsoft Excel for use cases such as automated data cleaning. 

\end{abstract}

\maketitle

% %%% do not modify the following VLDB block %%
% %%% VLDB block start %%%
% \pagestyle{\vldbpagestyle}
% \begingroup\small\noindent\raggedright\textbf{PVLDB Reference Format:}\\
% \vldbauthors. \vldbtitle. PVLDB, \vldbvolume(\vldbissue): \vldbpages, \vldbyear.\\
% \href{https://doi.org/\vldbdoi}{doi:\vldbdoi}
% \endgroup
% \begingroup
% \renewcommand\thefootnote{}\footnote{\noindent
% This work is licensed under the Creative Commons BY-NC-ND 4.0 International License. Visit \url{https://creativecommons.org/licenses/by-nc-nd/4.0/} to view a copy of this license. For any use beyond those covered by this license, obtain permission by emailing \href{mailto:info@vldb.org}{info@vldb.org}. Copyright is held by the owner/author(s). Publication rights licensed to the VLDB Endowment. \\
% \raggedright Proceedings of the VLDB Endowment, Vol. \vldbvolume, No. \vldbissue\ %
% ISSN 2150-8097. \\
% \href{https://doi.org/\vldbdoi}{doi:\vldbdoi} \\
% }\addtocounter{footnote}{-1}\endgroup
% %%% VLDB block end %%%

% %%% do not modify the following VLDB block %%
% %%% VLDB block start %%%
% \ifdefempty{\vldbavailabilityurl}{}{
% \vspace{.3cm}
% \begingroup\small\noindent\raggedright\textbf{PVLDB Artifact Availability:}\\
% The source code, data, and/or other artifacts have been made available at \url{\vldbavailabilityurl}.
% \endgroup
% }
% %%% VLDB block end %%%

\section{Introduction}
\label{sec:intro}

Recent language models, such as GPT~\cite{llm-gpt-3}, Llama~\cite{llm-llama} and Mistral~\cite{llm-mistral},  have shown remarkable abilities to perform diverse natural language tasks~\cite{eval-glue, eval-superglue, llm-gpt-3, llm-llama}. Such models show strong generalizability, in that they can be prompted 
%no longer need dataset-specific or task-specific training, but can be used directly, with 
with few-shot examples, to perform across a wide range of language tasks~\cite{llm-gpt-3, kojima2022large}.

However, when it comes to complex ``table tasks'', such as data transformation~\cite{data-transform-tde, data-transform-flashfill, auto-transform, auto-pipeline},  data cleaning~\cite{data-cleaning-survey, error-detection-unidetect, error-detection-holodetect}, 
%and NL-to-Code~\cite{nl2sql-spider, nl2sql-wikisql}, 
where the central object of interest is a \emph{structured relational table} (as opposed to natural language text), even the latest language models can struggle to perform well, despite prompt-level optimizations~\cite{table-gpt, table-llama, spreadsheetllm, mmtu}. This is likely because language models are trained predominately on one-dimensional natural language text, whereas tables are two-dimensional in nature, which are fundamentally different for models to understand and manipulate~\cite{table-gpt, sui2024table, spreadsheetllm}.

\begin{figure*}[t]
    \vspace{-15mm}
    \centering
    \includegraphics[width=0.65\linewidth]{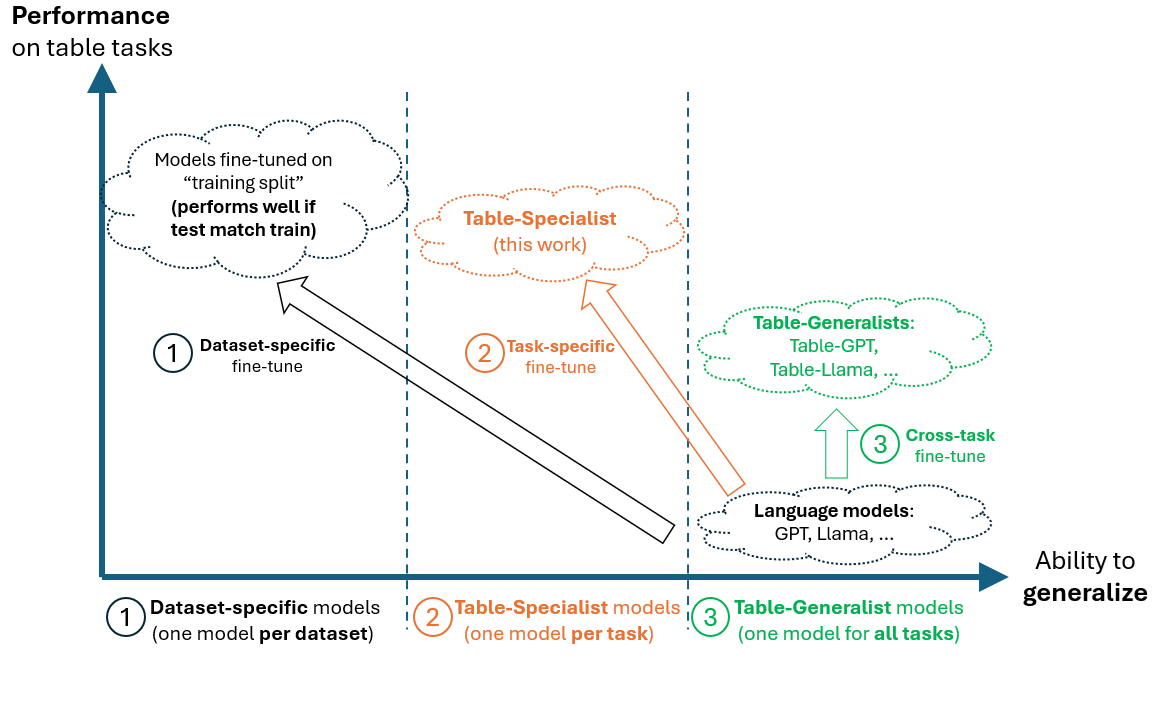}
    
    \vspace{-10mm}
    \caption{Performance vs. generalizability trade-offs: A visual comparison of different fine-tuning approaches for table-tasks. \underline{(1) Dataset-specific fine-tuning}: models are fine-tuned on benchmark ``training split'' of one dataset, which performs well on the corresponding ``test split'' (but may not generalize to a different datasets for the same task type). \underline{(2) Table-Specialist fine-tuning} (this work): we propose to fine-tune one model per table-task (e.g., data cleaning, data transformation, etc.), which generalizes well across datasets for the same task type.   \underline{(3) Table-Generalist fine-tuning}: methods that fine-tune one general-purpose model to handle many different table-tasks, which has good generalizability,  at the cost of lower-performance on individual tasks.}
    \vspace{-2mm}
    \label{fig:finetuning-spectrum}
\end{figure*}

\begin{figure}
    \centering
    \begin{subfigure}{.49\linewidth}
      \centering
      \includegraphics[width=\linewidth]{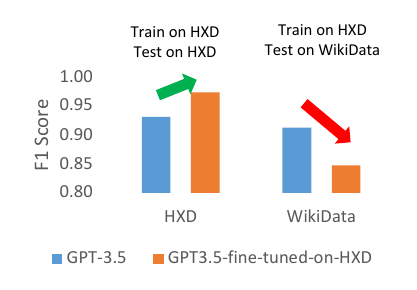}
      \caption{Schema Matching}
    \end{subfigure}    
    \unskip\ \vrule\
    \begin{subfigure}{.49\linewidth}
      \centering
      \includegraphics[width=\linewidth]{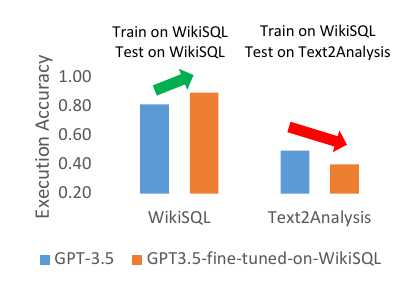}
      \caption{NL-to-SQL}
    \end{subfigure}
        \vspace{-5mm}
    \caption{``Dataset-specific fine-tuning'' using GPT-3.5 for  table-task $T$: (a) Schema matching, (b) NL-to-SQL. In both cases, while GPT-3.5 fine-tuned using the training-split of one dataset $D$ lead to performance gains on the test-split of the same $D$ (shown as green arrows pointing up), they also result in  significant performance loss on another dataset $D'$ for the same task type $T$ (red arrows down), \emph{relative to un-tuned vanilla GPT-3.5}, suggesting likely over-fitting on $D$. %, and hurt models ability to generalize to other datasets $D'$ for the same task $T$.  % GPT-3.5 fine-tuned on HXD leads to better performance on the test-split of HXD~\cite{zhang2024smutf}, but lower performance on wikiData~\cite{Schema matching-valentine}.  can Performance of vanilla GPT-3.5 v.s. Fine-Tuned-GPT35 with Dataset-specific Fine-Tuning for NL-to-SQL (\textit{left}) and schema matching (\textit{right}).
    }
    \vspace{-4mm}
    \label{fig:overfit}
\end{figure}

\textbf{Prior approaches: Fine-tuning for table tasks.} To overcome improve language-models performance on table-tasks, various fine-tuning approaches can be used, which we will review below, and show a visual comparison in Figure~\ref{fig:finetuning-spectrum}.

\underline{Dataset-specific fine-tuning}. A common approach to fine-tuning, shown on the left of Figure~\ref{fig:finetuning-spectrum} (marked as \tikz[baseline=(char.base)]{
    \node[shape=circle,draw,inner sep=1pt] (char) {1};
}), is what we refer to as ``dataset-specific fine-tuning''. Here, given a particular table-task $T$ (say data-transformation), we start from a vanilla ``base'' language-model like GPT or Llama (at the lower-right of the figure), and use a labeled dataset $D$ for task $T$ that is usually divided into train/test splits. The training split is then used to fine-tune language-models, which can often lead to significant performance gains when evaluated using the (highly correlated) test-split.

This is a common practice widely used in many of today's benchmarks for table-tasks, which often come with train/test splits~\cite{nl2sql-spider, nl2sql-wikisql, em-data-magellan, em-data-deepmatcher, zhang2024smutf, table-qa-wikitablequestions, table-qa-tabfact}, where the expectation is for models to be trained and tested on homogeneous splits of the \emph{same labeled dataset} $D$.

%which is then used to train or fine-tune underlying models that can often lead to substantial performance improvements for that specific dataset. For example, many benchmarks today have train/test splits, that are labeled by the same human workers on the same underlying data, where it is common practice to train/fine-tune on the training-split, and then report test performance on the  

However, despite their large capacity, large language models fine-tuned on the training-split of one dataset $D$ often do not generalize well to another dataset $D'$ for the same task type $T$. For instance, Figure~\ref{fig:overfit} shows the result of fine-tuning GPT-3.5-turbo on two table tasks, Schema matching (left) and NL-to-SQL (right). In both cases, we observe that fine-tuned GPT-3.5 using one benchmark dataset $D$ can lead to performance gains on the test-split of the same $D$ (as expected), but significantly lower performance on another benchmark dataset $D'$ of the same task, \emph{compared to the baseline of vanilla GPT-3.5 without fine-tuning}, for both Schema matching and NL-to-SQL.\footnote{Note that ``over-fitting'' is a well-known topic studied in the literature~\cite{overfit-1, overfit-2}, especially for small models. Our analysis serves to show that, for the common table-tasks that we care about, over-fitting can still happen even when we fine-tune large language models such as the  175B GPT-3.5  (which have a large capacity and are supposed to be more robust to over-fitting).}
\iftoggle{full}
{

    Specifically, in Figure~\ref{fig:overfit}(a), we take two separate Schema-matching benchmarks from two sources, HXD~\cite{zhang2024smutf} and Wikidata~\cite{schema-matching-valentine}. We then fine-tune GPT-3.5 using HXD, and test the test-splits of HXD and Wikidata using the resulting fine-tuned models. While substantial gain is observed when the model is trained and tested on HXD (the left half of the Figure~\ref{fig:overfit}(a)), we also note a significant drop in quality, when training is on HXD and test is on Wikidata (the right half of the Figure~\ref{fig:overfit}(a)). Similarly, in Figure~\ref{fig:overfit}(b), when we fine-tune GPT-3.5 using one NL-2-SQL dataset WikiSQL~\cite{nl2sql-wikisql}, and then test also on the test-split of WikiSQL we observe a quality improvement (left of the figure), but when this is tested  on another dataset Text2Analysis~\cite{DBLP:conf/aaai/HeZXMDDGJCHY024} we see another drop in quality (right of the figure).
}
{
    (More details of this analysis can be found in~\cite{full}).
}

We argue that the relatively narrow nature of one specific dataset, often manually labeled at a small scale, can lead to poor ``generalizability'' of fine-tuned language models with ``dataset-specific fine-tuning'', which we illustrate along the x-axis of Figure~\ref{fig:finetuning-spectrum}.

Given that it is hard to anticipate real test data from real users (e.g., in scenarios like data cleaning or NL-to-SQL, where user tables and queries are often not known a priori), we argue that ``dataset specific fine-tuning'' does not provide a robust way to develop models to handle diverse user requests reliably in practice.

\underline{Table-Generalist fine-tuning}. A second class of recent fine-tuning techniques, which we refer to as ``Table-Generalist''.  This class of approaches are inspired by the success of general-purpose chat models like ChatGPT and Llama-Chat, which are fine-tuned from their respective base models (GPT and Llama)~\cite{instruct-gpt, self-instruct, chatgpt}, and show great generalizability by handling diverse human instructions unseen during training, which we will refer to as ``Chat-Generalist''.

Inspired by the success of ``Chat-Generalist'' models like ChatGPT, ``Table-Generalist'' models  such as Table-GPT~\cite{table-gpt} and Table-Llama~\cite{table-llama}
are developed, which are fine-tuned similarly to ChatGPT by pooling diverse table-tasks as training data for multi-task table fine-tuning. The resulting ``Table-Generalist'' models are shown to handle diverse table-tasks, with better performance on a wide range of table tasks than the vanilla language models (GPT and Llama), including on \emph{new and unseen table-tasks} held out during fine-tuning~\cite{table-gpt, table-llama}.

This class of techniques is depicted on the right of Figure~\ref{fig:finetuning-spectrum}  (marked as \tikz[baseline=(char.base)]{
    \node[shape=circle,draw,inner sep=1pt] (char) {3};
}), which shows both strong generalizability (x-axis), and improved table-task performance compared to vanilla language-models (y-axis). However, their cross-task generality comes at a cost of performance, as there is often a performance gap between ``dataset-specific fine-tuning'' and Table-Generalists, like shown along the y-axis.

\textbf{\sys: a new approach to table fine-tuning.} In this work, we develop a new fine-tuning approach for table-tasks that aims to close the performance gap, which we call ``\sys'', shown in the middle of Figure~\ref{fig:finetuning-spectrum} (marked as \tikz[baseline=(char.base)]{
    \node[shape=circle,draw,inner sep=1pt] (char) {2};
}).

In this approach, each \sys model is fine-tuned by design to \emph{focus on one specific type of table task $T$} (e.g.,  one model for data transformation,  one model for error detection, one model for NL-to-SQL, etc.), which is unlike  Table-Generalist models (Table-GPT and Table-Llama) that can handle all types of table tasks. 

Importantly, we show that by being more specialized on one task $T$,  \sys can be (1) made much more performant than Table-Generalists (depicted on the y-axis), while (2) still being able to generalize to new and unseen datasets of the same task $T$ (in contrast to ``dataset-specific fine-tuning'', where models fine-tuned on $D$ may not generalize to $D'$, like depicted on the x-axis).

%Crucially, unlike ``dataset-specific fine-tuning'' (left of Figure~\ref{fig:finetuning-spectrum}) that uses a narrow set of manually-labeled training set from one dataset (often its training-split), which can perform well on the same dataset (its test-split),  but often cannot generalize to new datasets (e.g., Figure~\ref{fig:overfit}), \sys is designed to generalize across different datasets for the same table-task. 

At a high level, our \sys exploits  \emph{a duality of table tasks}, where a ``\emph{generative table-task}'' has a counter-part that is a ``\emph{classification table-task}'', and vice versa, forming two dual versions of the same task. Correspondingly, we propose a  ``Generator-Validator'' framework that can iteratively fine-tune a generative model and a classification model for the dual versions of the task, using training data  automatically ``generated-then-validated'' by the two models, leveraging unique characteristics of tables (e.g., permutation-invariance and execution-invariance). 

While the notion of dual-learning and dual-tasks is studied for machine-translation tasks (e.g., translating from language A to B, and from B to A)~\cite{dual-learning-mt-1, dual-learning-mt-2, dual-learning-mt-3} and computer-vision tasks (e.g., transferring from style A to B, and from B to A)~\cite{dual-learning-cv-1, dual-learning-cv-2, dual-learning-cv-3}, it is not explored in the context of tables, as we will review in Section~\ref{sec:related}.

We show that our approach utilizes diverse real tables to create rich training examples, making the resulting models much more generalizable than baselines such as ``dataset-specific fine-tuning''.

%is inspired by the success of using data generated by capable language-models, to train state-of-the-art small language models~\cite{synthetic-slm-phi15, synthetic-slm-phi3} and text-embedding models~\cite{synthetic-embedding-1}.  
%fine-tuning . We take a similar approach of using synthetic training data generated by language models to fine-tune specialist models.   

\begin{figure}
%\vspace{-5mm}
    \centering
    \begin{subfigure}{.49\linewidth}
      \centering
      \includegraphics[width=\linewidth]{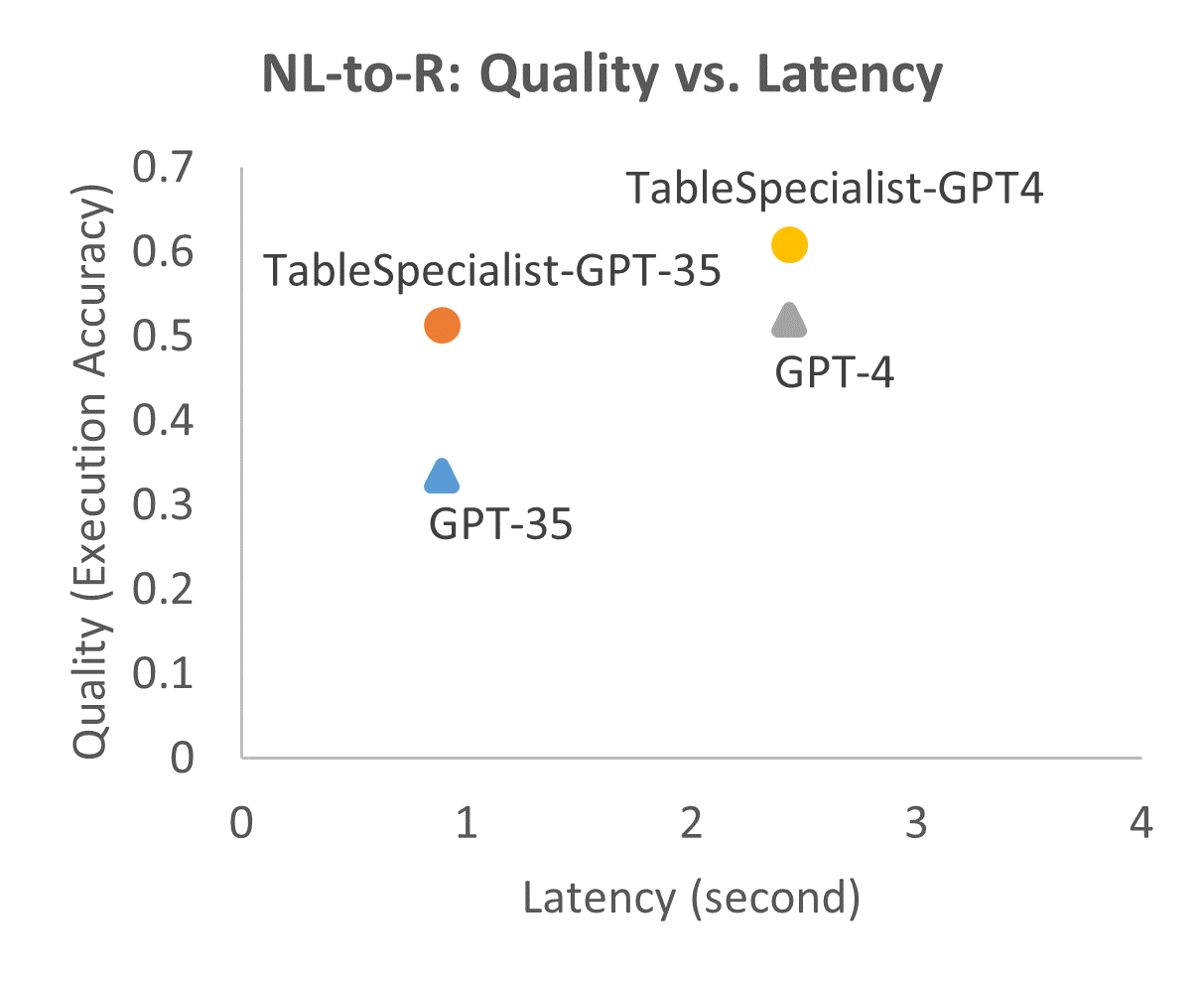}
        \vspace{-5mm}
      \caption{NL-to-R}
    \end{subfigure}    
    \unskip\ \vrule\
    \begin{subfigure}{.49\linewidth}
      \centering
      \includegraphics[width=\linewidth]{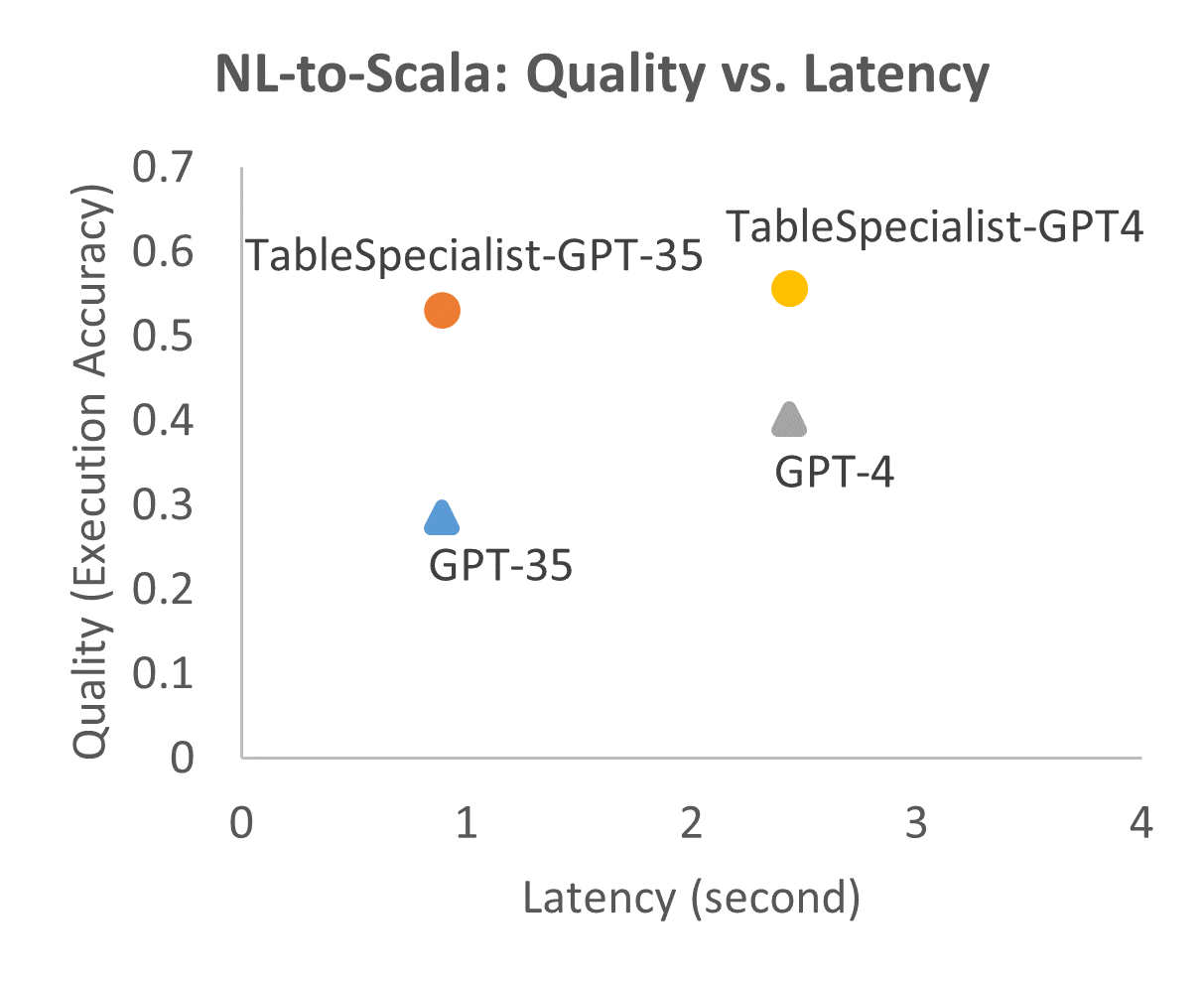}
      \vspace{-5mm}
      \caption{NL-to-Scala}
    \end{subfigure}
        \vspace{-5mm}
    \caption{``\sys fine-tuning'': Quality vs. latency comparison  on two table-tasks: (a) NL-to-R; (b) NL-to-Scala. In both cases, \sys-GPT-3.5 significantly outperforms vanilla GPT-3.5, and can even outperforms vanilla GPT-4 (shown on y-axis), making it possible to deploy \sys-GPT-3.5 over vanilla GPT-4 for these tasks, at substantially lower latency and costs (x-axis).
    }
    \vspace{-4mm}
    \label{fig:quality-vs-latency}
\end{figure}

\textbf{Key benefits of \sys}. 
To better illustrate the benefits of \sys, in Figure~\ref{fig:quality-vs-latency} we highlight our results on two table-tasks, NL-to-R and NL-to-Scala (which are similar to NL-to-SQL but translate natural-language questions to R and Spark-Scala code that can run on tables instead). 

In both figures, we can see that vanilla GPT-4 produces higher quality than vanilla GPT-3.5, but have 2-3x higher latency (and also higher financial cost to deploy), which is expected. The proposed \sys fine-tuned on GPT-3.5 and GPT-4 show strong quality gains over vanilla GPT-3.5 and GPT-4, respectively, \emph{without using any training data from the training-split of the test benchmarks}. 

More importantly, we can see that in both cases, \sys-GPT-3.5 can match or exceed the quality achieved by vanilla GPT-4. Furthermore, because these \sys-GPT-3.5 are fine-tuned on GPT-3.5, they have similar latency as vanilla GPT-3.5 (shown on x-axis). What this means is that for these table-tasks, \emph{we can deploy smaller specialized models (\sys-GPT-3.5) over larger general models (vanilla GPT-4), with comparable quality, but at significantly lower latency and costs.}

We summarize the key benefits of \sys below:
\begin{itemize}[noitemsep,  leftmargin=*] 
  \item \underline{Strong performance}. \sys outperforms vanilla language models such as GPT, as well as generalist models such as Table-GPT and Table-Llama. For example, we show that \sys fine-tuned on GPT-3.5  is not only consistently better than vanilla GPT-3.5, but also often surpasses vanilla GPT-4 for the same table-task, despite being orders of magnitude smaller. 
  \item \underline{Lower cost}.  Because \sys achieves GPT-4 level quality using fine-tuned GPT-3.5 models, it is substantially cheaper to deploy, both in terms of latency and financial costs.  
  \item \underline{Better generalizability}. Unlike ``dataset-specific fine-tuning'', \sys can reliably generalize to new and unseen datasets for the same task. We show in our experiments that \sys is benchmark-agnostic, and show consistent performance gains on multiple benchmarks of the same task, all \emph{without using any data from the training-split of these benchmarks}.
  \item \underline{Labeling-free}. Because the ``Generator-Validator'' framework in \sys leverages language-models to automatically generate-then-validate training data for fine-tuning, it is easier to scale to new table-tasks without expensive human labeling.
\end{itemize}

% Outline:

% LLM, vanilla model with zero-shot, few-shot, prompt-engineering ...

% Table-tuned models TableGPT, TableLLM, TableLlama ...

% Challenge for specific task:

% (1) perform not well on less-seen tasks

% (2) SFT over-fitting problem

% (3) annotation cost high

% unique technical aspects of \sys

% --- Gen-Val

% --- Table-specific shuffling and validation

% --- task generation using ``text-book''

% benefits of \sys

% --- lower cost and latency than larger models (GPT4)

% --- benchmark-agnostic, generalize better and more robust

% --- labeling free

% --- adapt to new tasks

%\input{sections/2.Preliminary}

\section{Related work}
\label{sec:related}

\iftoggle{full}
{
    
    \textbf{Table tasks.} A wide variety of tasks have been studied in the research literature that are centered around tables, which are also increasingly important in practice (e.g., in database and spreadsheet copilots~\cite{excel-copilot, bigquery-copilot, azure-sql-copilot, google-sheet-copilot}). 
    We give a brief overview of tasks considered in this work, and refer readers to surveys like~\cite{table-tasks-1, survey-tables, survey-data-cleaning} for a more comprehensive treatment of the subject.
    
    Table matching tasks, such as entity-matching~\cite{em-deepmatcher, em-book, em-ditto} and schema-matching~\cite{schema-matching-cupid, schema-mapping-survey, schema-matching-valentine}, address the problem of identifying related rows and columns from tables that refer to the same entity, which are usually framed as binary classification problems~\cite{em-ditto, em-deepmatcher}. 
    
    Data cleaning is a broad topic that includes tasks like error-detection~\cite{error-detection-survey, error-detection-holodetect, error-detection-unidetect, auto-detect, auto-test}, that try to identify data errors from tables (which can be seen as a binary classification problem), and error-repair~\cite{survey-data-cleaning, rekatsinas2017holoclean}, that attempt to identify possible fixes for erroneous table cells (which is generative in nature).
    
    In data transformation, problems like by-example program synthesis~\cite{data-transform-tde, data-transform-flashfill} have been  studied, which aim to generate transformation programs based on user-provided input/output examples in a table, where the generated transformations can target different languages (e.g., SQL, Python, etc.)
    
    %Additional tasks such as table question answering~\cite{table-qa-sqa, table-qa-wikitablequestions, table-qa-2, hitab, hybridqa}  (answering user questions based on a table), 
    
    NL-to-Code tasks, which translate user natural-language questions into code into different domain-specific languages that can execute on tables, such as NL-to-SQL~\cite{nl2sql-spider, nl2sql-wikisql, nl2sql-2},  NL-to-Pandas~\cite{DBLP:conf/aaai/HeZXMDDGJCHY024, nl2pandas}, NL-to-PySpark~\cite{nl-2-pyspark-english-query}, etc., are also popular topics of research. 
    %NL-to-spreadsheet-formula~\cite{nl2formula},   
    
    We consider all these common table tasks in our study, in order to evaluate the effectiveness of different fine-tuning approaches. 
    
    \textbf{Language models for table tasks.} 
    %Early versions of encoder-style language models, such as BERT~\cite{llm-bert} and RoBERTa~\cite{llm-roberta}, are already shown to be effective for table-tasks. However, fine-tuning is often required, on the training split of the dataset.
    Auto-regressive language models, such as GPT~\cite{llm-gpt-3} and Llama~\cite{llm-llama}, are capable of performing not only natural-language tasks, but also table-tasks~\cite{llm-can-wrangle-data, llm-vision}. 
    
    However, language models still struggle on complex table tasks~\cite{table-gpt, table-llama, llm-table-understanding-1, llm-table-understanding-2, zhang2023jellyfish}. This can be attributed to factors such as large table context~\cite{llm-table-understanding-1, spreadsheetllm}, two-dimensional reasoning~\cite{table-gpt, llm-table-understanding-1, llm-table-understanding-2}, as well as possible mismatch between pre-train data (one dimensional text) and test-time tasks (two-dimensional tables)~\cite{table-gpt, table-llama}.
    
    To improve language models' performance on table tasks, fine-tuning techniques have been developed, including dataset-specific fine-tuning and table-generalist fine-tuning~\cite{table-gpt, table-llama, generalist-ft-structlm}, as discussed in the introduction. In this work, we introduce a new class of ``specialist fine-tuning'', with better performance than table-generalists, while still being generalizable across datasets of the same task.   
}
{
    \textbf{Language models for table tasks.} A variety of tasks studied in the research literature that are centered around tables, which are also increasingly important in practice (e.g., in database and spreadsheet copilot/assistant scenarios~\cite{excel-copilot, bigquery-copilot, azure-sql-copilot, google-sheet-copilot}). 
    We study a sample of common table-tasks in this work, and refer readers to surveys like~\cite{table-tasks-1, survey-tables, survey-data-cleaning} for a more comprehensive survey of table-tasks.
    
    Auto-regressive language models, such as GPT~\cite{llm-gpt-3} and Llama~\cite{llm-llama}, are capable of performing not only natural-language tasks, but also table-tasks~\cite{llm-can-wrangle-data, llm-vision}. 
    However, language models still struggle on complex table tasks~\cite{table-gpt, table-llama, llm-table-understanding-1, llm-table-understanding-2, zhang2023jellyfish}. This can be attributed to factors such as large table context~\cite{llm-table-understanding-1, spreadsheetllm}, two-dimensional reasoning~\cite{table-gpt, llm-table-understanding-1, llm-table-understanding-2}, as well as possible mismatch between pre-train data (one dimensional text) and test-time tasks (two-dimensional tables)~\cite{table-gpt, table-llama}.
    
    To improve language models' performance on table tasks, fine-tuning techniques have been developed, including dataset-specific fine-tuning and table-generalist fine-tuning~\cite{table-gpt, table-llama, generalist-ft-structlm}, as discussed in the introduction. In this work, we introduce a new class of ``specialist fine-tuning'', with better performance than table-generalists, while still being generalizable across datasets of the same task.   
}

\textbf{Train language models using synthetic data.} In this work, we fine-tune models for individual table tasks, using synthetic training data  ``generated-then-validated'' by language models from diverse real tables, which is inspired by the success of using synthetic data to train state-of-the-art small language models~\cite{synthetic-slm-phi15, synthetic-nemotron, synthetic-orca} and text-embedding models~\cite{synthetic-embedding-1}, that are also trained using synthetic data  generated by language-models. In \sys, we leverage the duality of table tasks, and unique characteristics of tables (e.g., permutation-invariance, and execution-invariance) to validate training data, which are all specific to tables and table tasks.
%\haoyu{does it deserve discussions on works about instruction tuning? It may also achieve similar experiment results to show that fine-tuned GPT3.5 surpasses GPT-4}

%\textbf{Instruction fine-tuning vs. table fine-tuning.}

\textbf{Validation in table-tasks vs.  NLP reasoning tasks.} 
In our Generator-Validator fine-tuning process, we validate table training data based on result consistency (leveraging permutation-invariance and execution-invariance). Our approach can be seen as similar in spirit to consistency-based verification methods used in NLP reasoning tasks, such as ``self-consistency'' and ``tree-of-thoughts''~\cite{self-consistency, tree-of-thoughts, self-verification, shinn2024reflexion}, but is tailored specifically for tables.

\textbf{Teacher-student distillation vs. Self-training.} 
In standard teacher-student distillation, a larger teacher model (e.g., GPT-4) is used to train a smaller student model (e.g., GPT-3.5) for specific tasks~\cite{distillation-1, distillation-2, distillation-3}, where the smaller student model is expected to gain in quality when learning from a more capable teacher model. In our approach, however, both Generator and Validator are symmetric and use the same model (e.g., both are GPT-3.5 to create \sys-GPT-3.5, or both are GPT-4 to create \sys-GPT-4), which is close to a form of ``\emph{self-training}''~\cite{self-train-1, self-train-2, self-train-google-noisy-student}.

\textbf{Dual learning.} Dual learning is a concept in machine-learning where two related tasks are learned together via mutual reinforcement, which are used in machine-translation (where the dual tasks are translating from language A to B, and from B to A)~\cite{dual-learning-mt-1, dual-learning-mt-2, dual-learning-mt-3},  and computer-vision (where the dual tasks can be style-transfers from style A to B, and from B to A)~\cite{dual-learning-cv-1, dual-learning-cv-2, dual-learning-cv-3}. The duality of table tasks we study in this work is similar in spirit, but not explored in the literature to the best of our knowledge.

\begin{figure}[t]        \vspace{-8mm}

    \centering
        \centering
        \includegraphics[width=0.4\textwidth]{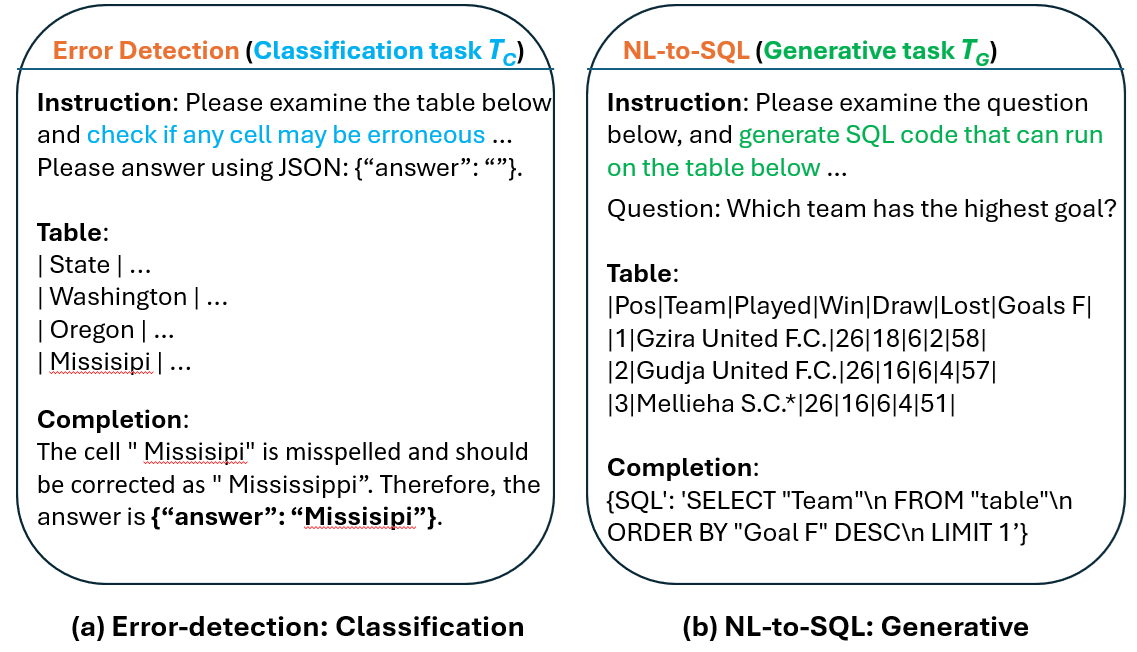}
        \vspace{-3mm}
        \caption{Example table-tasks: Error detection and NL-to-SQL}        \vspace{-5mm}

        \label{fig:example-tasks}
\end{figure}

\begin{table}[]
\caption{List of table-tasks: classification and generative}        \vspace{-3mm}
\begin{small}
\scalebox{0.7}
    {
    \begin{tabular}{|>{\centering\arraybackslash}p{1.4\linewidth}|}
    \hline
    \textbf{Classification table-tasks}% \\  \hline
    \begin{itemize}[leftmargin=0.5cm]
        \item \textbf{Error detection} (multi-class classification): check if any cell in a table column is erroneous
        \item \textbf{Schema matching} (binary classification): check if a pair of columns in two tables are related
        \item \textbf{Entity matching} (binary classification): check if a pair of rows in two tables refer to the same entity
        \item \textbf{Column type annotation} (multi-class classification): determine the type of a column from a list
        \item \textbf{Table fact verification} (binary classification): check if a statement about a table is true or not
        \item[] \centering  \ldots  
    \end{itemize}  \\\hline
    \textbf{Generative table-tasks}% \\  \hline
    \begin{itemize}[leftmargin=0.5cm]
        \item \textbf{NL-to-Code} (SQL, R, Pandas, \ldots): translate natural-language questions to executable code on a table
        \item \textbf{Data transformation by-example} (SQL, R, Pandas, \ldots): generate code for data transformation, based on given input/output examples

        \item \textbf{Table question answering}: answer a natural language question based on a table

        \item \textbf{Data imputation}: fill in missing values in a table, based on  table context
        \item \textbf{Table summarization}: summarize a table using natural language 
        \item[] \centering  \ldots
        % \\ \ldots
    \end{itemize}  \\\hline
    \end{tabular}
    }
\end{small}
\label{tbl:tasks}
  \vspace{-5mm}
\end{table}

\begin{figure*}[t]
\vspace{-18mm}
    \centering
        \centering
        \includegraphics[width=0.9\textwidth]{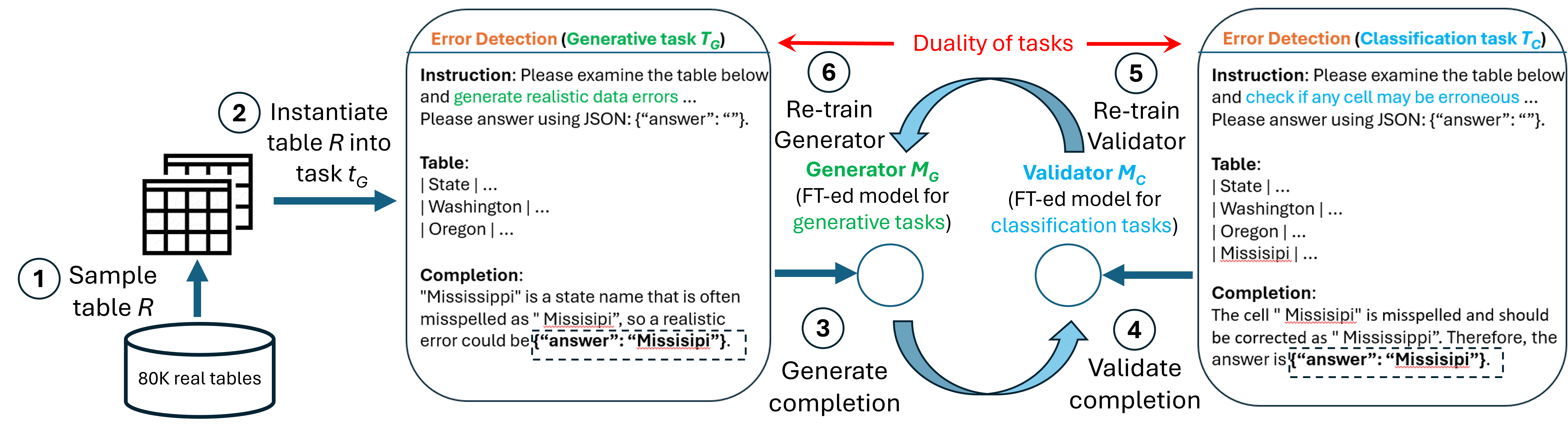}
        \vspace{-2mm}
        \caption{Architecture of \sys  using ``Generator-Validator'' fine-tuning for a given task type $T$ (Error detection in this example). (1) A real table $R$ is sampled from a corpus of diverse tables; (2) Table $R$ is used to instantiate an instance of the generative table task $T_G(R)$ (left box); (3) A ``Generator model'' $M_G$ (initially a vanilla language-model) is used to generate completion for $T_G(R)$, in this case a possible typo error ``Missisipi''; (4) The completion ``Missisipi'' is inserted into $R$, and used to instantiate a classification-version of the Error detection task $T_C$ (right box), which is validated by a ``Validator model'' for the classification task $M_C$  (initially also a vanilla language-model). If $M_C$ consistently produces ``Missisipi'' for $T_C$, then ``Missisipi'' is considered validated (i.e., likely a real error); (5-6) Validated training data is then used to re-train the Generator $M_G$ and Validator $M_C$, for more effective Generator and Validator models. We iteratively fine-tune $M_G$ and $M_C$, by repeating steps (1)-(6) .
    }
        \label{fig:gen-val-arch}
\end{figure*}

\section{Table-Specialist: Overview}
\label{sec:table-specialist-overview}
In this section, we will give a high-level overview of \sys.

\textbf{Preliminary: Generative and classification table-tasks.}
Many table-tasks have been studied in the literature. Table~\ref{tbl:tasks} shows a list of common table-tasks.   

Some of these tasks can be ``\emph{classification}'' in nature, where the output has to come from a predefined set of options. Example  classification table-tasks include Error detection (checking whether any cell in a table may be an error)~\cite{error-detection-survey, error-detection-unidetect, error-detection-holodetect}, Schema matching (checking whether two table columns match)~\cite{schema-mapping-survey, schema-matching-cupid, schema-matching-valentine}, etc.

Note that table tasks may also be ``\emph{generative}'' in nature, where new output needs to be generated. Example here include NL-to-Code (generating code that can execute on a table for a given natural language question)~\cite{nl-2-pyspark-english-query, nl2formula, nl2pandas, nl2sql-wikisql}, where the generated code can be in a target DSL such as SQL, R, Pandas, and Scala, etc.

Following prior work~\cite{table-gpt, table-llama}, from the perspective of using language models to solve table tasks,  we represent each instance of a table task as an ``(\code{instruction, table, completion})'' triple:
\begin{definition}
\label{def:table-task}
    [Table tasks]. An instance of a table task, denoted by $t$, is defined as a triplet $t = (Ins, R, C)$, where $Ins$ is the natural language instruction to describe the task, $R$ is the input table on which the task is performed, $C$ is the expected completion by following the instruction $Ins$ and performing the task on table $R$.
\end{definition}

We give concrete examples of table tasks below.

\begin{example}
\label{ex:table-task}    
    [Table tasks]. Figure~\ref{fig:example-tasks} (a) shows an instance of the Error detection task, which is a classification task that identifies values in a table column that may be erroneous.  Here, we have an \codeq{instruction} that describes the task, an actual \codeq{table} shown in the middle, and a \codeq{completion} that we expect language models to produce, which in this case  identifies \codeq{Missisipi} (a typo) from the table as a predicted error.
    
    Figure~\ref{fig:example-tasks}(b) shows an example generative table task, NL-to-SQL. In this task, SQL code that can execute on tables needs to be generated in \codeq{completion}. 
    Note that this task is generative, as the output is not chosen from a fixed set of options. %Also note that  \codeq{completion} in this case is code snippets, as opposed to natural language text.
\end{example}

\textbf{Goal: Train specialist models that can generalize, using diverse training data generated by language-models.} %  , using data generated by language-models.} 
Recall that a key motivation of this work, based on our observation shown in Figure~\ref{fig:overfit}, is that it is usually hard to fine-tune models on the training-split of a narrow benchmark dataset $D$ (often manually labeled on a small scale), and hope it can still generalize to diverse test cases beyond the test-split of $D$. 

We would like to build ``specialist models'' that specialize in a given type of table task $T$ (say NL-to-Code or Error detection). Crucially, such models need to \emph{generalize to any dataset of the task $T$}, without being over-fit on a particular dataset $D$, so that we can build a model once for each task-type $T$, and then  deploy it in production reliably  (without needing to fine-tune a new model every time a customer brings in a new dataset $D'$).

Given the success of using language-models (e.g., vanilla GPT-3.5 or GPT-4) to generate synthetic training data, for training  small and specialized language models (e.g., code, reasoning, and embedding models)~\cite{synthetic-embedding-1, synthetic-slm-phi15, synthetic-nemotron, synthetic-orca}, we explore a similar direction to train table-specialist models, by using language-models to generate large amounts of synthetic training data (beyond the scale possible with manual labeling), so that we can hope to fine-tune  models for task $T$ that can generalize to diverse test cases in $T$.

\textbf{Challenge: Need to validate training data generated by language-models.} The obvious challenge, however, is that vanilla language-models, denoted by $M$, do not automatically generate high quality synthetic training data for task $T$ -- in fact, if $M$ already understands the task $T$ well enough to consistently  generates accurate  training data, one could argue that there is no need to fine-tune $M$ for  task $T$ in the first place. %(e.g., since for the task of NL-to-SQL).

We observe that for many table-tasks in Table~\ref{tbl:tasks}, even state-of-the-art vanilla language models (e.g., GPT-4) still struggle to produce correct answers. This can be because of (1)  \underline{uncommon DSL}: for some code-related tasks (e.g., NL-to-Code), when the target DSL is relatively uncommon  (e.g., R or Spark-Scala that are sparse in the pre-training data),  language-models may generate incorrect code; (2) \underline{unfamiliar task}: for tasks like {``Data-transformations by-example using code''} in Table~\ref{tbl:tasks}, where the task is to observe input/output examples in a table, and generate transformation code, such tasks are sparse in its pre-training data, and unfamiliar to language-models;   (3) \underline{challenging task}: some tasks are inherently challenging, e.g., Error detection, which has a strong class imbalance where over 98\% of input values can be error-free, and true positives are found in only 2\% of the data, such that achieving both high recall and high precision simultaneously is challenging for any predictive models. % (4) \underline{complex table context}: two dimensional understanding and reasoning is often required for table tasks, e.g., to read tables horizontally and vertically, which are skills that vanilla language-models  still do not fully possess~\cite{table-gpt, llm-table-understanding-1}. 
% can add -- task itself is challenging, e.g., transform using SQL, error-detection which has strong class imbalance, etc.

Note that even when a table-task $T$ may not be particularly challenging for the aforementioned reasons, the training data directly generated by  vanilla language models for $T$ is still often random in nature and far from perfect. 

All of these call for a way to ``validate'' synthetic training data automatically generated by language models, before it can be reliably used  to fine-tune specialist models.

%\textbf{Our approach: Validate training data leveraging ``task duality'', for Generator-Validator fine-tuning.}
\textbf{Our approach: Validate training data using ``task duality''.}
In order to systematically validate training data, we first observe that there is a natural ``\emph{duality}'' in table tasks. Specifically, for each classification table task $T_C$, we can construct a ``dual''  generative task $T_G$, and vice versa, defined as follows:
%that makes it possible to perform validation between $T_C$ and $T_G$.
%We first define duality below.
%We demonstrate this duality using an example below.

\begin{definition}
    \label{def:task-duality}
    [Task duality]. Let $T_G$ be a generative table-task, and $T_G(R)$ be an instance of the task $T_G$ instantiated with  a table $R$. Similarly, let $T_C$ be a classification table-task, and $T_C(f(R))$ be an instance of the task $T_C$, instantiated with  table $f(R)$, where $f$ is a deterministic transformation function applied to $R$. Let $M$ be an oracle model that produces ground-truth completions for any task. The generative task $T_G$ is said to be a  \emph{dual task} of $T_C$, if for any table $R$,  we always have $M(T_G(R)) \equiv M(T_C(f(R)))$, using some fixed transformation $f$.\footnote{Duality in the other direction for generative-tasks can be defined similarly.}
\end{definition}

Intuitively, a task $T_G$ is the dual of $T_C$, if for any table $R$, $M(T_G(R))$ and $M(T_C(f(R)))$ are always expected to produce the same output. For example, we can see that the two tasks  constructed for Error detection in Figure~\ref{fig:gen-val-arch}, as well as the two tasks constructed for Schema matching in Figure~\ref{fig:example-tasks-schema-matching}, are expected to produce the same output, as marked in dashed rectangles, making them ``dual tasks''. 

We  illustrate duality and its construction $f$ in more detail below. 

\begin{example}
\label{ex:duality}
    [Task duality]. The task of Error detection is a multi-class classification task $T_C$, where the goal is to predict if any value in a given table column is an error, like shown in the right box of Figure~\ref{fig:gen-val-arch}.
    We can construct its generative dual, $T_G$, shown in the left box, which  simply asks a model to examine a given table column $R$ and produce a realistic data error.

    To see why $T_C$ and $T_G$ are dual tasks, let $T_G(R)$ be an instance of the generative Error detection task instantiated using a table $R$, like shown in the left-box of the figure. Let $c = M(T_G(R))$ be its completion, in this example $c=$\codeq{Missisipi}, a realistic typo error. Let $f(R) = insert(c, R)$ be a transformation that inserts $c$ into $R$ (creating the column on the right that contains the typo \codeq{Missisipi}). Now for the task $T_C(f(R))$ (identifying errors in $f(R)$), we expect the same $c=$\codeq{Missisipi} to always be returned by an oracle model $M$, ensuring $M(T_G(R)) \equiv M(T_C(f(R)))$, or the two tasks always produce the same output, like shown in the dashed boxes, making the two tasks dual. 

    As another example, we look at Schema matching. Recall that Schema matching is traditionally a classification task $T_C$, that identifies pairs of columns from two input tables as either match or non-match, shown in Figure~\ref{fig:example-tasks-schema-matching}(b). %The generative task shown in dual tasks  shown in Figure~\ref{fig:example-tasks-schema-matching} 
    Like for Error detection, we can construct its generative dual, $T_G$, shown in  Figure~\ref{fig:example-tasks-schema-matching}(a), which takes a Table-A as input, to generate another Table-B, as well as  column mappings between the two tables as output.

    To see why the two are also dual tasks, consider a transformation $f$ that combines Table-A and Table-B from $T_G$ in Figure~\ref{fig:example-tasks-schema-matching}(a), as input for $T_C$ in Figure~\ref{fig:example-tasks-schema-matching}(b), we can see that the two tasks $T_G$ and $T_C$ should now always produce the same output, like indicated by dashed boxes in the figure, because under this transformation $f$, the mappings generated by an oracle model $M$ on $T_G$ should always match that generated by $M$ on  $T_C$.
    %Similarly, for a generative table task, say NL-to-Code, which requires a model to generate a code snippet $c$ to answer a question $q$ on a given table $R$, we can construct its classification counterpart that classifies or validates, whether code $c$ correctly answers question $q$ on table $R$, where the output is expected to be true or false. % \yeye{may point to a figure later} 
\end{example}

\textbf{Generator-Validator fine-tuning.}
Given that two dual tasks are expected to always produce the same output for the same table $R$ (Definition~\ref{def:task-duality}), we leverage this duality to automatically ``generate-then-validate'' training data to fine-tune specialist models.

We give high-level overview of our ``\emph{Generator-Validator}'' fine-tuning framework, illustrated in Figure~\ref{fig:gen-val-arch}.  
Given a target classification task $T_C$ that we want to fine-tune, we first construct its dual generative task $T_G$ (and vice versa), shown as two boxes in the figure.
We then iteratively fine-tune: (1)
a ``\underline{\emph{Generator model}}'', $M_G$, for the generative table-task $T_G$, and (2) a ``\underline{\emph{Validator model}}'', $M_C$, for the classification table-task $T_C$,  in the middle of the figure.  

In Figure~\ref{fig:gen-val-arch}, in each iteration, we would first \tikz[baseline=(char.base)]{
    \node[shape=circle,draw,inner sep=1pt] (char) {1};
} sample a real table $R$ from a large corpus to \tikz[baseline=(char.base)]{
    \node[shape=circle,draw,inner sep=1pt] (char) {2};
} instantiate a task $T_G(R)$, and then \tikz[baseline=(char.base)]{
    \node[shape=circle,draw,inner sep=1pt] (char) {3};
} invoke $M_G$ (initially a vanilla language model) to generate a completion $c=M_G(T_G(R))$, which we know is also the expected completion for the corresponding classification task $T_C$, given the task-duality, which can then be used to ``train'' the classification model $M_C$. However, since such training data are not always correct, we \tikz[baseline=(char.base)]{
    \node[shape=circle,draw,inner sep=1pt] (char) {4};
} invoke $M_C$ (initially also a vanilla language model) to systematically ``validate'' training data. The resulting validated training data that can then be used to \tikz[baseline=(char.base)]{
    \node[shape=circle,draw,inner sep=1pt] (char) {5};
} fine-tune $M_C$ for $T_C$, and
\tikz[baseline=(char.base)]{
    \node[shape=circle,draw,inner sep=1pt] (char) {6};
} fine-tune $M_G$ for $T_G$, to create increasingly more capable $M_G$ and $M_C$ models. %(We will give a more detailed walk-through of these steps in the next section).

The validation step in  \tikz[baseline=(char.base)]{
    \node[shape=circle,draw,inner sep=1pt] (char) {4};
} is key to the success of the iterative fine-tuning, where we leverage unique properties of  tables, such as ``\underline{\emph{permutation-invariance}}'' (reordering rows and columns should not change the semantics of a table), and ``\underline{\emph{execution-invariance}}'' (executing  semantically equivalent code on sub-samples of a table should always produce identical results), etc., to better validate training data. %Note that these are similar in spirit to self-consistency~\cite{self-consistency} and self-critique~\cite{} used in reasoning tasks proposed for NLP tasks, but are tailored to table, leveraging unique characteristics of tables. 

In the end, by fine-tuning using diverse training data generated specifically for  $T_G$ and $T_C$, the resulting models $M_G$ and $M_C$ models become more effective ``specialist models'' in solving $T_G$ and $T_C$ than vanilla language models, which become our \sys models for different table-tasks. %(e.g., Error detection, Schema-matching, NL-to-Code, etc., in Table~\ref{tbl:tasks}).

\underline{Tasks \emph{not} suited for Generator-Validator fine-tuning.}
We want to point out upfront that not all table-tasks are suited for the proposed Generator-Validator fine-tuning. For example, this approach is not directly applicable to tasks that do not have precise ``ground-truth'', such as table summarization~\cite{table-summary-google, table-summary-2, table-summary-3}, as the lack of ground-truth makes it hard for perform validation easily.

There are also tasks that naturally come with ample training data, for which Generator-Validator would not be needed. For example, the task of Data-imputation~\cite{imputation-1, imputation-2} predicts the value for a missing cell in a table, where training data can be easily obtained (by masking out random cells in real tables, and use their ground-truth values for training). For such tasks, it would not be necessary to use of Generator-Validator for fine-tuning.

However, Generator-Validator fine-tuning can apply to tasks with precise ground-truth, and  traditionally require careful manual-labeling to generate ground-truth, such as Error detection, Schema matching, NL-to-Code, and Data-transformations in Table~\ref{tbl:tasks}, which we will evaluate in our experiments.

Next, we will describe in more detail our iterative fine-tuning for classification table-task (Section~\ref{sec:classification-tasks}) and generative table-tasks (Section~\ref{sec:generative-tasks}), respectively. %, using relevant table-tasks as concrete examples to support the discussion. 

\section{\sys: Classification Task}
\label{sec:classification-tasks}
Many table tasks studied in the literature are classification in nature, which can be binary-classifications (e.g., Schema matching, Entity-matching, Table-fact-verification), or multi-class classification (e.g., Error detection, Column-type-annotation, etc.).

Given a target classification table-task $T_C$ that we want to fine-tune, and its dual generative table-task $T_G$ that we can construct, in this section, we describe how the Generator-Validator framework can fine-tune for $T_C$. %to leverage language-models to automatically generate training data and perform fine-tuning. % We use Error detection~\cite{error-detection-survey,error-detection-unidetect,error-detection-holodetect} and Schema-matching~\cite{sm-other-1,schema-mapping-survey,schema-matching-cupid,schema-matching-valentine} as  examples here, but the technique generalizes to other classification tasks as well.  

\renewcommand{\algorithmicrequire}{\textbf{Input:}}
\renewcommand{\algorithmicensure}{\textbf{Output:}}

%\begin{small}
\begin{algorithm}
\scalebox{0.8}
{
    \begin{minipage}{1.2\linewidth}
    \begin{algorithmic}[1]
    \SetKw{kwReturn}{return}
     \REQUIRE{A corpus of real table $\mathcal{R}$, a vanilla language-model $M$, a generative table-task $T_G$, a corresponding classification table-task $T_C$}
     \ENSURE{Fine-tuned specialist model $M_G$ for task $T_G$, and $M_C$ for task $T_C$}
    
    \STATE $M_G \leftarrow M$  \tcp{\footnotesize{initialize the generative model $M_G$ as vanilla $M$}}  \label{ln:init-mg}
    
    \STATE $M_C \leftarrow M$ \tcp{\footnotesize{initialize the classification model $M_C$ as vanilla $M$}}  \label{ln:init-mc}

    \FOR{$i$ in 1 to $k$ iterations}  \label{ln:iter-train}
        \STATE $\text{Train}_G \leftarrow \{\}$ \tcp{\footnotesize{initialize the validated training set for $T_G$}}  \label{ln:init-train-g} 
        \STATE $\text{Train}_C \leftarrow \{\}$  \tcp{\footnotesize{initialize the validated training set for $T_C$}} \label{ln:init-train-c}

        \FOR{$j$ in 1 to step-size}  \label{ln:iter-step}

            \STATE Sample $R \in \mathcal{R}$ \tcp{\footnotesize{sample a real table}} \label{ln:sample}
            
            \STATE Instantiate $t_G \leftarrow T_G(R)$ \tcp{\footnotesize{instantiate a generative task $t_G$ using $R$}}  \label{ln:init-tg}
            
            \STATE $c \leftarrow M_G(t_G)$ \tcp{\footnotesize{invoke $M_G$ to  compute the completion $c$ for $t_G$}} \label{ln:invoke-mg}
            
            \STATE Construct $t_C \leftarrow T_C(R, c)$ \tcp{\scriptsize{construct a classification task $t_C$ with $R$, $c$}}      \label{ln:construct-tc} 
            
            \tcp{\footnotesize{check $c$ is a valid completion of $t_G$, by calling Validate()}}             
            \IF{Validate($M_C$, $t_C$, $c$)}   \label{ln:validate-data}
            
                \STATE $\text{Train}_G \leftarrow \text{Train}_G \cup (t_G, c)$  \tcp{\footnotesize{add the validated $(t_G, c)$ into $\text{Train}_G$}}  \label{ln:add-train-g}
                
                \STATE $\text{Train}_C \leftarrow \text{Train}_C \cup (t_C, c)$ \tcp{\footnotesize{add the validated $(t_C, c)$ into $\text{Train}_C$}} \label{ln:add-train-c}
                
            \ENDIF
            
        \ENDFOR
        
        \STATE Fine-tune $M_G$ using $\text{Train}_G$ \tcp{\footnotesize{fine-tune $M_G$ using validated training data}} \label{ln:ft-mg}
        
        \STATE Fine-tune $M_C$ using $\text{Train}_C$\tcp{\footnotesize{fine-tune $M_C$ using validated training data}} \label{ln:ft-mc}
        
    \ENDFOR

    \kwReturn $M_G, M_C$ \tcp{\footnotesize{return fine-tuned models $M_G, M_C$}}
    \caption{Generator-Validator fine-tuning}
    \label{alg:gen-val}
    \end{algorithmic}
    
    \end{minipage}
}

\end{algorithm}
%\end{small}

Algorithm~\ref{alg:gen-val} shows the general steps of the Generator-Validator approach.  
We start by initializing both the generative model $M_G$ for the generative task $T_G$, and the classification model $M_C$ for the classification task $T_C$, as a vanilla language model $M$ (Line~\ref{ln:init-mg}-\ref{ln:init-mc}). 
We then start our iterative fine-tuning (Line~\ref{ln:iter-train}), by first initializing training set for $T_G$ and $T_C$ as empty sets (Line~\ref{ln:init-train-g}-\ref{ln:init-train-c}). In each fine-tuning iteration, we iteratively perform \textit{step-size} number of sampled steps (Line~\ref{ln:iter-step}), where each time, we sample a real table $R$ from the corpus (Line~\ref{ln:sample}), which we use to instantiate an instance of the generative task $t_G = T_G(R)$  (Line~\ref{ln:init-tg}). We then  invoke $M_G$ on $t_G$, to produce a completion $c$ (Line~\ref{ln:invoke-mg}). We use $c$ and $R$ to construct a corresponding classification task $t_C$ (Line~\ref{ln:construct-tc}). At this point, we perform the crucial validation step by calling the Validate() subroutine (Line~\ref{ln:validate-data}, which calls Algorithm~\ref{alg:validate-for-classification-task} and will be explained next). Once the validation passes, we add $(t_G, c)$ and $(t_C, c)$ as validated training examples for $T_G$ and $T_C$, respectively, because by duality $c$ will be a correct completion for both $t_G$ and $t_C$ (Line~\ref{ln:add-train-g}-\ref{ln:add-train-c}). After performing step-size number of samples, the validated training data will be used to fine-tune $M_G$ and $M_C$ (Line~\ref{ln:ft-mg}-\ref{ln:ft-mc}), to conclude one iteration of the fine-tuning process. We repeat $k$ such iterations, and return the resulting  $M_G$ and $M_C$ as our \sys models.

\renewcommand{\algorithmicrequire}{\textbf{Input:}}
\renewcommand{\algorithmicensure}{\textbf{Output:}}

%\begin{small}
\begin{algorithm}
\scalebox{0.8}
{
    \begin{minipage}{1.2\linewidth}
    \begin{algorithmic}[1]
    \SetKw{kwReturn}{return}
     \REQUIRE{A classification model $M_C$, an instance of classification task $t_C$,  and its expected output $c$}
     \ENSURE{True or False} \tcp{\footnotesize{validate whether $c$ is the correct completion for $t_C$}}

    \STATE $R \leftarrow t_C.R$ \tcp{\footnotesize{get the table $R$ used in task $t_C$}}  \label{ln:assign-R} 

    \FOR{$i$ in 1 to $N$}  \label{ln:iter-validate}
        \STATE $R' \leftarrow $ Permute($R$) \tcp{\footnotesize{permute row and columns of table $R$}}  \label{ln:shuffle} 

        \STATE $t_C' \leftarrow T_C(R')$ \tcp{\footnotesize{instantiate a new $T_C$ task, using the permuted $R'$}}  \label{ln:shuffle} 
        
        \STATE $c' \leftarrow M_C(t_C')$ \tcp{\footnotesize{get
         completion $c'$ for $t_C'$, using classification model $M_C$}} \label{ln:mc-after-shuffle}
        
        \IF{($c' \neq c$)} \label{ln:check-eq}

            \STATE \kwReturn False \tcp{\footnotesize{Not-validated: unsure if $c$ is correct  completion for $t_C$}}
             
        \ENDIF

    \ENDFOR

    \kwReturn True \tcp{\footnotesize{Validated: $c$ is likely the correct completion for $t_C$}}
    \caption{Validate($M_C, t_C, c$): validate for classification tasks}
    \label{alg:validate-for-classification-task}
    \end{algorithmic}
    \end{minipage}
}
\end{algorithm}

Algorithm~\ref{alg:validate-for-classification-task} shows the validation subroutine (Line~\ref{ln:validate-data} of Algorithm~\ref{alg:gen-val}), which is necessary for the following reason. 
Recall that  $c = M_G(t_G)$ is a completion generated by invoking $M_G$ on task $t_G$, which we expect to also be the completion of the corresponding dual task $t_C$ (by task-duality in Definition~\ref{def:task-duality}), such that we can use $(t_C, c)$ as training data to train model $M_C$ for our target classification task $T_C$. 
However, $M_G$ is often not perfect in many table-tasks as we discussed, so that $c = M_G(t_G)$ may not be the correct completion for $t_G$, and thus also not the correct completion for $t_C$, in which case $(t_C, c)$ pairs should not be used for training. We therefore use the subroutine in Algorithm~\ref{alg:validate-for-classification-task} for this validation.
%validates the whether $c$ is the correct output for the classification task $t_C$ (only then can we use $(t_C \rightarrow c)$ as a high-quality training example to fine-tune $M_C$). 

Here, we use a property unique to tables known as ``\emph{permutation-invariance}'' described below, to help validate $(t_C, c)$.

\begin{proposition}
\label{prop:permutation-invariance}
[Permutation-invariance]. Given a task $T$ on a table $R$, let $R'$ be any permuted version of $R$, whose rows and columns may be reordered. \emph{Permutation-invariance} states that because the permuted $R'$ does not change the semantics of the original table $R$, we should always have $T(R) \equiv T(R')$.\footnote{With the exception of tasks that specifically depend on row and column orders, such as ``removing the second row'', which however are uncommon (e.g., not seen in Table~\ref{tbl:tasks}).}
\end{proposition}

This property is intuitive, and is used in  Algorithm~\ref{alg:validate-for-classification-task} as follows. In the pseudo-code, we start by assigning $R$ as the table used in $t_C$. We then iteratively perform  $N$ validations. In each validation iteration, we first permute rows and columns in $R$ to get $R'$, which we use to instantiate a new task $t'_C=T_C(R')$, that is identical to $t_C=T_C(R)$ except that its table $R'$ is a permuted version of $R$ in $t_C$. We then invoke $M_C$ on $t_C'$ to get its completion $c'$. Note that by permutation-invariance, we should have $c = c'$ at this point, but if we verify that $(c' \neq c)$, then we know either $c$ or $c'$ is not a valid completion. Since we are not sure which is valid, we  return \codeq{False} and discard this data point $(t_C, c)$. If $N$ repeated validation iterations all pass, we know $c$ is consistently returned as a completion of $t_C$ despite permutations (not due to flukes or randomness from the model’s non-deterministic behavior), and likely a valid answer. We  return \codeq{True} in such cases, which would make $(t_C, c)$ and $(t_G, c)$ valid training data for $M_C$ and $M_G$, respectively, in Algorithm~\ref{alg:gen-val}. %(treating $c$ as a valid completion of $t_C$ for training), only if $N$ repeated validations on permuted $R'$ all produce the expected completion $c$. % (if $c$ is the correct completion for $t_C$, then we expect $t_C'$ to produce the same completion $c$, given that $t_C'$ and $t_C$ only differ in how the rows/columns of the tables are ordered, and tables are supposed to be permutation-invariant).

\begin{figure}[t]
    \vspace{-6mm}
    \centering
        \centering
        \includegraphics[width=0.4\textwidth]{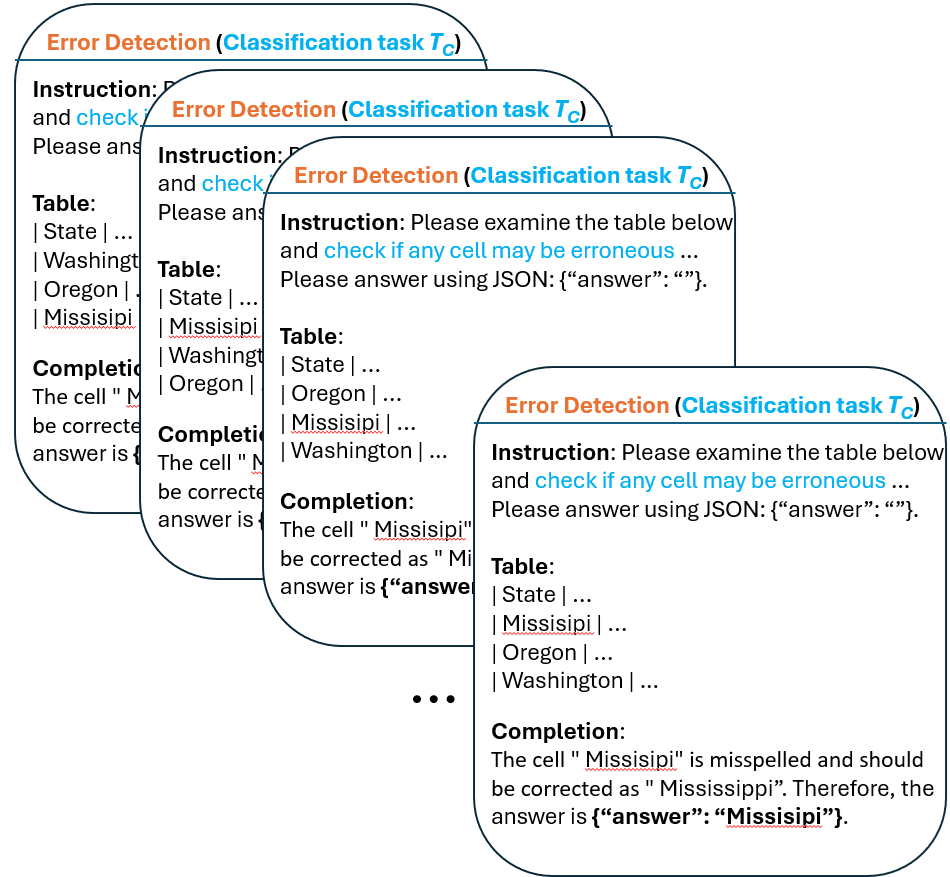}
            \vspace{-3mm}
        \caption{Validate training data by permutation invariance: we permute rows and columns in $R$, and repeatedly invoke $M_C$ to check whether a consistent completion is produced.}
        \label{fig:model-validation-error}
    \vspace{-4mm}
\end{figure}

Connecting Algorithm~\ref{alg:gen-val} and  Algorithm~\ref{alg:validate-for-classification-task} together, we use the following example to illustrate the fine-tuning process end-to-end.

\begin{example}
\label{ex:error-detect}
[Error detection]. We revisit Figure~\ref{fig:gen-val-arch}, and explain the  Error detection task end-to-end.

First, both the Generator and Validator models, $M_G$ and $M_C$, are initialized as a vanilla language model $M$. In each fine-tuning iteration, we sample a real table $R$ and instantiate a generative task $t_G = T_G(R)$, by adding table $R$ into task template $T_G$, like shown in the box on the left of Figure~\ref{fig:gen-val-arch}, which in this case, samples a table $R$ with column \code{states}, that asks the Generator model $M_G$ to create a realistic error based on $R$.

Invoking $M_G$ on $t_G$ creates an actual completion, $c = M_G(t_G)$, shown in the lower half of the left box, which in this case is a typo error \codeq{Missisipi} that may realistically occur in $R$. 

This completion $c$ is then used to construct a classification-version of Error detection $t_C$, where we perform transformation $f$ by inserting the created error \codeq{Missisipi} into the original column $R$, to create the input table for $t_C$ shown in the right box.

Next, we invoke the Validate() subroutine (Algorithm~\ref{alg:validate-for-classification-task}), to validate whether $c$ (the completion \codeq{Missisipi}) is the correct completion for our $t_C$. Specifically, like illustrated in Algorithm~\ref{alg:validate-for-classification-task}, we would perform repeated ``permutation'' of the  table in task $t_C$, creating many variants $t_C'$ shown in Figure~\ref{fig:model-validation-error} (note that rows inside each task box in the figure are ordered differently).  We then invoke $M_C$ on each $t_C'$, and we expect the completion $c$  (\codeq{Missisipi}) to be consistently produced if $c$ is an actual error\footnote{This is assuming that there are no additional error in the original table -- if the original table has other error, then the completion of $t_C'$ would not be consistently  $c$ (``{Missisipi}''), and we will also not validate this $(t_C, c)$ pair for downstream training.}. Note that for classification, we would also need negative examples, in this case we directly sample real table column, and perform validation also using Algorithm~\ref{alg:validate-for-classification-task}, where the expected $c$ is an empty set.

If a pair ($t_C, c$)   can be consistently validated using $M_C$  with permutation, we treat $(t_G, c)$ and $(t_C, c)$ as good training examples for $M_G$ and $M_C$, which we add into their respective training sets (in this case, the tasks shown in the left and right box of Figure~\ref{fig:model-validation-error}, and their completion \codeq{Missisipi}).

We iterate the preceding steps to sample and validate a training data point, for ``step-size'' number of times (e.g., 3000), and then fine-tune $M_G$ and $M_C$ on validated training data, where the hope is that fine-tuning the resulting $M_G$ and $M_C$ models can be better than the original vanilla models $M$. 
We iterate the fine-tuning process for a few iterations (up to 3), and return the resulting models as our specialist models. % In this case, the fine-tuned $M_C$ model is our desired model that specializes in detecting errors (better than the vanilla $M$). 
\end{example}

Note that in the fine-tuning process, because we sample diverse real tables $R$ to construct table-tasks  $t_C$ and $t_G$ for training (instead of using a small manually-labeled dataset), the resulting models are less likely to ``over-fit'', and are more likely generalize well.

%Also note that through repeated validation, our validation procedure only retains high-quality training examples $t_C$ and $t_G$ for fine-tuning, ensuring the quality of the resulting fine-tuned models. 

\begin{figure}[t]
        \vspace{-4mm}
    \centering
        \centering
        \includegraphics[width=0.4\textwidth]{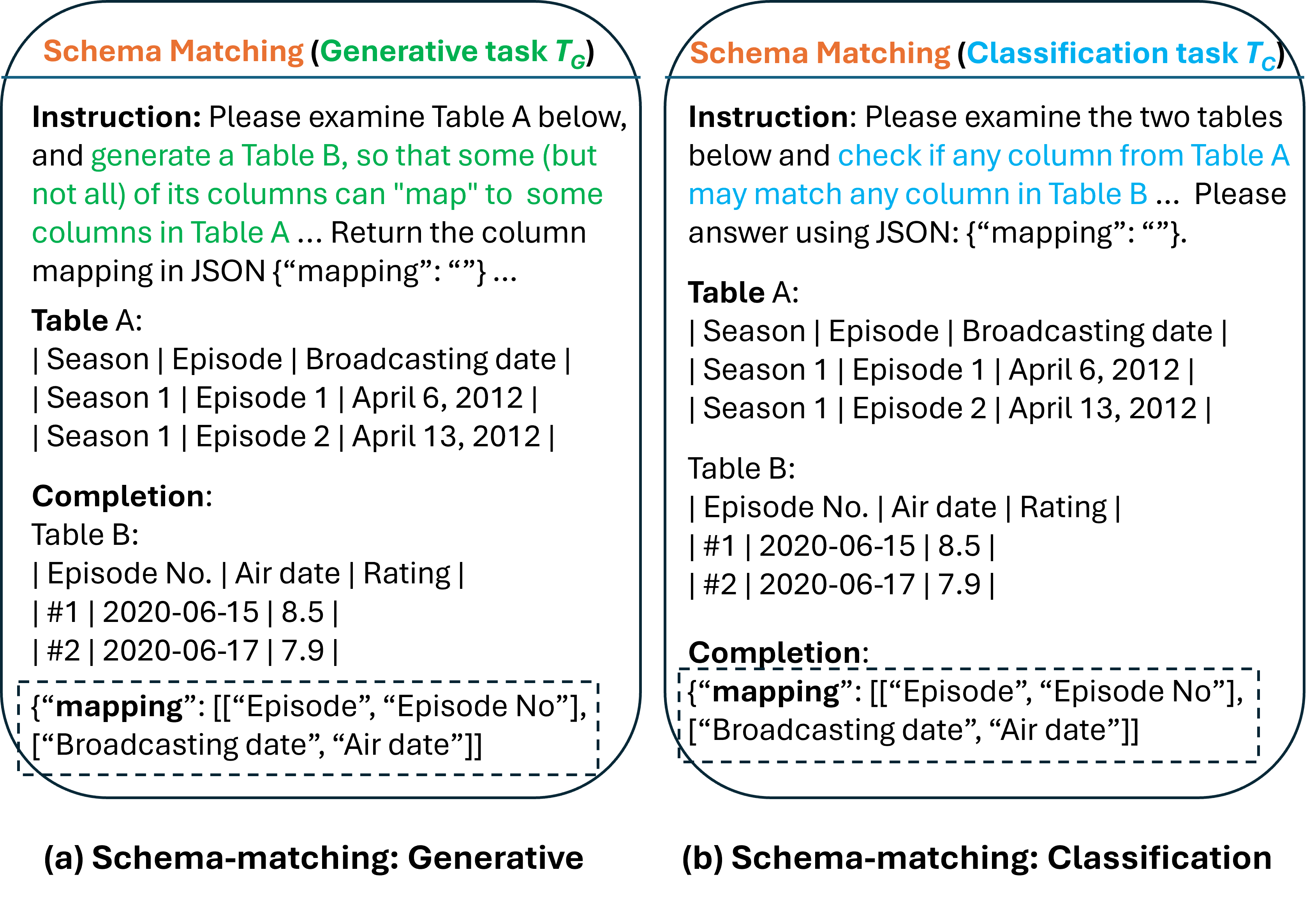}
        \vspace{-5mm}
        \caption{Schema matching task duality: (a) its generative version; and (b) classification version. We can perform Generator-Validator fine-tuning for Schema matching, by using these two boxes in Figure~\ref{fig:gen-val-arch}.}
        \vspace{-5mm}
        \label{fig:example-tasks-schema-matching}
\end{figure}

As another example, we explain the end-to-end process for a different task, Schema matching~\cite{schema-matching-cupid, schema-matching-valentine, schema-mapping-survey}, that takes \emph{two tables} as input (recall that in Schema matching, our goal is to identify column pairs that refer to the same concept from two input tables).

\begin{figure*}[t]
\vspace{-19mm}
    \centering
        \centering
        \includegraphics[width=0.9\textwidth]{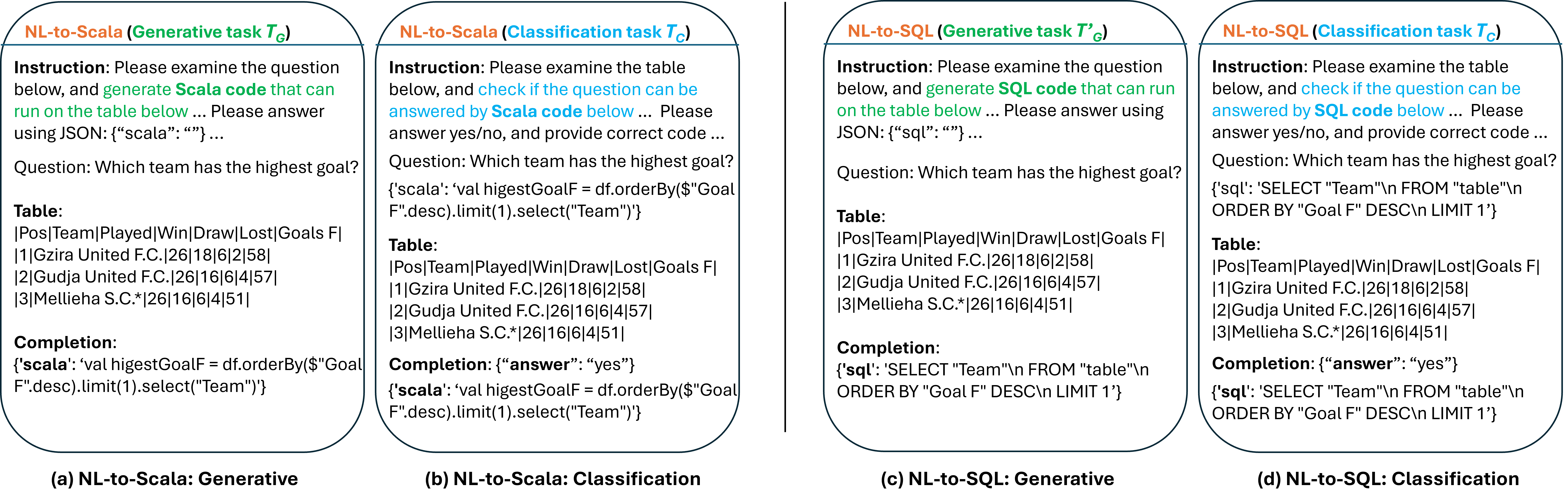}
                  \vspace{-4mm}
        \caption{Two example NL-to-Code tasks that translate natural-language questions to code: (a, b) dual versions of the NL-to-Scala task; (c, d) dual versions of the NL-to-SQL task. ``{Execution-invariance}'': observe that in (a)  NL-to-Scala, and (c)  NL-to-SQL, given the same question, the generated Scala and SQL code should generate identical results when executed on the same input table.}
        \vspace{-3mm}
        \label{fig:example-tasks-NL2Code}
\end{figure*}

\begin{example}
\label{ex:schema-matching}
[Schema matching]. Figure~\ref{fig:example-tasks-schema-matching}(b) shows the traditional, classification-version of Schema matching studied in the literature, which is verbalized as a task that requires language-models to inspect two tables, Table-A and Table-B, and produce column-mappings in a JSON list. Exploiting ``duality'', we construct its generative-version, shown in Figure~\ref{fig:example-tasks-schema-matching}(a), which takes one table, Table-A, as input, and is asked to produce another table, Table-B, with possible column-mappings between the two tables. 

We can then use Figure~\ref{fig:example-tasks-schema-matching}(a) and (b) to replace the two boxes in Figure~\ref{fig:gen-val-arch}, and similarly perform Generator-Validator fine-tuning.

We start by sampling a real table $R_A$ and populate it as Table-A in the generative-version of Schema matching  (Figure~\ref{fig:example-tasks-schema-matching}(a)), to produce an instance of this task $t_G$. Invoking the Generator model $M_G$ (initially the vanilla model $M$) produces the completion (lower half of Figure~\ref{fig:example-tasks-schema-matching}(a)), which includes another table $T_B$ and mapping between $(T_A, T_B)$, denoted by $c_m$ (a completion for mapping), shown as a JSON list at the bottom. 

We then construct a corresponding task $t_C$ shown in Figure~\ref{fig:example-tasks-schema-matching}(b), by concatenating $(T_A, T_B)$ as input for Schema matching (which is the transformation $f$ in Definition~\ref{def:task-duality}). By duality, if the generative-task $t_G$ generates $c_m$ correctly, then we expect its classification dual $t_C$ would produce the same $c_m$ as completion, as marked in two rectangle boxes with the same JSON content in the figure.

We therefore repeatedly permute both tables $T_A$ and $T_B$ in task $t_C$, into a new task $t_C'$, and invoke  $M_C(t_C')$ to see if we  get the expected $c_m$ consistently\footnote{Note that we parse the two JSON lists for order-insensitive equality comparisons, and do not require the two JSON to be identical verbatim.}. If validated, we use the corresponding $(t_G \rightarrow T_B, c_m)$ and $(t_C \rightarrow c_m)$ as training data, to fine-tune $M_G$ and $M_C$, respectively. 
\end{example}

With Example~\ref{ex:error-detect} and~\ref{ex:schema-matching}, we can see how Generator-Validator fine-tuning in Algorithm~\ref{alg:gen-val} can  apply to other classification table-tasks. 
\iftoggle{full}
{
    Details of the prompts used in our generative-task $T_G$ and classification-task $T_C$ for Schema matching and Error detection, can be found in Appendix~\ref{apx:tasks}.
}
{
    Additional details for these tasks can be found in our technical report~\cite{full}.
}

\section{Table-Specialist: Generative Tasks}
\label{sec:generative-tasks}
% overview of code generation task

% Fine-tune the "generator" model

% NL-to-Code: 

%     - high/low resource?
    
%     - Two pair of pairwise execution validation

% Table to question:

% text-book like generation

% validation:

%     -- no ground truth: cross validation across PL (NL-to-Code); shuffle row-column for multiple execution runs to improve validity
    
%     -- with ground truth: validation with GT directly (R2R )

% % ------------------------------------------

% Generative tasks are ubiquitous in table related applications, particularly in automating data manipulation and transformation. These tasks require generating new outputs, such as code snippets, based on given inputs. In this section, we focus on two critical generative tasks: Natural Language to Domain Specific Language (NL-to-Code) and Row-to-Row Transformation (R2R). These tasks demonstrate the versatility and power of the \sys framework for \textit{generative} table tasks.

In this section, we describe how Generator-Validator fine-tuning can be applied to generative table-tasks (the lower half of Table~\ref{tbl:tasks}), such as NL-to-Code and Data-transformation, etc. Figure~\ref{fig:example-tasks-NL2Code} show two generative NL-to-Code tasks on tables, NL-to-Scala and NL-to-SQL, and their respective  classification duals.

Our fine-tuning process for generative table-tasks uses the same Generator-Validator approach in Algorithm~\ref{alg:gen-val}, thanks the the symmetry between generative/classification tasks in our setup.

As additional opportunities, we observe that for a subset of generative table-tasks, such as code-generation (e.g., NL-to-Code and Data-transformations), where the target code can be in languages such as SQL, R, Scala, Pandas,  then in addition to using the model-based validation in Line~\ref{ln:validate-data} of Algorithm~\ref{alg:gen-val} (which invokes Algorithm~\ref{alg:validate-for-classification-task}), we can also leverage a unique property of executing code on tables for validation, that we call ``\emph{execution-invariance}'' described below.

\begin{proposition}
\label{prop:execution-invariance}
[Execution invariance]. Given a task $T$ specified on a table $R$, let $c^L$ be the generated code in a language $L$ that can execute on $R$ to correctly solve $T$, and $c^{L'}$ be the generated code in a different language $L'$ that can also  solve $T$.
Let $R_S \subseteq R$ be a table with a subset of rows of $R$, then  for any $R_S$, we have $c^L(R_S) \equiv c^{L'}(R_S)$, meaning that the execution of $c^L$ and $c^{L'}$ on any $R_S \subseteq R$ should  always produce identical results. 
\end{proposition}

This is intuitive, because given any task $T$, if the code $c^L$ and $c^{L'}$ generated in two languages can both correctly perform task $T$, they must be semantically equivalent, and their execution results must be identical, on $R$ or any of its subsets $R_S \subseteq R$.  

\begin{example}
    \label{ex:execution-invariance}
    [Execution invariance]. Figure~\ref{fig:example-tasks-NL2Code}(a) and (c) show two generative NL-to-Code tasks, NL-to-Scala and NL-to-SQL, respectively. Given the same question (e.g., \codeq{which team has the highest goal}), the generated Scala and SQL code shown at the bottom of the boxes, should always produce the same results when executed on the same table $R$ (or its subset $R_S$), shown in the figure.
\end{example}

The execution-invariance property provides us with an alternative to model-based validation (Algorithm~\ref{alg:validate-for-classification-task}), by using execution-based validation, which we explain in Algorithm~\ref{alg:validate-for-generative-task} below. 
%serves as a drop-in replacement of  Algorithm~\ref{alg:validate-for-classification-task} (in Line~\ref{ln:validate-data} of Algorithm~\ref{alg:gen-val}).

%\yeye{change main demo task into NL-to-scala?}

%\begin{small}
\begin{algorithm}[h]
\scalebox{0.8}
{
    \begin{minipage}{1.2\linewidth}
    \begin{algorithmic}[1]
    \SetKw{kwReturn}{return}
    \caption{Validate($M_G^L, M_G^{L'}, t_G$): for code-generative tasks}
    \label{alg:validate-for-generative-task}
     \REQUIRE{A generative model $M_G^L$ for generating code in a target language $L$, another generative model $M_G^{L'}$ for generating code in a second language $L'$, an instance of classification task $t_G$}
     \ENSURE{True or False} %\tcp{\footnotesize{validate whether $c$ is the correct completion for $t_C$}}

    \STATE $R \leftarrow t_G.R$ \tcp{\footnotesize{get the table $R$ used in task $t_G$}}  \label{ln:assign-R-code-task} 
    
    \STATE $c^L \leftarrow M_G^{L}(t_G)$ \tcp{\footnotesize{generate target code $c^L$ in language $L$}}  \label{ln:gen-code-lang-L} 
    
    \STATE $c^{L'} \leftarrow M_G^{L'}(t_G)$ \tcp{\footnotesize{generate target code $c^{L'}$ in language $L'$}}  \label{ln:gen-code-lang-Lprime} 

    \FOR{$i$ in 1 to $N$}  \label{ln:iter-validate}
        \STATE $R_S \leftarrow $ Sample($R$) \tcp{\footnotesize{sample rows in table $R$}}  \label{ln:sample-R} 

        \STATE $r \leftarrow \text{Execute}(c^L, R_S)$ \tcp{\footnotesize{execute $c^L$ on table $R_S$ to get $r$}}  \label{ln:execute-1} 
        
        \STATE $r' \leftarrow \text{Execute}(c^{L'}, R_S)$ \tcp{\footnotesize{execute $c^{L'}$ on table $R_S$ to get $r'$}}  \label{ln:execute-2} 
        
        \IF{($r \neq r'$)} \label{ln:check-eq-result}

            \STATE \kwReturn False \tcp{\footnotesize{Not-validated: unsure if $c^L$ is correct  completion for $t_G$}}
             
        \ENDIF

    \ENDFOR

    \kwReturn True \tcp{\footnotesize{Validated: $c^L$ is likely a correct completion for $t_G$}}
    \end{algorithmic}
    \end{minipage}
}
\end{algorithm}

In Algorithm~\ref{alg:validate-for-generative-task}, we are given a generative model $M_G^L$ that can generate code on tasks $t_G$ in a target language $L$ (e.g., NL-to-Scala). We use a second model  $M_G^{L'}$ that generates code for the same task $t_G$ but in a different language $L'$ (e.g., NL-to-SQL), in order to validate code generated by $M_G^L$.

We start by assigning $R$ as the table used in $t_G$, then invoke $M_G^L$ and $M_G^{L'}$ (both are initially vanilla language models), to generate code $c^L$ and $c^{L'}$ respectively. Then in $N$ iterations, we repeatedly sample rows to generate $R_S \subseteq R$, and execute   $c^L$ and $c^{L'}$ on $R_S$, to produce results $r$ and $r'$, respectively. If in any iteration we have $(r \neq r')$, then by execution-invariance we know that $c^L$ and $c^{L'}$ is not semantically equivalent, and at least one of the two is incorrect, which is why we return \codeq{False} to signify that $c^L$ cannot be validated so that it will not be used in training later. Otherwise, if we cannot find contradictions in $N$ iterations, we consider   $(t_G, c^L)$ a valid training example and return \codeq{True} for this data point to fine-tune $M_G^L$. Note that $(t_G, c^{L'})$ is also a valid training example, so that we can  fine-tune $M_G^{L'}$ for a different language $L'$ in parallel.

We illustrate Algorithm~\ref{alg:validate-for-generative-task} using NL-to-Code as an example.
\begin{example}
    \label{ex:validate-code-gen-task}
    Consider the task of NL-to-Scala, or generating Scala code that can run on Spark, as shown in Figure~\ref{fig:example-tasks-NL2Code}(a). Like in Figure~\ref{fig:gen-val-arch}, as pre-processing steps, we would first sample a real table $R$, and then ask language-models to brainstorm a question relevant to table $R$, e.g., \codeq{which team has the highest goal} for the table in the figure, to create a generative task $t_G$.  The classification version of the task is shown in Figure~\ref{fig:example-tasks-NL2Code}(b), which asks a model to predict whether a code snippet can execute to answer a given natural-language question.     
    With these two tasks, we can already perform Generator-Validator fine-tuning using Algorithm~\ref{alg:gen-val} and~\ref{alg:validate-for-classification-task}. 

    Leveraging execution-invariance, we can perform a different type of validation, that invokes Algorithm~\ref{alg:validate-for-generative-task} (in place of Algorithm~\ref{alg:validate-for-classification-task}). Specifically, when validating training data (Line~\ref{ln:validate-data} of Algorithm~\ref{alg:gen-val}), we  invoke Algorithm~\ref{alg:validate-for-generative-task}, where we use the same task, but require code to be generated in a different language -- Figure~\ref{fig:example-tasks-NL2Code}(c) shows an NL-to-SQL task that directly corresponds to Figure~\ref{fig:example-tasks-NL2Code}(a) but requires generated code to be in SQL.

    Let $M_G^L$ be the NL-to-Scala model that we iteratively fine-tune, and  $M_G^{L'}$ be a NL-to-SQL model (which can be a vanilla language-model, or another model that we also iteratively fine-tune in lockstep),  we can then proceed to invoke Algorithm~\ref{alg:validate-for-generative-task}. We first generate code in both Scala and SQL for the same question, like shown in the bottom of Figure~\ref{fig:example-tasks-NL2Code}(a) and (c), and then execute both Scale and SQL repeatedly on sub-samples $R_S \subseteq R$, to compare their execution results. If we cannot find contradictions in any iteration, we consider $(t_G, c^L)$ and $(t_G, c^{L'})$  validated, which we can use to iterative fine-tune $M_G^L$ and $M_G^{L'}$.   (This in effect changes the right-half of the architecture in Figure~\ref{fig:gen-val-arch}, by  replacing the mode-based validation, into an execution-based validation). 
\end{example}

Note that the execution-based validation applies to other generative tasks involving code, such as Data-transformation by-example, or generating code to perform transformations specified by input/output examples, using a target language (e.g., SQL, R, Scala, etc.), which we will also study in our experiments.

%\textbf{Textbook-like generation.}

\iftoggle{full}
{
        
    \underline{``Textbook-like'' generation}~\cite{synthetic-slm-phi3, synthetic-slm-phi15}. 
    There are additional details in our initial data generation process, where we use a curriculum-guided process to direct language-models to compose textbook constructs so that they can generate diverse questions of varying levels of difficulty that are  relevant to a given table $R$ that we will explain below.

    Recall that for generative tasks such as NL-to-Code and Data-transformations, there is an initial preprocessing step in which we need to generate a reasonable ``task'' $t$ for a sampled table $R \in \mathcal{C}$, so that the the question $t$ can then become part of the instruction and used as training data to fine-tune language models (e.g., for NL-2-Code, this $t$ would be a natural language question that needs to be answered based on the content of $R$, for Data-transformation, this $t$ would be a ground-truth transformation that we want models to predict based on input/output examples). 
    
    While language-models can by themselves generate reasonable tasks $t$ given a table $R$, we find benefit in guiding language-models towards constructing  diverse $t$ by composing basic building-blocks from programming language ``textbooks''. 
    
    For example, for NL-2-Code, we find it beneficial to  decompose the question-generation task, into an explicit list of requirements based on atomic SQL constructs, such as where, group-by, order-by, etc.,  using database textbooks~\cite{db-textbook}, and then ask language-models to brain-storm a question using a given table $R$, based on the constraints, like below:    
    \begin{itemize}
        % \item 
    % \end{itemize}
        \item *THREE (3)* of filtering predicate(s) in WHERE clause
        \item *TWO (2)* GROUP BY clause, with aggregation function
        \item *ONE (1)* ORDER BY command
        \item ...
    \end{itemize}
    Note that the numbers 3/2/1 shown above are examples, which are randomly sampled from a range of $[0,k]$, and dynamically inserted into the prompt when generating a question $q$ on a table $R$. This produces  diverse questions with varying degrees of difficulty, that can be answered using SQL or otherwise. (In comparison, if language models are asked to generate questions unconstrained, they tend to produce similar questions on different tables that are less diverse, which we find to be  less effective as training data). 

    Similarly, for Data-transformation, we decompose the task of generating reasonable transformations that can be performed on a given table $R$, also into a list of atomic building blocks, using basic Python constructs, such as string-transformation (split, concatenate, sub-string, etc.), number transformation, array transformations, etc., using Python textbooks~\cite{python-textbook}. We sample requirements from the list, in order for language-models to generate diverse transformation examples that can be used as training data. 

    Details of the prompts used in generating task $t$ for NL-2-Code and Data-transformation, can be found in Appendix~\ref{apx:tasks}.
    
}
{
    \underline{Additional details.} Details of our fine-tuning, such as task data generation (e.g., using ``textbook-like generation''~\cite{synthetic-slm-phi3, synthetic-slm-phi15}, or a curriculum to guide language-models to compose textbook constructs so that they can generate diverse questions of varying levels of difficulty that are  relevant to a given table $R$). 
    We provide these additional details in our technical report~\cite{full}.  
}

\iftoggle{full}
{
    \underline{Things we tried but were not effective.} In addition to what is reported, we also tried many things that did not turn out to be effective, which we will report below.

    Since it is standard to use confidence scores as soft-labels in self-supervised and semi-supervised learning~\cite{self-train-1,self-train-2, self-train-google-noisy-student}, in \sys we also tried to extract confidence scores of training examples from language-models during the generation process. For example, we used log-probabilities~\cite{llm-confidence-openai-logprob} as well as verbalization techniques~\cite{llm-table-understanding-1, llm-confidence-2} to extract confidence from language-models, which we use to find confident training examples in the self-training / iterative fine-tuning process, which however was not always beneficial.

    During the data validation process, we produce lots of negative (invalidated) examples, in addition to positive (validated) examples. In one variant of our fine-tuning, we try to use both positive (validated) and negative (invalidated) examples (e.g., using a format suggested in~\cite{ft-neg-examples}), to prefix positive and negative examples with leading special-tokens, such as [POS] and [NEG], respectively, which was not helpful in our experiments. 

     Since some of the table-tasks we test are pretty challenging (e.g., Data-transformations, which requires trial-and-test, and reflect on previous mistakes), we also tried agentic self-reflection style fine-tuning using trajectories, by allowing language-models to make multiple attempts in generating transformation-programs, each looking at the output from previous attempts to reflect on previous errors (e.g., compilation errors in previous execution, or output from a previous execution does not match the intended output), similar to~\cite{shinn2024reflexion} in NLP tasks. While it provides modest benefit in terms of overall success rate for challenging tasks like Data-transformation, it substantially increases the latency for training data generation, and complicates the overall architecture, which we decided not to include in the end. 
}
{
    \underline{Things we tried but were not effective.} In addition to what is reported, we also tried many things that did not turn out to be effective. For example, we tried to extract confidence scores of training examples from language-models (e.g., using log-probabilities)~\cite{llm-table-understanding-1, llm-confidence-2, llm-confidence-openai-logprob}, as a form of soft-labels in our self-training, which was not beneficial. We tried to fine-tune using both positive (validated) and negative (invalidated) examples~\cite{ft-neg-examples}, but that was not helpful. We also tried agentic self-reflection style fine-tuning using trajectories, which was also not too beneficial~\cite{shinn2024reflexion}. We give more details of these attempts in~\cite{full}.
}

\section{Experiments}
\label{sec:exp}

We perform extensive experiments, using GPT-3.5, GPT-4, and Llama-3.1-8B as base models. Our code is available at \href{https://github.com/microsoft/Table-Specialist}{\faGithub~microsoft/Table-Specialist}.

%\yeye{mention why not large models teach small models}

\subsection{Experiment Setup}
\label{sec:exp-setup}

\begin{table}[th!]
    \centering
    
    \resizebox{\linewidth}{!}
    {
    \begin{tabular}{|c|c|c|c|c|}
    \hline
        \textbf{Table-task group} & \textbf{\makecell{Evaluation \\ metric}} & \textbf{\makecell{Task \\ category}}  & \textbf{Dataset} & \textbf{Size}  \\\hline
        \multirow{5}{*}{\makecell{NL-to-Code \\ (NL-to-SQL, NL-to-R,\\ NL-to-Scala)} } & \multirow{5}{*}{\makecell{Execution\\Accuracy}} & \multirow{5}{*}{\makecell{easy}} & WikiSQL & 1000  \\\cline{4-5}
                                         &                   &  & Spider   & 1198 \\\cline{4-5}
                                         &                   &  & BIRD   & 356 \\\cline{4-5}
                                         &                   &  & WikiTQ   & 1000 \\\cline{4-5}
                                         &                   &  & Text2Analysis   & 271  \\\hline

    \multirow{3}{*}{{Table-QA}}    & \multirow{3}{*}{Accuracy}  & \multirow{3}{*}{\makecell{easy}} & FinQA & 1000  \\\cline{4-5}
                                         &                   &  & TableBench   & 424  \\\cline{4-5}
                                         &                   &  & WikiTQ       & 1000  \\
                                         \hline
                                         
        \multirow{2}{*}{\makecell{Data transformation \\ (generating SQL, R, Pandas)}} & \multirow{2}{*}{\makecell{Execution\\Accuracy}} & \multirow{2}{*}{\makecell{hard}} & TDE & 570 \\\cline{4-5}
                                         &                   &  & Transform-text   &  335 \\\hline        
        \multirow{3}{*}{Schema matching} & \multirow{3}{*}{F1}  & \multirow{3}{*}{\makecell{easy}} & DeepM & 42 \\\cline{4-5}
                                         &                   &  & WikiData   & 24   \\\cline{4-5}
                                         &                   &  & HXD   & 468 \\\hline
        \multirow{2}{*}{Error detection} & \multirow{2}{*}{F1}  & \multirow{2}{*}{\makecell{hard}} & Spreadsheet-Tables & 1126  \\\cline{4-5}
                                         &                   &  & Relational-Tables   & 1081  \\\hline

    \end{tabular}
    }
    % \end{small}
    \caption{Table task and benchmark data for evaluation}
    \label{tbl:benchmarks}
    % \vspace{-5mm}
\end{table}

\stitle{Table tasks and benchmarks.} 
% Table~\ref{tbl:benchmarks} shows the list of table tasks and their corresponding benchmarks used in our evaluation. 
For a comprehensive evaluation, we use three sets of three generative tasks, NL-to-Code (generating SQL, R, Scala), Data-transformation (generating SQL, R, Pandas) and Table-QA; as well two  classification tasks, Error detection and Schema matching, for a total of 9 table tasks. Each task is extensively evaluated using 2-5 benchmarks from the literature, as shown in Table~\ref{tbl:benchmarks}, for a total of 29 evaluated benchmarks
(each benchmark corresponds to a row in our main result in Table~\ref{table:quality_comparison} and Table~\ref{tbl:quality_comparison_tqa}).

\begin{small}
\begin{table*}[t]
 % \vspace{-10mm}
\centering
    % \vspace{-4mm}
\resizebox{\textwidth}{!}
% \scalebox{0.6}
{
\begin{tabular}{|c|c|c||c|c||c|c||c|c|}
\hline
\small
\multirow{3}{*}{\textbf{Task Type}} & \multirow{3}{*}{\textbf{Task}} & \multirow{3}{*}{\textbf{Dataset}} & \multicolumn{2}{c||}{\textbf{\makecell{Specialist Fine-tuning\\(GPT-3.5)}}}  & \multicolumn{2}{c||}{\textbf{\makecell{Specialist Fine-tuning\\(GPT-4)}}} & \multicolumn{2}{c|}{\textbf{\makecell{Specialist Fine-tuning\\(Llama3.1-8B)}}}  \\\cline{4-9}
 & & & Vanilla & \makecell{\textsc{Table}\\\textsc{Specialist}} & Vanilla & \makecell{\textsc{Table}\\\textsc{Specialist}} & Vanilla & \makecell{\textsc{Table}\\\textsc{Specialist}}  \\\hline

%%%%                                                                                    | GPT-3.5  | Ours  |35 w/o | TableLlama | TableGPT | gpt4|   
\multirow{7}{*}{Classification} & \multirow{4}{*}{Schema Matching}     & DeepM              & 0.984  & \textbf{1}      & 1 & 1  & 0.857 & \textbf{1} \\%\cline{3-9}
                            &                                      & WikiData           & 0.913   & \textbf{0.918}  & 0.952 & 0.952 & \textbf{0.912} & 0.766\\%\cline{3-9}
                            &                                      & HXD                & 0.878   & \textbf{0.897}  & 0.924 & \textbf{0.935} & 0.749 & \textbf{0.852}\\%\cline{3-9}
                            &                                      & \textbf{Average}   & 0.925   & \textbf{0.938}  & 0.959 & \textbf{0.965} & 0.839 & \textbf{0.873}\\ \cline{2-9}
                            & \multirow{3}{*}{Error Detection}     & Spreadsheet-Tables & 0.136  & \textbf{0.207}  & 0.403 & \textbf{0.458}  & 0.071 & \textbf{0.136} \\%\cline{3-9}
                            &                                      & Relational-Tables  & 0.340  & \textbf{0.457}  & 0.465 & \textbf{0.529}  & 0.108 & \textbf{0.161} \\%\cline{3-9}
                            &                                      & \textbf{Average}   & 0.238  & \textbf{0.332}  & 0.434 & \textbf{0.494}  & 0.090 & \textbf{0.148} \\ \hline

%%%%                                                                                                            | GPT-3.5 | Ours |35 w/o | TableLlama | TableGPT | gpt4|   
\multirow{31}{*}{Generative} 
                            & \multirow{6}{*}{\makecell{NL-to-SQL %\\(SQL -- SparkScala)
                            }}                                                              & WikiSQL           & 0.823  & \textbf{0.855}   & 0.869  & \textbf{0.874} & 0.525 & \textbf{0.816} \\%\cline{3-9}
                            &                                                               & WikiTQ            & 0.421  & \textbf{0.513}   & 0.559  & \textbf{0.597} & 0.300 & \textbf{0.449} \\%\cline{3-9}
                            &                                                               & Text2Analysis     & 0.498  & \textbf{0.517}   & \textbf{0.581} & 0.572 & 0.273 & \textbf{0.465} \\%\cline{3-9}
                            &                                                               & Spider            & 0.650  & \textbf{0.684}   & 0.694 & \textbf{0.704} & 0.670 & \textbf{0.690}  \\%\cline{3-9}
                            &                                                               & BIRD              & 0.452  & \textbf{0.514}   & 0.528 & \textbf{0.556} & 0.388 & \textbf{0.438}  \\%\cline{3-9}
                            &                                                               & \textbf{Average}  & 0.569  & \textbf{0.616}   & 0.647 & \textbf{0.661} & 0.431 & \textbf{0.572} \\\cline{2-9}
                            & \multirow{6}{*}{\makecell{NL-to-R 
                            %\\(R -- SparkScala)
                            }}                                                              & WikiSQL           & 0.567  & \textbf{0.776~$\mathbf{^*}$} & 0.759 & \textbf{0.827} & 0.331 & \textbf{0.409} \\%\cline{3-9}
                            &                                                               & WikiTQ            & 0.209  & \textbf{0.404}               & 0.416 & \textbf{0.550} & 0.138 & \textbf{0.257} \\%\cline{3-9}
                            &                                                               & Text2Analysis     & 0.227  & \textbf{0.358}               & 0.382 & \textbf{0.446} & 0.103 & \textbf{0.199} \\%\cline{3-9}
                            &                                                               & Spider            & 0.530  & \textbf{0.565~$\mathbf{^*}$} & 0.563 & \textbf{0.605} & 0.503 & \textbf{0.536}  \\%\cline{3-9}
                            &                                                               & BIRD              & 0.317  & \textbf{0.404}               & 0.430 & \textbf{0.475} & 0.225 & \textbf{0.261} \\%\cline{3-9}
                            &                                                               & \textbf{Average}  & 0.370  & \textbf{0.502}               & 0.510 & \textbf{0.582} & 0.260 & \textbf{0.333}  \\\cline{2-9}
                            & \multirow{6}{*}{\makecell{NL-to-Scala
                            %\\(SparkScala -- R)
                            }}                                                              & WikiSQL           & 0.510 & \textbf{0.794~$\mathbf{^*}$} & 0.745 & \textbf{0.815} & 0.359 & \textbf{0.728} \\%\cline{3-9}
                            &                                                               & WikiTQ            & 0.109 & \textbf{0.426~$\mathbf{^*}$} & 0.198 & \textbf{0.476} & 0.043 & \textbf{0.188}\\%\cline{3-9}
                            &                                                               & Text2Analysis     & 0.236 & \textbf{0.373~$\mathbf{^*}$} & 0.258 & \textbf{0.373} & 0.129 & \textbf{0.214} \\%\cline{3-9}
                            &                                                               & Spider            & 0.308 & \textbf{0.504~$\mathbf{^*}$} & 0.294 & \textbf{0.466} & 0.249 & \textbf{0.356} \\%\cline{3-9}
                            &                                                               & BIRD              & 0.188 & \textbf{0.360~$\mathbf{^*}$} & 0.189 & \textbf{0.407} & 0.100 & \textbf{0.228}  \\%\cline{3-9}
                            &                                                               & \textbf{Average}  & 0.270 & \textbf{0.491~$\mathbf{^*}$} & 0.337 & \textbf{0.507} & 0.176 & \textbf{0.343} \\\cline{2-9}
                            & \multirow{3}{*}{\makecell{Data-transformation \\ (Pandas)}}  & TDE                            & 0.293 & \textbf{0.346}    & 0.418 & \textbf{0.456} & 0.137 & \textbf{0.161} \\%\cline{3-9}
                            &                                                                   & Transform-Text            & 0.227 & \textbf{0.230}    & \textbf{0.297} & 0.296 & 0.090 & \textbf{0.122}\\%\cline{3-9}
                            &                                                                   & \textbf{Average}          & 0.260  & \textbf{0.300}   & 0.357 & \textbf{0.376} & 0.113 & \textbf{0.142} \\\cline{2-9}
                            & \multirow{3}{*}{\makecell{Data-transformation \\ (R)}}       & TDE                            & 0.200  & \textbf{0.235}   & 0.305 & \textbf{0.318} & 0.063 & \textbf{0.105} \\%\cline{3-9}
                            &                                                                   & Transform-Text            & 0.164  & \textbf{0.215}   & \textbf{0.222} & 0.218 & 0.051 & \textbf{0.075} \\%\cline{3-9}
                            &                                                                   & \textbf{Average}          & 0.182  & \textbf{0.225}   & 0.264 & \textbf{0.268}  & 0.057 & \textbf{0.090} \\\cline{2-9}
                            & \multirow{3}{*}{\makecell{Data-transformation \\ (SQL)}}     & TDE                            & 0.144  & \textbf{0.168}  & 0.194 & \textbf{0.202} & 0.051 & \textbf{0.089}\\%\cline{3-9}
                            &                                                                   & Transform-Text            & 0.128  & \textbf{0.172}  & 0.216 & \textbf{0.227} & 0.063 & \textbf{0.066} \\%\cline{3-9}
                            &                                                                   & \textbf{Average}          & 0.136  & \textbf{0.170}  & 0.205 & \textbf{0.214} & 0.057 & \textbf{0.078}  \\\hline
                             % & \multirow{4}{*}{\junjie{TableQA}} & FinQA              & 0.222 & \textbf{0.261} & \multicolumn{2}{c||}{\multirow{4}{*}{n.a.\tablefootnote{\junjie{\code{GPT-4-0613} has retired at time of this revision.}}}}  & 0.066 & \textbf{0.145} \\%\cline{3-9}
                             % &                          & TableBench         & 0.322 & \textbf{0.336} & \multicolumn{2}{c||}{}                       & \textbf{0.277} & 0.261 \\%\cline{3-9}
                             % &                          & WikiTQ             & 0.546 & \textbf{0.579} & \multicolumn{2}{c||}{}                       & 0.465 & \textbf{0.486} \\%\cline{3-9}
                             % &                          & \textbf{Average}   & 0.364 & \textbf{0.392} & \multicolumn{2}{c||}{}                       & 0.270 & \textbf{0.296} \\\hline
\end{tabular}
}
\caption{Quality comparisons, between Vanilla models (GPT-3.5, GPT-4, Llama3.1-8B), and fine-tuned models. We use \textbf{bold} to indicate better performance after \sys fine-tuning, and we use $\mathbf{^*}$ to indicate fine-tuned GPT-3.5 models that can outperform vanilla GPT-4.}
\label{table:quality_comparison}
\end{table*}
\end{small}

\begin{table}[]
    \centering
\resizebox{\columnwidth}{!}
% \scalebox{0.6}
{
    \begin{tabular}{|c|c|c||c|c|}
    \hline
    \multirow{3}{*}{\textbf{Dataset}} & \multicolumn{2}{c||}{\textbf{\makecell{Specialist Fine-tuning\\(GPT-3.5)}}}  & \multicolumn{2}{c|}{\textbf{\makecell{Specialist Fine-tuning\\(Llama3.1-8B)}}}  \\\cline{2-5}
                           & Vanilla & \makecell{\textsc{Table}\\\textsc{Specialist}} & Vanilla & \makecell{\textsc{Table}\\\textsc{Specialist}}  \\\hline
         FinQA              & 0.222 & \textbf{0.261}  & 0.066 & \textbf{0.145} \\%\cline{3-9}
         TableBench         & 0.322 & \textbf{0.336}  & \textbf{0.277} & 0.261 \\%\cline{3-9}
         WikiTQ             & 0.546 & \textbf{0.579}  & 0.465 & \textbf{0.486} \\%\cline{3-9}
         \textbf{Average}   & 0.364 & \textbf{0.392}  & 0.270 & \textbf{0.296} \\\hline
    \end{tabular}
}
    \caption{{Quality comparisons between vanilla models, and fine-tuned \sys models, on more open-ended generative task (Table-QA).}}
    \label{tbl:quality_comparison_tqa}
\end{table}

\begin{table}[]
    \resizebox{\linewidth}{!}
    {
    \begin{tabular}{c|c|cc}
    \toprule
        \multirow{2}{*}{\textbf{Task}}    & \multirow{2}{*}{\makecell{\textsc{Table}\\\textsc{Specialist}}}    & \multicolumn{2}{c}{Generalist Fine-tuning} \\
                                          &                          & TableLlama & TableGPT \\\midrule
        Schema Matching  & \textbf{0.938}     & 0.918      & 0.896  \\ \hline
        Error Detection  & \textbf{0.332}     & 0.227      & 0.222  \\ \hline
        NL-to-SQL        & \textbf{0.616}     & 0.576      & 0.570  \\ \hline
        NL-to-R          & \textbf{0.502}     & 0.373      & 0.370  \\ \hline
        NL-to-Scala      & \textbf{0.491}     & 0.279      & 0.304  \\ \hline
        \makecell{Data-transformation (Pandas)}& \textbf{0.300}     & 0.241      & 0.253  \\\hline
        \makecell{Data-transformation (R)}     & \textbf{0.225}     & 0.191      & 0.158  \\\hline
        \makecell{Data-transformation (SQL)}   & \textbf{0.170}     & 0.146      & 0.137  \\\bottomrule
    \end{tabular}
    }
    \caption{Comparisons between \sys and Table-Generalists (fine-tuned on GPT-3.5)}
    \label{tbl:quality_more}
\end{table}

\iftoggle{full}
{
We describe each task and benchmark dataset in turn below.

\underline{NL-to-Code (NL-to-SQL, NL-to-R, NL-to-Scala)}. The generative NL-to-Code task takes a table and a natural-language question as input, for a model to produce code that can be executed to answer the given question (Figure~\ref{fig:example-tasks-NL2Code}(a) shows an example).  We test generation in three target languages, SQL, R, and Scala, and  refer to the corresponding task as NL-to-SQL, NL-to-R, and NL-to-Scala.

We use five benchmarks for NL-to-Code: (1) WikiSQL~\cite{nl2sql-wikisql} is a common benchmark for NL-to-SQL; (2) Spider~\cite{nl2sql-spider} is another popular NL-to-SQL benchmark; (3) BIRD~\cite{nl2sql-bird} is a more recent and challenging benchmark for NL-to-SQL; (4) WikiTableQuestions (WikiTQ)~\cite{table-qa-wikitablequestions} is a popular table question-answering (QA) benchmark, which we use to test code generation by matching code-execution against QA ground-truth; and (5) Text2Analysis~\cite{DBLP:conf/aaai/HeZXMDDGJCHY024} is a recent benchmark for generating code to perform data analysis intents that are expressed in natural language. 

In each task, NL-to-SQL, NL-to-R, and NL-to-Scala, we use all five benchmarks above, for a total of $3 \times 5 = 15$ benchmark tests, to evaluate result quality based on the standard ``execution accuracy''~\cite{nl2sql-2, nl2sql-wikisql}. For example, we use the WikiSQL benchmark to evaluate not only NL-to-SQL, but also NL-to-R and NL-to-Scala tasks, by comparing the results of executing generated R and Scala, against the ground-truth execution results.

\underline{Data-transformation (generating SQL, R, Pandas)}. In this generative task, we are given pairs of user-specified input/output examples, and the goal is to synthesize the desired data-transformation program~\cite{data-transform-tde, data-transform-flashfill}. We generate target programs in SQL, R, and Pandas, respectively, as three table-tasks.
We use two existing benchmarks, TDE~\cite{data-transform-tde} and Transform-Text~\cite{prose-benchmark}, for a total of $3 \times 2 = 6$ benchmark tests, and report ``execution accuracy'' by comparing the result of executing generated code with ground truth.

Note that unlike NL-to-Code (a task language-models are familiar with), vanilla language-models struggle with the task of Data-transformation by-example, likely because relevant data is sparsely populated in the pre-training data, and  therefore represents a \codeq{hard} generative task to test, like shown in Table~\ref{tbl:benchmarks}.\footnote{While language models can often predict output examples directly by using input/output examples as few-shot prompts, generating executable code from these examples is more practical for scaling to large tables, which however, remains challenging.}

\underline{Schema matching.} Schema matching is the classification task to predict whether a pair of columns from two tables refer to the same concept (Figure~\ref{fig:example-tasks-schema-matching} shows an example). We used three benchmarks: DeepM and WikiData from~\cite{schema-matching-valentine}, and HXD from~\cite{zhang2024smutf}. We use the standard F1 score to report the quality of schema matching results~\cite{schema-mapping-survey, schema-matching-cupid, schema-matching-valentine}.

\underline{Error detection.}
Error detection is the classification task of predicting whether any value in a given column is erroneous or not (Figure~\ref{fig:example-tasks} shows an example). We use two benchmarks, Spreadsheet-Tables  and Relational-Tables, with over 1000 columns sampled from real spreadsheet tables and relational tables, respectively. Real errors in both benchmarks are manually labeled and available in~\cite{Full}.
 We also report the standard F1 score for result quality~\cite{error-detection-survey}.

Note that  Error detection  has a strong class imbalance (e.g., 98\% of real data are error-free), and often have  data that is not standard English (e.g., code-names and proprietary vocabularies), which makes it hard for language-models to produce high precision and high recall, and is therefore a \codeq{hard} classification task in our test.

%Details of the test benchmarks, such as their evaluation metrics and size statistics, can be seen in Table~\ref{tbl:benchmarks}.
}
{}

\stitle{Methods Compared.} We compare the following:
\begin{itemize}[leftmargin=*]
    \item \textbf{Vanilla base models}: \uline{GPT-3.5}\footnote{We use \code{GPT-3.5-turbo-1106}}, \uline{GPT-4}\footnote{We use \code{GPT-4-0613}}, and \uline{Llama-3.1-8b}.
    % \begin{itemize}[leftmargin=2mm]
    %     \item \uline{GPT-3.5}
    %     % ~\cite{llm-gpt-3}
    %     : We use \code{GPT-3.5-turbo-1106} %, to test its performance on table tasks. % This is also a base model we use, to fine-tune  \sys and other models.
    %     \item \uline{GPT-4}
    %     % ~\cite{llm-gpt4}
    %     : We use \code{GPT-4-0613}. 
    % \end{itemize}

    \item \textbf{Specialist Fine-Tuning}:
\uline{\sys}: Our proposed method, fine-tuned on GPT-3.5, GPT-4, and Llama-3.1-8b, respectively; \uline{FT-no-validation}: \sys fine-tuned model without the validation step, to isolate its impact.

\item \textbf{Generalist Fine-Tuning}: \uline{Table-GPT}~\cite{table-gpt} \footnote{Retrained using~\cite{tablegpt-training-data}}; \uline{TableLlama}~\cite{table-llama} \footnote{Retrained using ~\cite{tablellama-training-data}}.
% \begin{itemize}[leftmargin=2mm]
    % \item \uline{Table-GPT}~\cite{table-gpt}: A generalist model trained on diverse table tasks (retrained using~\cite{tablegpt-training-data}).
    % \item \uline{TableLlama}~\cite{table-llama}: Similarly re-implemented using its training data~\cite{tablellama-training-data} (as the released 7B model underperforms on our tasks).
% \end{itemize}

    % \item Generalist Fine-Tuning:
    % \begin{itemize}[leftmargin=2mm]
    %     \item \uline{Table-GPT}~\cite{table-gpt}. This is a Table-Generalist model, trained using a pool of table tasks, and designed to handle new and unseen table tasks. For a fair comparison, we use its  released training data~\cite{tablegpt-training-data} to fine-tune a model based on GPT-3.5.
    %     \item \uline{TableLlama}~\cite{table-llama}. This is another Table-Generalist model, and since their release model is  small (7B) and not performing well on our tasks, we also use its released training data~\cite{tablellama-training-data} to re-create TableLlama  based on GPT-3.5, for a fair comparison.
    % \end{itemize}
    
    % \item Dataset-specific Fine-Tuning:
    % \begin{itemize}[leftmargin=2mm]
    %     \item \uline{Dataset-specific Fine-Tuning (DS-FT)}. TOFILL
    % \end{itemize}

\end{itemize}

% \iftoggle{full}
% {
    We use Lora fine-tuning~\cite{hu2021lora}, with learning-rate multiplier of 0.5, and a batch size that is 1\% of training-data-size  (to ensure that each epoch has 100 steps), which is consistent across all methods.
% }
 %We use a step-size of 3000 examples (Algorithm~\ref{alg:gen-val}) for all tasks except Data-transformation, where we use a larger step-size of 30000 as the task is challenging and a large number of generated examples fail to be validated as training data).

%\stitle{Hyper-parameters.} We used the Azure OpenAI fine-tuning service~\cite{azure-open-ai-finetuning}. Each fine-tuning job was trained for two epochs with a learning-rate multiplier of 0.5. The batch size was adjusted so that each epoch consisted of 100 steps.

%\stitle{Generalist Fine-Tuning}. For a fair comparison between \sys and generalist models (i.e., Table-GPT and TableLlama), we fine-tuned two models based on GPT-3.5 using their released training data~\cite{tablegpt-training-data,tablellama-training-data}. For Table-GPT, we excluded training data for the four tasks implemented by \sys: NL-to-SQL, data transformation, error detection, and schema matching. Both models were trained for two epochs with a learning-rate multiplier of 0.01 and a batch size of 32. 

\iftoggle{full}
{
\begin{figure*}
\centering
\begin{subfigure}{.33\textwidth}
  \centering
  \includegraphics[width=\linewidth]{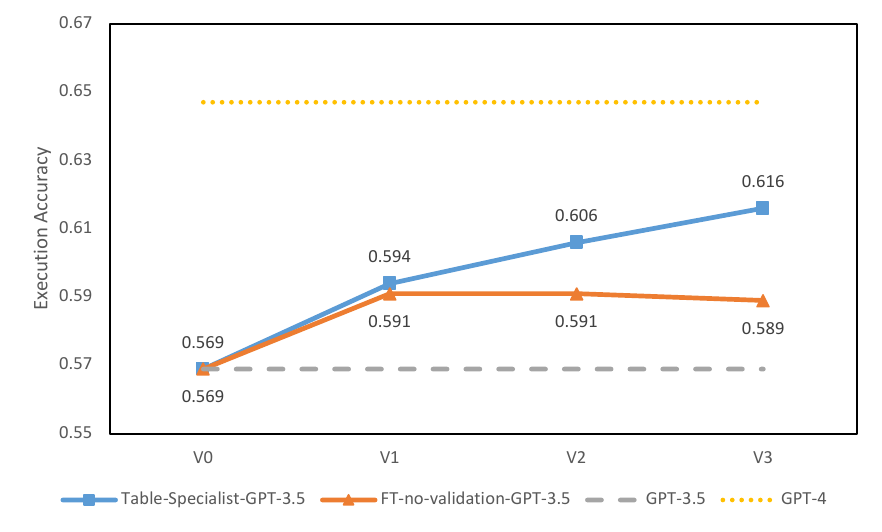}
  \caption{NL-to-SQL}
\end{subfigure}\hfill
\begin{subfigure}{.33\textwidth}
  \centering
  \includegraphics[width=\linewidth]{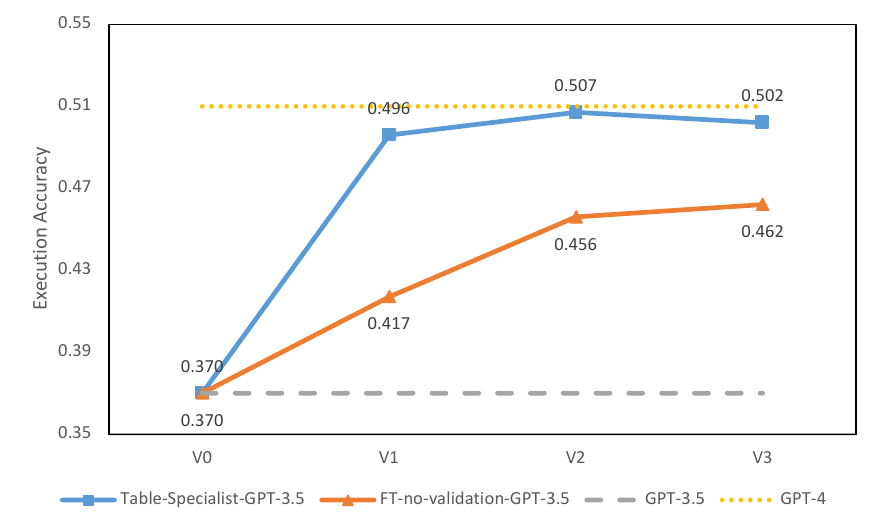}
  \caption{NL-to-R}
\end{subfigure}\hfill
\begin{subfigure}{.33\textwidth}
  \centering
  \includegraphics[width=\linewidth]{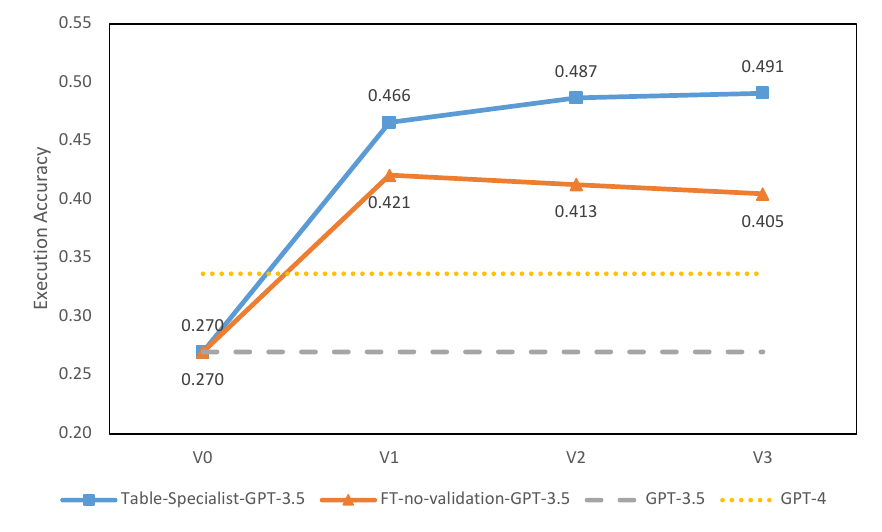}
  \caption{NL-to-Scala}
\end{subfigure}
\begin{subfigure}{.33\textwidth}
  \centering
  \includegraphics[width=\linewidth]{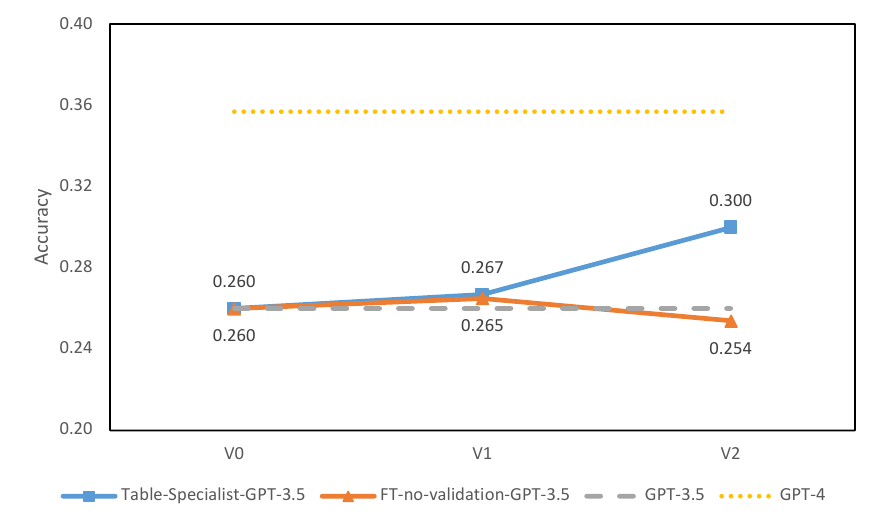}
  \caption{Data-transformation (Pandas)}
\end{subfigure}
\begin{subfigure}{.33\textwidth}
  \centering
  \includegraphics[width=\linewidth]{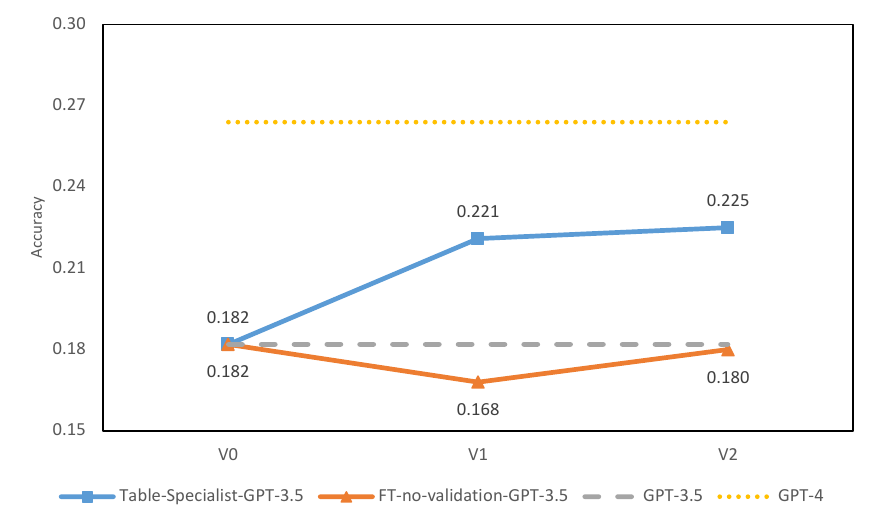}
  \caption{Data-transformation (R)}
\end{subfigure}
\begin{subfigure}{.32\textwidth}
  \centering
  \includegraphics[width=\linewidth]{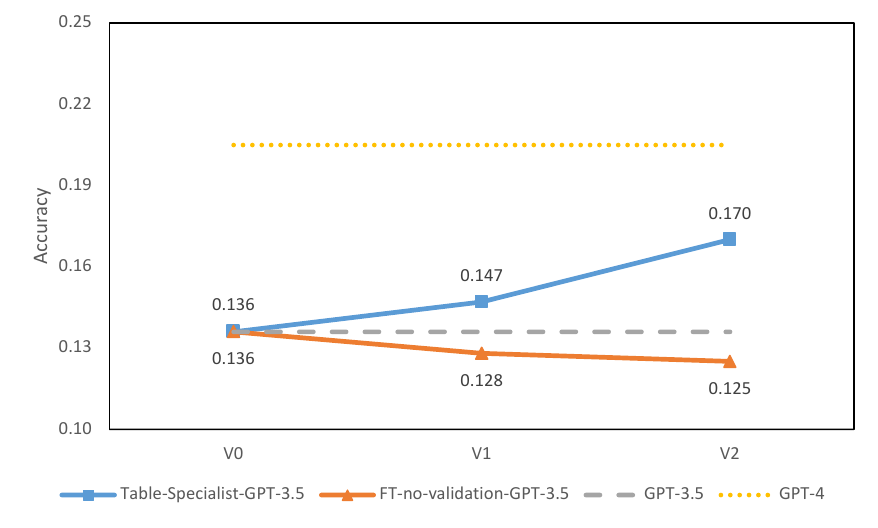}
  \caption{Data-transformation (SQL)}
\end{subfigure}
% \hspace*{\stretch{1}} % Margin on the left
% \begin{subfigure}{.33\textwidth}
%   \centering
%   \includegraphics[width=.9\linewidth]{figures/FT_by_iteration/ED-by-it.pdf}
%   \caption{Error Detection}
% \end{subfigure}\hfill
% \begin{subfigure}{.33\textwidth}
%   \centering
%   \includegraphics[width=.9\linewidth]{figures/FT_by_iteration/SM-by-it.pdf}
%   \caption{Schema Matching}
% \end{subfigure}
% \hspace*{\stretch{1}} % Margin on the right
          % \vspace{-6mm}
\caption{Quality of \sys by Iteration on Generative Tasks}
\label{figure:perf_by_it_generation}
\end{figure*}
}
{

}

\subsection{Main Results}
\label{sec:exp-quality}

\underline{Fine-tuning on GPT-3.5, GPT-4 and Llama-3.1-8B.}
Table~\ref{table:quality_comparison}  shows detailed comparisons on all table tasks and benchmarks between each vanilla base model, and its corresponding \sys model.

It can be seen that, for all generative and classification table tasks, \sys improves its base models (e.g., \sys-GPT-3.5 improves over GPT-3.5 on all benchmarks, 
% It is the best method in 24 out of 26 benchmark tests (ranking second in the remaining 2), 
and even surpassing vanilla GPT-4 on 7 benchmarks). Importantly, since we do not use the training split of any benchmark data during fine-tuning, it demonstrates that the fine-tuned models are capable of generalizing to multiple unseen benchmarks, as discussed in the introduction. %\sys is also better than other fine-tuning methods in 24 out of 26 benchmarks. 
% FT-no-validation ranks second in 13 out of 26 benchmarks, but lags behind \sys.

Additionally, Table~\ref{tbl:quality_comparison_tqa} shows that even on more open-ended generative tasks such as Table-QA, by pairing it with related generative tasks such as NL-to-code, \sys can still operate and improve the seemingly open-ended tasks.

%At the dataset level, \sys outperforms GPT-3.5 and FT-no-validation in all 20 tests and surpasses GPT-4 in 7 out of 20 tests. At the task level, \sys consistently outperforms GPT-3.5 and FT-no-validation across all 8 tasks and exceeds GPT-4 in 2 tasks.

\underline{Table-Specialist vs. Table-Generalist.}
 Table~\ref{tbl:quality_more} illustrates a comparison on all table tasks between \sys and Generalist Fine-Tuning models (shown at the task level, in the interest of space). \sys substantially outperforms both generalist models (TableLlama and TableGPT) on all tasks, showing the benefit of specialization over generalist models.

\iftoggle{full}
{
\underline{Iterative fine-tuning.}
Figure~\ref{figure:perf_by_it_generation} and~\ref{figure:perf_by_it_classification} show a comparison of \sys and vanilla GPT models (GPT-3.5 and GPT-4) on generative and classification tasks, respectively. The  x-axis here represents intermediate fine-tuned models, where ``V1'', ``V2'' represent models from the  first and second fine-tuning iteration, etc., and ``V0'' represents the base model. 

As we can see, \sys-GPT-3.5 demonstrates consistent quality improvement in consecutive iterations (trending up as we move right), matching or surpassing the GPT-4 level quality in some cases, which shows the benefit of iterative fine-tuning. In comparison, FT-no-validation is substantially less effective, showing the importance of validation.
}
{
\underline{Iterative fine-tuning.}
Figure~\ref{figure:perf_by_it_generation} shows an analysis of \sys by fine-tuning iterations. The x-axis here represents fine-tuned iterations, where ``V1'', ``V2'' represent models from the first and second iteration, etc., and ``V0'' represents the base model. 
As we can see, \sys demonstrates a consistent quality improvement in consecutive iterations (trending upward as we move to the right), matching or surpassing GPT-4-level quality in some cases. In comparison, FT-no-validation is substantially less effective.
}

% \underline{Fine-tuning on GPT-4 and  Llama3.1-8B.}
%     Table~\ref{table:quality_comparison} also shows the results when we fine-tune \sys models on GPT-4 and Llama31-8b, across all 8 tasks and 26 benchmark datasets. \sys fine-tuned on GPT-4 outperforms vanilla GPT-4 on 20 out of 26 benchmarks, while \sys fine-tuned on Llama3.1-8b consistently outperforms vanilla Llama3.1-8b on 25 out of 26 benchmarks, validating the effectiveness of \sys even on state-of-the-art frontier models. It is therefore possible to specialize models to perform even better than frontier models on individual table-tasks, at the same level of cost/latency.

% \begin{figure*}[h]
% \centering
% \hspace*{\stretch{1}} % Margin on the left
% \begin{subfigure}{.33\textwidth}
%   \centering
%   \includegraphics[width=\linewidth]{figures/FT_by_iteration/ED-by-it-V5.pdf}
%   \caption{Error Detection}
% \end{subfigure}\hfill
% \begin{subfigure}{.33\textwidth}
%   \centering
%   \includegraphics[width=\linewidth]{figures/FT_by_iteration/SM-by-it-V5.pdf}
%   \caption{Schema Matching}
% \end{subfigure}
% \hspace*{\stretch{1}} % Margin on the right
% \caption{Performance of \sys by Iteration on Classification Tasks}
% \label{figure:perf_by_it_classification}
% \end{figure*}

% \begin{figure}[ht!]
%     \centering
%     \includegraphics[width=0.75\linewidth]{figures/latency.pdf}
%     \caption{Average Latency for \sys (fine-tuned on GPT-3.5) and vanilla GPT-4}
%     \label{fig:latency}
% \end{figure}

\iftoggle{full}
{
\begin{figure*}
  \centering
  % Minipage for Figure 1
  \begin{minipage}[t]{0.64\textwidth}
    \centering
    % Subfigure for the first part of Figure 1
    \begin{subfigure}[t]{0.48\textwidth}
      \includegraphics[width=\linewidth]{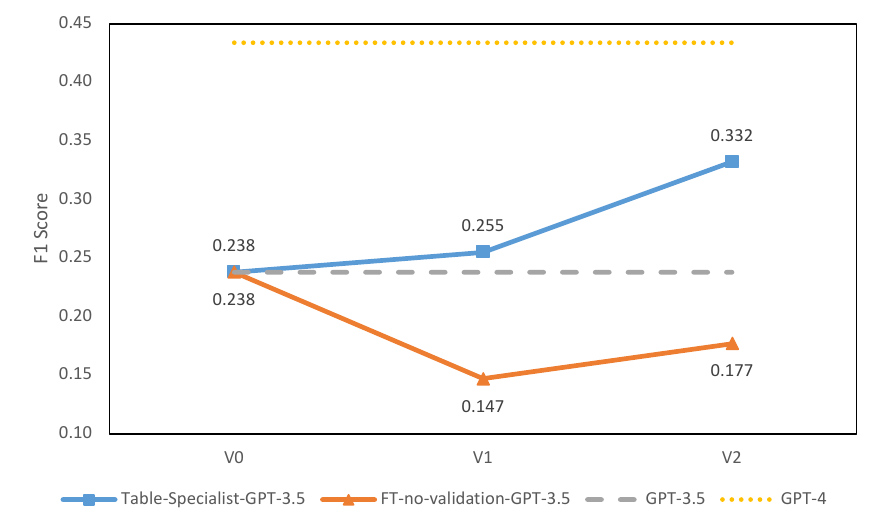}
      \caption{Error Detection}
      \label{fig:by_it_ed}
    \end{subfigure}
    \hfill
    % Subfigure for the second part of Figure 1
    \begin{subfigure}[t]{0.48\textwidth}
      \includegraphics[width=\linewidth]{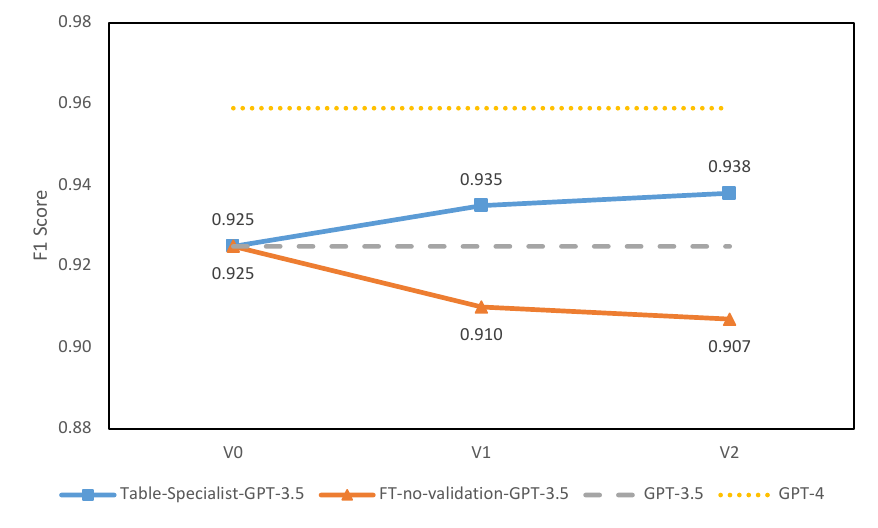}
      \caption{Schema Matching}
      \label{fig:by_it_sm}
    \end{subfigure}
    % Caption for Figure 1
    \caption{Quality of \sys by iteration on Classification Tasks}
    \label{figure:perf_by_it_classification}
  \end{minipage}
    \hspace{0.005\textwidth}
  \begin{tikzpicture}
    \draw [line width=0.5pt, dotted]
      (0,0) -- (0,\dimexpr\ht\strutbox+\dp\strutbox+80pt\relax);
  \end{tikzpicture}
  \hspace{0.005\textwidth}
  \hfill
  % Minipage for Figure 2
  \begin{minipage}[t]{0.32\textwidth}
    \centering
    \includegraphics[width=\linewidth]{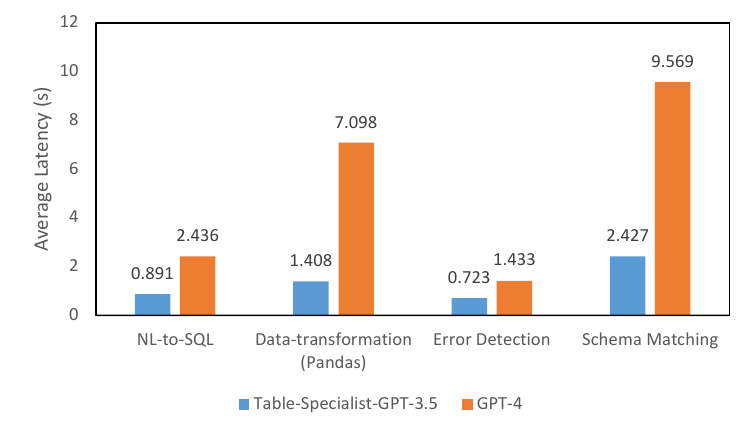}
    % Caption for Figure 2
    \captionof{figure}{Average Latency of \sys-GPT-3.5 vs. vanilla GPT-4}
    \label{fig:latency}
  \end{minipage}
\end{figure*}
}
{
   
}

%Each chart illustrates the performance of \sys and FT-no-validation across fine-tuning iterations. The baseline performance of vanilla GPT models is indicated by dotted horizontal lines. ``V0'' represents the base model (i.e. GPT-3.5), while denote the fine-tuned models at the first, second and third iterations, respectively. 

%The results demonstrate that \sys models consistently outperforms the base model and shows incremental gains with additional fine-tuning iterations. Notably, \sys even surpasses GPT-4 on several tasks. Conversely, FT-no-validation models sometimes underperform GPT-3.5, highlighting the importance of the training data validation process in the \sys fine-tuning approach.

\subsection{Latency and Cost Comparisons}
\label{sec:exp-cost}

% \begin{figure}[ht!]
%     \centering
%     \includegraphics[width=0.75\linewidth]{figures/latency.pdf}
%     \caption{Average Latency for \sys (fine-tuned on GPT-3.5) and vanilla GPT-4}
%     \label{fig:latency}
% \end{figure}

In Figure~\ref{fig:latency}, we compare the average latency of \sys models (fine-tuned on GPT-3.5) and GPT-4, on various task, averaged over all benchmark test cases. Because the fine-tuned \sys models are smaller, on average they are 3.42 times faster than vanilla GPT-4 (while still having comparable quality).  Figure~\ref{fig:quality-vs-latency} shows another analysis for NL-to-Code tasks, with similar latency reductions. Since serving online queries and ensuring interactivity is key in many  user-facing workloads, this highlights a crucial benefit of \sys as it allows us to employ smaller models to  reduce latency significantly.
%This demonstrates another advantage of applying fine-tuning instead of using more powerful models that incur longer latency.

In Figure~\ref{figure:cost_analysis}, we compare the cost\footnote{We use the published pricing~\cite{openai-pricing} to calculate the cost of fine-tuning and inference.} of fine-tuning and serving \sys using GPT-3.5, vs. serving directly using vanilla GPT-4, on two table tasks (results for other tasks are similar).  

The detailed unit price is listed in Table~\ref{tab:openai_price}. We  estimate the cost per API call using the average number of prompt and completion tokens for each tasks, as listed in Table~\ref{tab:inference_stats}.

\begin{small}
\begin{table}[ht!]
    \centering
    \caption{The Unit Price for Inference and Training, Per 1K Tokens, for Vanilla GPT-3.5, GPT-4, and Fine-tuned GPT-3.5. As of July 2, 2024~\cite{openai-pricing}}
    \begin{tabular}{|l||c|c|c|}
    \hline
        \textbf{Model}  & \textbf{Input} & \textbf{Output} & \textbf{Training} \\\hline
        GPT-3.5 & 0.001 & 0.002  & 0.008    \\
        FT(GPT-3.5) & 0.003 & 0.006  & N.A.    \\
        GPT-4 & 0.03  & 0.06 & N.A. \\\hline
    \end{tabular}
    \label{tab:openai_price}
\end{table}
\end{small}

\begin{small}
\begin{table}[ht!]
    \centering
    \caption{Average Number of Prompt and Completion Tokens, and Average Latency for \sys and GPT-4}
    \begin{tabular}{|c||c|c||c|c|}
    \hline
        \multirow{2}{*}{\textbf{Task} } & \multicolumn{2}{c||}{\textbf{Average \# of Tokens}} & \multicolumn{2}{c|}{\textbf{Average Latency (s)}} \\\cline{2-5}
                                         &   Prompt  & Completion & \sys & GPT-4 \\\hline
        NS & 969  & 38  & 0.891 & 2.436 \\
      R2RP & 678  & 92  & 1.408 & 7.098 \\
        ED & 701  & 10  & 0.723 & 1.433 \\
        SM & 1168 & 206 & 2.427 & 9.569 \\\hline
    \end{tabular}
    \label{tab:inference_stats}
\end{table}
\end{small}

%In addition to performance, we evaluated the cost-effectiveness of \sys compared to GPT-4 for long-term deployment. The cost analysis accounted for the total input, output and training tokens used in the generation, validation and fine-tuning processes.  %The estimated cost of serving \sys versus GPT-4 on NL-to-SQL(NS), Data Pandas (R2RP), Error Detection(ED) and Schema Matching(SM) is illustrated in .

We can see in Figure~\ref{figure:cost_analysis} that \sys-GPT-3.5 has to pay an upfront cost of fine-tuning, which is why it starts with a non-zero cost (on y-axis) to serve the first query (on x-axis). This however, is amortized over future queries, and takes less than 5000 queries (for Schema matching) or 10000 queries (for Data-transformation) for \sys to  break even with using vanilla GPT-4 directly. We argue that the cost saving in the long run, together with significant latency reductions, makes \sys a viable option, especially in user-facing online settings.

%This significant cost reduction, combined with comparable or superior performance, underscores the practical benefits of Table-Specialist for real-world applications.
% The result clearly demonstrates the cost-effectiveness of \sys fine-tuning compared to GPT-4 while maintaining comparable performance levels. These findings highlight the importance of considering both the short-term and long-term cost implications when choosing between fine-tuning specialized models and using generalist models like GPT-4. While \sys fine-tuning requires an initial investment in training, the ongoing operational savings make it a more cost-effective solution for large-scale applications.

\begin{small}
\begin{figure}
\centering
% \begin{subfigure}{.25\textwidth}
%   \centering
%   \includegraphics[width=\linewidth]{figures/Cost_Analysis/NS-Cost-V4.pdf}
%       \vspace{-6mm}
%   \caption{NL-to-SQL}
% \end{subfigure}\hfill

% \begin{subfigure}{.25\textwidth}
%   \centering
%   \includegraphics[width=\linewidth]{figures/Cost_Analysis/ED-Cost-V4.pdf}
%       \vspace{-6mm}
%   \caption{Error Detection}
%\end{subfigure}\hfill
\begin{subfigure}{.22\textwidth}
  \centering
  \includegraphics[width=\linewidth]{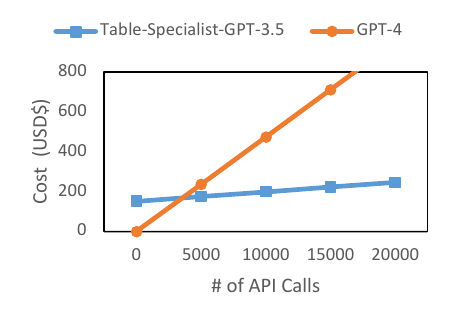}
      \vspace{-6mm}
  \caption{Schema Matching}
\end{subfigure}
\begin{subfigure}{.22\textwidth}
  \centering
  \includegraphics[width=\linewidth]{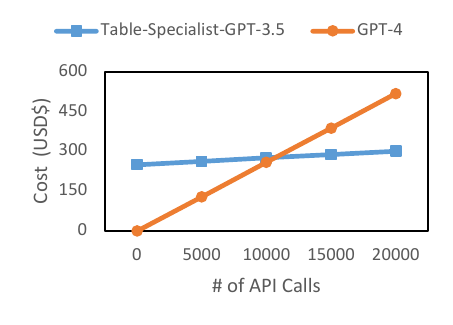}
      \vspace{-6mm}
  \caption{Data-transformation}
\end{subfigure}  %\hfill
    \vspace{-2mm}
\caption{Total Cost Analysis: \sys v.s. GPT-4}
\label{figure:cost_analysis}
\end{figure}
\end{small}

\subsection{Ablation Studies}

We conducted ablation studies to understand the impact of various components in \sys. 

\begin{table*}[th!]
    \caption{Ablation study: No ``textbook-like'' constrained task generation.}
    \label{tbl:ablation-no-textbook}
    \scalebox{0.8}
    {
    \begin{tabular}{c|cccccc}
    \toprule
        \textbf{Ablation}        & \textbf{NL-to-SQL}    & \textbf{NL-to-R}     & \textbf{NL-to-Scala}   & \textbf{Transform-Pandas}  & \textbf{Transform-R}  & \textbf{Transform-Scala} \\\midrule
           \sys           & 0.609 & 0.498  & 0.510 & 0.267 & 0.222 & 0.133 \\ 
         NoTextBook     & 0.601 & 0.497  & 0.501 & 0.257 & 0.200 & 0.135 \\
         %EasyHard       & 0.613 & 0.498  & 0.500 & 0.263 & 0.211 & 0.129 \\
         \toprule
    \end{tabular}
    }
\end{table*}

\begin{table}
    \caption{Ablation: No row/column permutation in validation.}
    \label{tbl:ablation1}
    \scalebox{0.65}
    {
    \begin{tabular}{c|ccccc}
    \toprule
        \textbf{Ablation}   & \textbf{Error-detect}    & \textbf{Schema-match}     & \textbf{NL-to-SQL}   & \textbf{NL-to-R}    & \textbf{NL-to-Scala}  \\\midrule
          \sys-GPT          & 0.255                    & 0.935                     & 0.594                & 0.496               & 0.466  \\ 
         No-Permutation     & 0.242                    & 0.912                     & 0.598                & 0.495               & 0.410  \\ \hline
         (Vanilla GPT-3.5)  & 0.238                    & 0.925                     & 0.569                & 0.370               & 0.270  \\     \toprule
    \end{tabular}
    }
\end{table}

\stitle{No permutation of tables.} Permutation-invariance is an important property we leverage to validate training data. 
In this ablation, we remove row sampling in NL-to-Code, and row/column shuffling in Error detection and Schema matching tasks.  Table~\ref{tbl:ablation1} shows that it leads to clear degradation in result quality. %, showing the importance of leveraging permutation-invariance in our table tasks.
%For this study, all row and column shuffling/sampling operations were excluded from the fine-tuning process in \sys. This includes the row sampling used in NL-to-Code pairwise validation (Section~\ref{sec:NL2Code}), and the row and column shuffling in self-consistent validation for schema matching (Section~\ref{sec:sm}). Results are presented in Table~\ref{tbl:ablation1} for the models fine-tuned in the first iteration.

\begin{table}
   % \vspace{-2mm}
    \caption{Ablation: No fine-tuning of Generators (and use vanilla GPT-3.5 as Generators instead), in classification tasks.}
        \vspace{-4mm}
    \label{tbl:ablation2}
    \scalebox{0.75}
    {
    \begin{tabular}{c|cc}
    \toprule
        \textbf{Ablation}       & \textbf{Error detection}    & \textbf{Schema matching}       \\\midrule
           \sys-GPT-3.5         & 0.332 & 0.938   \\ 
        No-Generator-fine-tune  & 0.310 & 0.931   \\ \hline
        (Vanilla GPT-3.5)       & 0.238 & 0.925   \\        
        \toprule
    \end{tabular}
    }
        \vspace{-2mm}
\end{table}

\stitle{No fine-tuning of Generator models.} In Table~\ref{tbl:ablation2}, we show the quality of not fine-tuning the Generator models in two classification tasks (the vanilla GPT-3.5 is used as the Generator in all iterations). We can see that this has a negative impact on result quality, suggesting that iterative fine-tuning of both Generator and Validator models are beneficial.

\begin{table}
    \caption{Ablation: No execution-based validation (and use language-models as validator instead), in generative tasks.}
    \label{tbl:ablation3}
    \scalebox{0.75}
    {
    \begin{tabular}{c|ccc}
    \toprule
        \textbf{Ablation}   & \textbf{NL-to-SQL}    & \textbf{NL-to-R} & \textbf{NL-to-Scala} \\\midrule
           \sys-GPT-3.5              & 0.594 & 0.496 & 0.466 \\ 
        No-execution-validation      & 0.595 & 0.489 & 0.457 \\ \hline
           (Vanilla GPT-3.5)         & 0.569 & 0.370 & 0.270 \\         \toprule
    \end{tabular}
    }
\end{table}

\stitle{No execution-based validator.} In Table~\ref{tbl:ablation3}, we show the result of not using execution-based validation (Algorithm~\ref{alg:validate-for-generative-task}), and vanilla model is used for validation instead,  for three generative NL-to-Code tasks.  We can see that 
``no-execution-validation'' still improves over vanilla GPT-3.5 (suggesting that they are still viable options), but generally has a negative impact on result quality when compared to execution-based validation in \sys-GPT-3.5.  %Note that in ``no-execution-validation'', language-models are used as validators instead, whose quality would , suggesting that language-models as validators are still viable options in our proposed \sys framework, even for generative tasks (even if they are not as performant as execution-based validation). 

\stitle{No ``textbook-like'' Constrained Task Generation.} We removed the ``textbook'' constrained task generation for the two generative tasks, and report resulting quality in Table~\ref{tbl:ablation-no-textbook}. We observe that having textbook-like curriculum-guided data generation improves the diversity of training data, and generally has a positive effect on the final model quality. % This study was conducted in two variants: \textbf{NoTextBook}, where no requirements were applied, and \textbf{EasyHard}, where task difficulty (easy, medium, hard) was specified for the table question generation (Equation~\ref{eq:t2q}) and transformed column generation (Equation~\ref{eq:t2dc}). 

\subsection{Sensitivity Analysis}
\label{sec:sensitivity}

We perform various types of sensitivity analysis in \sys.

% \begin{figure}[htbp]
%     \centering
%     \begin{minipage}[b]{0.23\textwidth}
%         \centering
%         \includegraphics[width=\linewidth]{figures/sensitivity/NS-NSS-VP-V4.pdf}
%         \caption{Vary Prompt Templates}
%         \label{fig:sensitivity-vary-prompt}
%     \end{minipage}
%     %\hfill
%     \begin{minipage}[b]{0.23\textwidth}
%         \centering
%         \includegraphics[width=\linewidth]{figures/sensitivity/NS-NSS-VarySize-V2.pdf}
%         \caption{Vary Training Data Size}
%         \label{fig:sensitivity-amount-training-data}
%     \end{minipage}    
% \end{figure}

\begin{figure}[h]
    %\hfill
    \begin{subfigure}[b]{0.22\textwidth}
        \centering
        \includegraphics[width=\linewidth]{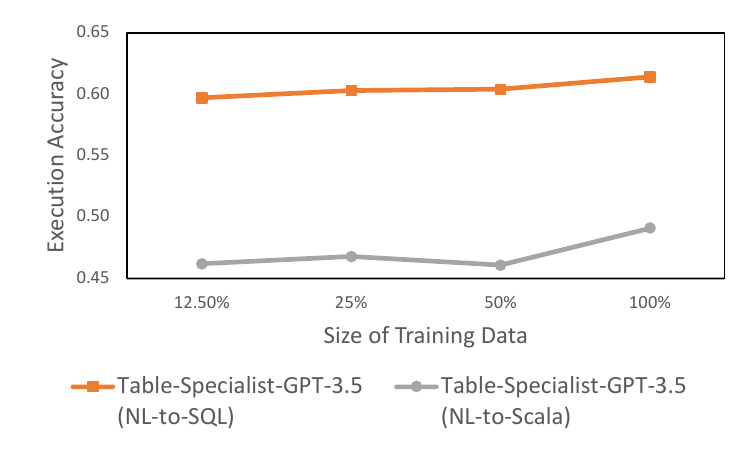}
        \vspace{-5mm}
        \caption{Vary Training Size}
        \label{fig:sensitivity-amount-training-data}
    \end{subfigure}  
    \centering
    \begin{subfigure}[b]{0.22\textwidth}
        \centering
        \includegraphics[width=\linewidth]{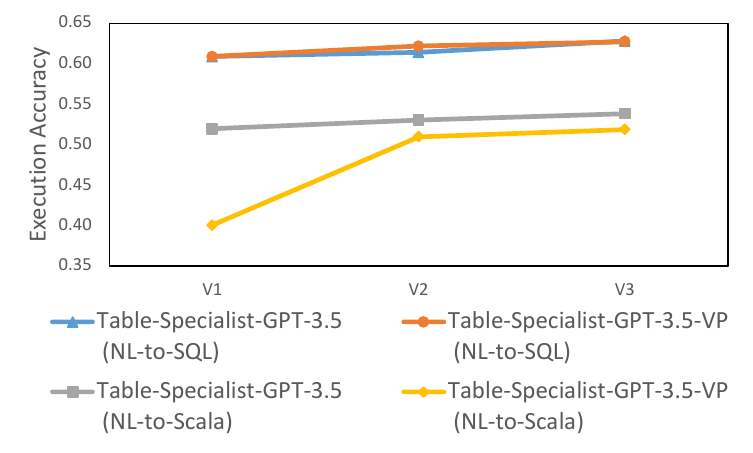}
        \vspace{-5mm}
        \caption{Vary Prompt Templates}
        \label{fig:sensitivity-vary-prompt}
    \end{subfigure}
    \vspace{-3mm}
    \caption{Sensitivity analysis}
    \vspace{-3mm}
    
\end{figure}

% \begin{small}
% \begin{figure*}[h]
% \centering
% \begin{subfigure}{.3\textwidth}
%   \centering
%   \includegraphics[width=\linewidth]{figures/sensitivity/NS-NSS-VP-V2.pdf}
%   \caption{NL-to-SQL}
%   \label{fig:sensitivity-vary-prompt}
% \end{subfigure}\hfill
% \begin{subfigure}{.23\textwidth}
%   \centering
%   \includegraphics[width=\linewidth]{figures/sensitivity/NS-NSS-VarySize.pdf}
%   \caption{Data-transformation Pandas}
%   \label{fig:sensitivity-amount-training-data}
% \end{subfigure}\hfill
% \begin{subfigure}{.23\textwidth}
%   \centering
%   \includegraphics[width=\linewidth]{figures/sensitivity/NS-VaryBase.pdf}
%   \caption{Error Detection}
% \end{subfigure}\hfill
% \begin{subfigure}{.23\textwidth}
%   \centering
%   \includegraphics[width=\linewidth]{figures/sensitivity/NSS-VaryBase.pdf}
%   \caption{Schema Matching}
% \end{subfigure}

% \caption{Cost Analysis: \sys v.s. GPT-4 Total Cost by Number of API Calls}
% \label{figure:cost_analysis}
% \end{figure*}
% \end{small}

% (1) prompt variation x5 (paraphrased prompt), for x1 task (NL-to-Code?)

% \begin{figure}[h!]
%     \centering
%     \includegraphics[width=.9\linewidth]{figures/sensitivity/NS-NSS-VP.pdf}
%     \caption{Caption}
%     \label{fig:sensitivity-vary-prompt}
% \end{figure}

% (2)vary amount of training data, for x1 task (NL-to-Code?)

\stitle{Vary the Amount of Training Data.} Figure~\ref{fig:sensitivity-amount-training-data} shows \sys quality, when we vary the amount of training data produced by the Generator from 100\% to 50\%, 25\% and 12.5\% of the original data (x-axis). %where we reduce the amount of training data produced by the Generator in each table task.   
We can see that increasing the amount of training data generally has a positive effect on result quality.
%It is evident that increasing the amount of training data consistently enhances performance across both NS and NSS tasks.

\stitle{Vary Prompt Templates.} To test the robustness of \sys, we vary our prompt templates used in each task, by giving our original prompt to ChatGPT and asking it to paraphrase into five different prompts, for the NL-to-SQL and NL-to-Scala tasks.  Figure~\ref{fig:sensitivity-vary-prompt} shows that using variants of the prompt (abbreviated as VP in the figure), lead to comparable quality.  %(especially in the final V3 iteration)
%To evaluate the impact of varying prompt templates on \sys fine-tuning, we created five distinct prompt variants (paraphrased using ChatGPT) and fine-tuned the \sys models on the NS and NSS tasks with the identical training data. The models trained with varied prompts were evaluated using the same variants. As shown in Figure~\ref{fig:sensitivity-vary-prompt}, the performance on NS remained largely stable, while there was a slight decline when using varied prompt templates on NSS.

% \begin{figure}[h!]
%     \centering
%     \includegraphics[width=.9\linewidth]{figures/sensitivity/Amount-Training-data.pdf}
%     \caption{Caption}
%     \label{fig:sensitivity-amount-training-data}
% \end{figure}

\iftoggle{full}
{
    % (3)base-vanilla vs. continous-FT (NL-to-Code)
    \stitle{Vary the Base Model for Fine-Tuning.} We test two alternatives of iterative fine-tuning, where in each iteration, we initialize the base model either as the vanilla model (e.g., GPT-3.5), or the model from the last iteration (continuous fine-tune). 
    %\sys model during the fine-tuning process: starting with a vanilla model (Base Vanilla) and using a model fine-tuned in a previous iteration (Continuous Fine-Tuning). 
    Figure~\ref{fig:vary_base} shows that using vanilla GPT as the base models are consistently better than using the check-point from the last iteration (continuous fine-tune), for both NL-to-SQL and NL-to-Scala.% the performance across iterations on NS and NSS for both strategies. The result reveals that models fine-tuned from vanilla consistently outperforms those from continuous fine-tuning on both tasks. 

    \begin{figure}[htbp]
        \centering
        \hfill
        \begin{minipage}[b]{0.48\textwidth}
            \begin{subfigure}{0.5\textwidth}
                \centering
                \includegraphics[width=\linewidth]{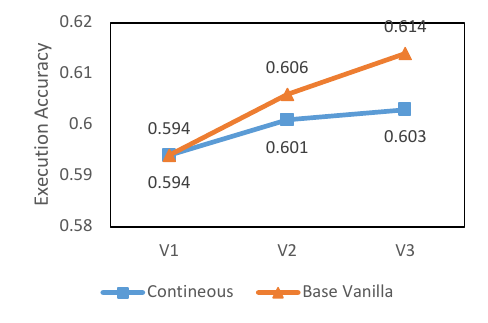}
                \caption{NL-to-SQL}
                \label{fig:ns_vary_base}
            \end{subfigure}
            \hfill
            \begin{subfigure}{0.5\textwidth}
                \centering
                \includegraphics[width=\linewidth]{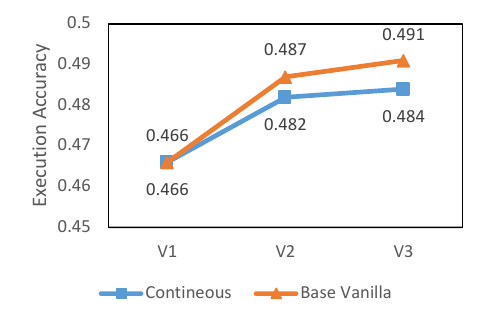}
                \caption{NL-to-Scala}
                \label{fig:nss_vary_base}
            \end{subfigure}
            \caption{Vary Base Model}
            \label{fig:vary_base}
        \end{minipage}
    \end{figure}
}

\section{Conclusions and Future Work}
In this work, we develop a new fine-tuning approach \sys specifically designed for table-tasks. We show that it can fine-tune small models specialized for individual table tasks, while still being performant and generalizable. Future directions include testing the method on additional base models, and tasks beyond table tasks.  
% In this work, we propose a new paradigm called \sys fine-tuning, to fine-tune large-large models like GPT-3.5 and GPT-4 for specific table tasks with self-generated and validated training data. We utilize no existing benchmark data for any guidance. Our experiment show that \sys fine-tuned models exhibit improved performance across different benchmarks for the same table task. 

% ...

%This work introduces \sys, an innovative fine-tuning framework for language models focused on enhancing their performance on specific table tasks. Unlike previous approaches that either required extensive human-labeled datasets or generic fine-tuning that did not cater to specific tasks, \sys leverages a self-training method that iteratively fine-tunes the dual models for generative and classification table tasks. This approach significantly reduces the dependency on human-labeled data and mitigates overfitting by utilizing synthetically generated and validated training data. Our experiments have demonstrated that \sys models not only outperform baseline models, including high-performing ones like GPT-4, but also offer reduced inference costs and improved latency, making them particularly suitable for real-world applications where efficiency and cost-effectiveness are critical.

% Looking ahead, several directions appear promising for extending the capabilities of the \sys framework: (1) Expansion to Multimodal Inputs

\iftoggle{full}
{
    % removed for revision
    % \revised{}
    \clearpage
    \appendix
    \onecolumn
\section{Sample Prompts}
\label{apx:tasks}

% NL2Code
\begin{lstlisting}[title=Table Question Generation]
Task description: Given the input table below, your task is to generate questions that can be answered by the given table with a SQL query.

Input:
Table:
|Rank|Country|2016|2013|2012|2011|
|---|---|---|---|---|---|
|1|Qatar|0.08%|0.11%|0.10%|0.02%|
|2|Malta|0.60%|0.61%|0.61%|0.72%|
|3|Saudi Arabia|1.14%|1.32%|1.31%|1.26%|
|4|Barbados|1.32%|1.16%|1.15%|2.44%|
|5|Grenada|1.42%|1.44%|1.46%|2.29%|
|6|Iceland|1.52%|1.55%|1.53%|1.56%|
|7|Bahrain|1.69%|1.81%|1.81%|1.66%|
|8|Kiribati|1.78%|1.78%|1.78%|1.88%|
|9|United Arab Emirates|1.97%|2.10%|2.07%|4.09%|
|10|Sweden|2.12%|2.26%|2.15%|2.00%|
|11|Norway|2.19%|2.35%|2.31%|2.28%|
|12|Finland|2.21%|2.28%|2.24%|2.06%|
|13|Singapore|2.27%|2.49%|2.54%|2.85%|
|14|Egypt|2.29%|2.34%|2.33%|2.38%|
|15|Israel|2.30%|2.49%|2.43%|2.60%|
|16|Estonia|2.36%|2.52%|2.50%|2.25%|
|17|Switzerland|2.37%|2.61%|2.59%|2.55%|
|18|Luxembourg|2.43%|2.68%|2.65%|2.70%|
|19|Seychelles|2.55%|2.58%|2.60%|2.68%|
|20|France|2.62%|2.79%|2.78%|2.76%|
|21|Oman|2.64%|2.74%|2.72%|2.80%|
|22|Cyprus|2.68%|2.77%|2.81%|3.46%|
|23|Denmark|2.89%|3.10%|3.09%|2.86%|
|24|Lithuania|2.92%|3.18%|3.23%|2.89%|
|25|Germany|2.95%|3.24%|3.27%|2.96%|
|26|Ukraine|2.97%|3.14%|3.19%|3.02%|
|27|Canada|3.01%|3.18%|3.18%|2.57%|
|28|Spain|3.05%|3.38%|3.40%|3.29%|
|29|Belgium|3.07%|3.42%|3.48%|3.51%|
|30|Mongolia|3.08%|3.10%|3.24%|3.43%|
|31|Belarus|3.11%|3.31%|3.32%|2.98%|
|32|Poland|3.20%|3.46%|3.53%|3.42%|
|33|Kuwait|3.28%|3.70%|3.71%|3.71%|
|34|Latvia|3.31%|3.48%|3.51%|3.09%|
|35|Czech Republic|3.37%|3.61%|3.67%|4.15%|
|36|Slovakia|3.39%|3.63%|3.69%|3.38%|
|37|Austria|3.39%|3.80%|3.75%|3.41%|
|38|Slovenia|3.41%|3.69%|3.81%|3.72%|
|39|Portugal|3.45%|3.80%|3.82%|3.62%|
|40|Paraguay|3.48%|3.85%|3.84%|4.12%|
|41|United Kingdom|3.54%|3.71%|3.65%|3.61%|
|42|Kazakhstan|3.56%|3.84%|3.87%|4.04%|
|43|Argentina|3.56%|3.76%|3.80%|3.77%|
|44|Russia|3.58%|3.78%|3.83%|3.56%|
|45|United States|3.76%|3.99%|3.99%|3.72%|
|46|Libya|3.79%|3.93%|3.80%|3.67%|
|47|Croatia|3.97%|4.24%|4.35%|3.71%|
|48|Uruguay|4.03%|4.09%|4.12%|3.94%|
|49|Brazil|4.09%|4.26%|4.30%|4.26%|
|50|Bahamas|4.14%|3.99%|4.17%|4.52%|


Please return one natural language question, and ensure that when your question is answered in SQL, the corresponding SQL would contain all of the below:
(1) *ONE (1)* number preprocessing step (e.g., extracting numbers from a column for comparisons/sorting)
(2) *ONE (1)* sorting then take top-K step

Return your answer in the following JSON format: {"question": <QUESTION>}.

Output:

{"question": "What are the top 5 countries with the lowest percentage of GDP in 2016?"}
\end{lstlisting}
\begin{lstlisting}[title=NL-to-SQL (Generative)]
# Task description:
Natural language to SQL, also known as NL2SQL, is the problem of generating a SQL query that can be executed on a input table, for a given natural language question.

You will be given an input table called "table" in the markdown format, as well as a question. Your task is to generate an SQL query that can be executed on the input table to answer the question.

Please ensure that the generated SQL only return relevant columns being specifically asked in the question (e.g., in SQL please don't return more than one columns in SELECT if only one column is needed to answer the given question, and similarly please refrain from using SELECT * unless the entire row is needed to answer the question).
 
When generating SQL code, please use the SQLite syntax to ensure that your SQL is executable on SQLite3. Use only columns that are present in the table, always use quotes around your column and table names (some column names have punctuation or special characters), and refer to the columns using the column-names exactly as they appear in the table, do not add or remove punctuation (underscore, slash, etc.)

No explanation, return the code only with the following markdown codeblock format : ```sql<SQL CODE>```.

For example:
 
Input:
Table:
|City|Population (2015)|Area|Density|City class|Income class|Province|
|---|---|---|---|---|---|---|
|Bacolod|561,875|62.81|9,100|Highly urbanized|1st|Negros Occidental|
|Bago|170,981|154.9|1,100|Component|2nd|Negros Occidental|
|Bais|76,291|123.41|620|Component|3rd|Negros Oriental|
|Bayawan|117,900|269.92|440|Component|2nd|Negros Oriental|
|Cadiz|154,723|202.54|750|Component|2nd|Negros Occidental|
|Canlaon|54,509|66.0|830|Component|4th|Negros Oriental|
|Dumaguete|131,377|12.98|10,000|Component|2nd|Negros Oriental|
|Escalante|94,070|74.43|1,300|Component|4th|Negros Occidental|
|Guihulngan|95,969|150.02|650|Component|5th|Negros Oriental|
|Himamaylan|106,880|141.71|750|Component|3rd|Negros Occidental|
|Kabankalan|181,977|269.25|670|Component|1st|Negros Occidental|
|La Carlota|64,469|53.01|1,200|Component|4th|Negros Occidental|
|Sagay|146,264|127.54|1,100|Component|3rd|Negros Occidental|
|San Carlos|132,536|174.33|750|Component|2nd|Negros Occidental|
|Silay|126,930|82.93|1,500|Component|3rd|Negros Occidental|
|Sipalay|70,070|146.63|470|Component|4th|Negros Occidental|
|Talisay|102,214|77.68|1,300|Component|4th|Negros Occidental|
|Tanjay|80,532|106.58|750|Component|4th|Negros Oriental|
|Victorias|87,933|51.71|1,700|Component|4th|Negros Occidental|

question: What is the top-3 most populous city in 2015 in the table?

Output:
```sql
SELECT "City"
FROM "table"
ORDER BY CAST(REPLACE("Population (2015)", ',', '') AS INTEGER) DESC
LIMIT 3;
```
 
Input:
Table:
|Contestant ID|Finalist|Gender|
|---|---|---|
|SSJ01|Monisha|Female|
|SSJ02|Alka Ajith|Female|
|SSJ03|Balasarangan|Male|
|SSJ04|Sahana|Female|
|SSJ05|Manisha|Female|
|SSJ06|Prasanna Sundar|Male|
|SSJ07|Sowmya|Female|
|SSJ08|Oviya|Female|
|SSJ09|Vishnucharan|Male|
|SSJ10|Sivaranjani|Female|
|SSJ11|Nithyashree|Female|
|SSJ12|Harish|Male|
|SSJ13|Shri Prasanna|Male|
|SSJ14|Sanjana|Female|
|SSJ15|Shravan|Male|
|SSJ16|Srinisha|Female|
|SSJ17|Shrihari|Male|
|SSJ18|Sharath|Male|
|SSJ19|Roshan|Male|
|SSJ20|Srikanth|Male|
|SSJ21|Priyanka|Female|
|SSJ22|Soundarya|Female|
|SSJ23|Sathyanarayanan|Male|
|SSJ24|Srinidhi|Female|
|SSJ25|Ponmozhi|Female|

question: How many female contestants are there in total?
 
Output: 
```sql
SELECT COUNT("Contestant ID")
FROM "table"
WHERE "Gender" = 'Female';
```
 
Input:
Table:
|Club|Coach|Replaced Coach|Home stadium|
|---|---|---|---|
|FC Dynamo Kyiv|Yozhef Sabo|Mykhailo Fomenko|Republican Stadium|
|FC Shakhtar Donetsk|Valery Yaremchenko||Shakhtar Stadium|
|FC Chornomorets Odessa|Viktor Prokopenko||Black Sea Shipping Stadium|
|FC Dnipro Dnipropetrovsk|Mykola Pavlov||Meteor Stadium|
|FC Karpaty Lviv|Myron Markevych||Ukraina Stadium|
|FC Kryvbas Kryvyi Rih|Volodymyr Brukhtiy|Ihor Nadein|Metalurh Stadium|
|FC Nyva Ternopil|Leonid Buriak||City Stadium|
|SC Tavriya Simferopol|Anatoliy Zayaev||Lokomotyv Stadium|
|FC Temp Shepetivka|Leonid Tkachenko||Temp Stadium|
|FC Nyva Vinnytsia|Yukhym Shkolnykov||Central City Stadium|
|FC Volyn Lutsk|Roman Pokora||Avanhard Stadium|
|FC Veres Rivne|Vyacheslav Kobyletskyi|Mykhailo Dunets|Avanhard Stadium|
|FC Torpedo Zaporizhia|Ihor Nadein|Viktor Matviyenko|AvtoZAZ Stadium|
|FC Zorya-MALS|Volodymyr Kobzarev|Anatoliy Shakun|Avanhard Stadium|
|FC Kremin Kremenchuk|Evhen Rudakov|Boris Streltsov 8 games Tiberiy Korponay 9 games|Dnipro Stadium|
|FC Metalurh Zaporizhia|Anatoliy Kuksov|Janis Skredelis Hryhoriy Vul|Metalurh Stadium|
|FC Bukovyna Chernivtsi|Oleksandr Pavlenko||Bukovyna Stadium|
|FC Metalist Kharkiv|Viktor Kamarzayev|Oleksandr Dovbiy Yevhen Lemeshko|Metalist Stadium|

question: What is the name of the home statium of the football club FC Kryvbas Kryvyi Rih?"
 
Output: 
```sql
SELECT "Home stadium"
FROM "table"
WHERE "Club" = 'FC Kryvbas Kryvyi Rih';
```
 
Input:
Table:
|Title|Year|Director(s)|
|---|---|---|
|"Red Morning Light"|2003|Douglas Biro|
|"Molly's Chambers"|2003|Honey|
|"Wasted Time"|2003|Mark Pellington|
|"The Bucket"|2004|Patrick Daughters|
|"Four Kicks"|2005|Patrick Daughters|
|"King of the Rodeo"|2005|Patrick Daughters|
|"On Call"|2007|Adria Petty|
|"Charmer"|2007|Robert Hales|
|"Crawl"|2008|Nick Wickham|
|"Sex on Fire"|2008|Sophie Muller|
|"Use Somebody"|2008|Sophie Muller|
|"Notion"|2009|Phil Griffin|
|"Radioactive"|2010|Sophie Muller|
|"Pyro"|2010|Max Goldman, Paul Greenhouse, Casey McGrath|
|"Back Down South"|2011|Casey McGrath|
|"Supersoaker"|2013|W.I.Z.|
|"Beautiful War"|2013|Casey McGrath|
|"Temple"|2014|Casey McGrath|
|"Waste a Moment"|2016|Dimitri Basil|
|"Walls"|2016|Dimitri Basil|
|"Around the World"|2016||

question: What are the names of the directors that have directed movies since 2010, and how many movies have they produced?
 
Output: 
```sql
SELECT "Director(s)", COUNT("Title")
FROM "table"
WHERE "Year" >= 2010
GROUP BY "Director(s)";
```

Now, for the input table and question below,  generate its corresponding SQL query for the natural language question. No explanation, return the code only with the following markdown codeblock format : ```sql<SQL CODE>```.

Input:
Table:
|Date from|Nationality|Name|To|Fee|Ref.|
|---|---|---|---|---|---|
|1 July 2017||Fabian Bailey|Free agent|Released||
|1 July 2017||Lee Camp|Free agent|Released||
|1 July 2017||Chris Dawson|Free agent|Released||
|1 July 2017||Joel Ekstrand|Free agent|Released||
|1 July 2017||Stephen Kelly|Free agent|Released||
|1 July 2017||Tom Rose|Free agent|Released||
|1 July 2017||Richard Smallwood|Free agent|Released||
|1 July 2017||Tom Thorpe|Free agent|Released||
|1 July 2017||Danny Ward|Cardiff City|Undisclosed||
|1 July 2017||Kelvin Wilson|Free agent|Released||
|4 July 2017||Kirk Broadfoot|Kilmarnock|Undisclosed||
|12 July 2017||Dexter Blackstock|Free Agent|Released||


question: Which players were released as free agents on July 1, 2017?

Output:

```sql
SELECT "Name"
FROM "table"
WHERE "Date from" = '1 July 2017' AND "Fee" = 'Free agent';
```
\end{lstlisting}
\begin{lstlisting}[title=NL-to-SQL (Classification)]
Task description:
Natural language to SQL, also known as NL2SQL, is the problem of generating a SQL query that can be executed on a input table, for a given natural language question. You will be given an input table called "table" in the markdown format below, a question in natural language for this table, and a corresponding SQL query for the natural language question. You need to: (1) determine if the SQL query is grammarly correct, (2) determine if the SQL query can truthfully represent the natural language question being asked.  Return your answer in the following JSON format: {"grammarly_correct": <YES or NO>, "SQL-correct": <YES or NO>}
 
Table:
|Title|Episode #|Writers|Directors|Summary|
|---|---|---|---|---|
|"That's What Friends Are For"|125|Elaine Overbey|Michael Dimich|Robin tries to help her friend Becky (played by Natasha Pearce) overcome her shyness. Songs: "Open Your Eyes", "Big Girls Don't Cry", "Games", "Hold On", "When I See You Smile" *Note*: The episode's title is a reference to the song That's What Friends Are For.|
|"P*lace Alone"|108|||Attempting to prove they're "grown up"; Haylie and Robin try to protect the P*lace from a burglar. Songs: "Get Up! (Before the Night Is Over)", "Release Me", "I Can Do Anything", "The Power", "Save Me"|
|"Secret Agent Girl"|126|||Haylie gets a spy kit, but when she mistakes a comment the others made about getting rid of a doll and thinks they're talking about her, she threatens to quit. Songs: "Get Ready", "Impulsive", "Read My Mind", "Lies", "More Than Ever" Note: This episode is sometimes called "The Spy Kit." "Secret Agent Girl", the episode's title is a parody of the song Secret Agent Man.|
|"Breaking Up is Hard to Do"|115|Ken Lipman|David Grossman|Ana's father (played by Barry Williams) comes to town on business; and Ana feels neglected when he doesn't spend much time with her. Songs: "Pump Up the Jam", "This Old Heart of Mine (Is Weak for You)", "If You Asked Me To", "Real Love", "Don't Ever Go"|
|"Music Lessons"|117|||Kenny attempts to cope with the sudden death of his music teacher. Songs: "Miss You Much", "Dancing Machine", "Power of Love/Love Power", "How Can I Go On", "Tomorrow"|
|"Earth Day Festival"|124|||The kids hold a special festival for Earth Day. Songs: "After the Rain", "Yakety Yak", "Please Save Us the World", "This is the Right Time", "Save the Trees" Note: Please Save Us the World would later become the only Kids Incorporated original to appear on a cast member's album when it appeared in Jennifer Love Hewitt's debut album Love Songs Note: The three original songs are the most to appear in a single episode.|
|"Family Matters"|113|||Ana's attempts to outdo Robin causes a rift between the two cousins. Songs: "Jealous Again", "You Can't Deny It", "Don't Treat Me Bad", "If You Don't Know Me By Now", "You're the One"|
|"My Fair Ana"|120|||Ana develops a crush on a boy she has math class with. Songs: "The Girl I Used to Know", "The Great Pretender", "You Need a New Attitude", "Individuality", "Waiting for Love"|
|"Pipe Dreams"|111|||The kids get excited about an upcoming concert until Haylie accidentally loses the tickets. Songs: "U Can't Touch This", "How Can We Be Friends", "Give a Little Love", "Follow it Around", "Voices That Care"|
|"Teen Spotlight"|109|||Kids Incorporated is featured in a segment on a teen-oriented show. Songs: "I'm Breaking Free", "Do You Love Me", "If Wishes Came True", "Friends", "I'll Be Good to You"|
|"History in the Making"|112|||Eric is having trouble in his history class. Songs: "Don't You Want to Be Mine", "Mercy Mercy Me (The Ecology)", "Two to Make It Right", "History", "I'll Be Your Shelter"|
|"Five Kids and a Dog"|121|||Ana takes in a stray dog. Guest appearance by Paul Benvenuti and Foster (as the dog). Songs: "We'll Be Rockin'", "Cherish", "Walkin' the Dog", "Everything", "Downtown Train"|
|"Flip Out"|116|Elaine Overbey|Michael Dimich|When Flip's sneaker fortune runs out; a financial advisor (played by guest star Karla DeVito) suggests making the P*lace a tea room for adults. Songs: "Club at the End of the Street", "(Can't Live Without Your) Love and Affection", "Cry for Help", "Fairweather Friend", "This House"|
|"New Twist"|114|Elaine Overbey|Michael Dimich|The annual tongue-twister contest is approaching at Robin's school; and her classmates name her the school's representative. Songs: "Can't Stop", "Hold On", "I Can Fly", "Get On Your Feet", "Forever Young"|
|"Double Trouble"|123|||The kids meet a fun-loving kid new in town and his twin brother. Scott Wolf (credited as Scott Tyler Wolf) guest stars as Billy and Bobby. Songs: "Feels Good", "Lollipop", "How Can We Be Friends", "All Around the World", "Swear to Your Heart"|
|"While the Cat's Away"|118|||An urgent meeting forces Flip to leave Kenny and Eric in charge. Songs: "Step by Step", "Let the Good Times Roll", "Here We Go (Let's Rock & Roll)", "All Play and No Work", "Rhythm of My Heart"|
|"Thirteensomething"|110|||A girl named Cynthia visits; and Kenny attempts to impress her. Songs: "The Book of Love", "Another Day in Paradise", "Don't Wanna Lose You", "Something Happened on the Way to Heaven", "High Enough"|
|"A Hard Date's Night"|107|||Stacy, Devyn, and Richie have left the band. They are replaced by Haylie, Ana, and Eric. Eric, noticing how some girls are paying attention to him, decides to date three of them at once. Songs: "Romeo", "The Shoop Shoop Song (It's in His Kiss)", "It's the Right Time", "Heat of the Moment", "King of Wishful Thinking" Note: The episode's title is a parody of The Beatles song A Hard Day's Night (song)|
|"Mummy Dearest"|122|||A mix-up in a delivery leads to a mummy being sent to the P*lace. Songs: "Where There's a Will", "Boris the Spider", "Coming Out of the Dark", "Call It Superstition", "If I Could Turn Back Time"|
|"Tall Order"|119|||Haylie becomes self-conscious about being the shortest member of the band. Guest appearance by Mandy Carr and Randy Lathrop. Songs: "Gonna Make You Sweat (Everybody Dance Now)", "Little Darlin'", "When I'm Back on My Feet Again", "Wild Child", "Follow Your Dream"|


Question:
Which episodes were directed by Michael Dimich and had the word 'Girl' in the title?

SQL:
SELECT "Title"
FROM "table"
WHERE "Directors" = 'Michael Dimich' AND "Title" LIKE '%Girl%';

Please determine if the SQL query is grammarly correct and if it can truthfully represent the natural language question being asked. Return your answer in the following JSON format: {"grammarly_correct": <YES or NO>, "SQL-correct": <YES or NO>}.

Output:

{"grammarly_correct": "YES", "SQL-correct": "YES"}
\end{lstlisting}
\begin{lstlisting}[title=NL-to-R (Generative)]
# Task description:
Natural language to R script, is a problem of generating R script  that can be executed on a input CSV table, for a given natural language question.

You will be given an input table called "table" in the markdown format, as well as a question. Your task is to generate a R script that can be executed on the input table to answer the question.

Please ensure that the generated R script only return relevant columns being specifically asked in the question.

When generating R script, you can assume that the input table is always stored in a file named "data.csv", so always start your R script with the boilerplate prefix:
"library(dplyr)

data <- read.csv("data.csv", check.names = FALSE)
"

After you generate R script code to produce results, please always save your final results into a file named "result.csv" with no index column, so your R script should always end with:
"write.csv(final_result, "result.csv", row.names = FALSE)"

No explanation, return the code only with the following markdown codeblock format : ```r<R CODE>```.

For example:
 
Input:
Table:
|City|Population (2015)|Area|Density|City class|Income class|Province|
|---|---|---|---|---|---|---|
|Bacolod|561,875|62.81|9,100|Highly urbanized|1st|Negros Occidental|
|Bago|170,981|154.9|1,100|Component|2nd|Negros Occidental|
|Bais|76,291|123.41|620|Component|3rd|Negros Oriental|
|Bayawan|117,900|269.92|440|Component|2nd|Negros Oriental|
|Cadiz|154,723|202.54|750|Component|2nd|Negros Occidental|
|Canlaon|54,509|66.0|830|Component|4th|Negros Oriental|
|Dumaguete|131,377|12.98|10,000|Component|2nd|Negros Oriental|
|Escalante|94,070|74.43|1,300|Component|4th|Negros Occidental|
|Guihulngan|95,969|150.02|650|Component|5th|Negros Oriental|
|Himamaylan|106,880|141.71|750|Component|3rd|Negros Occidental|
|Kabankalan|181,977|269.25|670|Component|1st|Negros Occidental|
|La Carlota|64,469|53.01|1,200|Component|4th|Negros Occidental|
|Sagay|146,264|127.54|1,100|Component|3rd|Negros Occidental|
|San Carlos|132,536|174.33|750|Component|2nd|Negros Occidental|
|Silay|126,930|82.93|1,500|Component|3rd|Negros Occidental|
|Sipalay|70,070|146.63|470|Component|4th|Negros Occidental|
|Talisay|102,214|77.68|1,300|Component|4th|Negros Occidental|
|Tanjay|80,532|106.58|750|Component|4th|Negros Oriental|
|Victorias|87,933|51.71|1,700|Component|4th|Negros Occidental|

question: What is the top-3 most populous city in 2015 in the table?

Output: 
```r
library(dplyr)

data <- read.csv("data.csv", check.names = FALSE)

data$`Population (2015)` <- as.numeric(gsub(",", "", data$`Population (2015)`))

top_3_cities <- data %>%
  arrange(desc(`Population (2015)`)) %>%
  slice(1:3)

final_result <- top_3_cities %>%
  select(City)

write.csv(final_result, "result.csv", row.names = FALSE)
```
 
Input:
Table:
|Contestant ID|Finalist|Gender|
|---|---|---|
|SSJ01|Monisha|Female|
|SSJ02|Alka Ajith|Female|
|SSJ03|Balasarangan|Male|
|SSJ04|Sahana|Female|
|SSJ05|Manisha|Female|
|SSJ06|Prasanna Sundar|Male|
|SSJ07|Sowmya|Female|
|SSJ08|Oviya|Female|
|SSJ09|Vishnucharan|Male|
|SSJ10|Sivaranjani|Female|
|SSJ11|Nithyashree|Female|
|SSJ12|Harish|Male|
|SSJ13|Shri Prasanna|Male|
|SSJ14|Sanjana|Female|
|SSJ15|Shravan|Male|
|SSJ16|Srinisha|Female|
|SSJ17|Shrihari|Male|
|SSJ18|Sharath|Male|
|SSJ19|Roshan|Male|
|SSJ20|Srikanth|Male|
|SSJ21|Priyanka|Female|
|SSJ22|Soundarya|Female|
|SSJ23|Sathyanarayanan|Male|
|SSJ24|Srinidhi|Female|
|SSJ25|Ponmozhi|Female|

question: How many female contestants are there in total?
 
Output: 
```r
library(dplyr)

data <- read.csv("data.csv", check.names = FALSE)

female_count <- data %>%
  filter(Gender == "Female") %>%
  summarise(Count = n())

write.csv(female_count, "result.csv", row.names = FALSE)
```
 
Input:
Table:
|Club|Coach|Replaced Coach|Home stadium|
|---|---|---|---|
|FC Dynamo Kyiv|Yozhef Sabo|Mykhailo Fomenko|Republican Stadium|
|FC Shakhtar Donetsk|Valery Yaremchenko||Shakhtar Stadium|
|FC Chornomorets Odessa|Viktor Prokopenko||Black Sea Shipping Stadium|
|FC Dnipro Dnipropetrovsk|Mykola Pavlov||Meteor Stadium|
|FC Karpaty Lviv|Myron Markevych||Ukraina Stadium|
|FC Kryvbas Kryvyi Rih|Volodymyr Brukhtiy|Ihor Nadein|Metalurh Stadium|
|FC Nyva Ternopil|Leonid Buriak||City Stadium|
|SC Tavriya Simferopol|Anatoliy Zayaev||Lokomotyv Stadium|
|FC Temp Shepetivka|Leonid Tkachenko||Temp Stadium|
|FC Nyva Vinnytsia|Yukhym Shkolnykov||Central City Stadium|
|FC Volyn Lutsk|Roman Pokora||Avanhard Stadium|
|FC Veres Rivne|Vyacheslav Kobyletskyi|Mykhailo Dunets|Avanhard Stadium|
|FC Torpedo Zaporizhia|Ihor Nadein|Viktor Matviyenko|AvtoZAZ Stadium|
|FC Zorya-MALS|Volodymyr Kobzarev|Anatoliy Shakun|Avanhard Stadium|
|FC Kremin Kremenchuk|Evhen Rudakov|Boris Streltsov 8 games Tiberiy Korponay 9 games|Dnipro Stadium|
|FC Metalurh Zaporizhia|Anatoliy Kuksov|Janis Skredelis Hryhoriy Vul|Metalurh Stadium|
|FC Bukovyna Chernivtsi|Oleksandr Pavlenko||Bukovyna Stadium|
|FC Metalist Kharkiv|Viktor Kamarzayev|Oleksandr Dovbiy Yevhen Lemeshko|Metalist Stadium|

question: What is the name of the home statium of the football club FC Kryvbas Kryvyi Rih?"
 
Output: 
```r
library(dplyr)

data <- read.csv("data.csv", check.names = FALSE)

home_stadium <- data %>%
  filter(Club == "FC Kryvbas Kryvyi Rih") %>%
  select(`Home stadium`)

write.csv(home_stadium, "result.csv", row.names = FALSE)
```
 
Input:
Table:
|Title|Year|Director(s)|
|---|---|---|
|"Red Morning Light"|2003|Douglas Biro|
|"Molly's Chambers"|2003|Honey|
|"Wasted Time"|2003|Mark Pellington|
|"The Bucket"|2004|Patrick Daughters|
|"Four Kicks"|2005|Patrick Daughters|
|"King of the Rodeo"|2005|Patrick Daughters|
|"On Call"|2007|Adria Petty|
|"Charmer"|2007|Robert Hales|
|"Crawl"|2008|Nick Wickham|
|"Sex on Fire"|2008|Sophie Muller|
|"Use Somebody"|2008|Sophie Muller|
|"Notion"|2009|Phil Griffin|
|"Radioactive"|2010|Sophie Muller|
|"Pyro"|2010|Max Goldman, Paul Greenhouse, Casey McGrath|
|"Back Down South"|2011|Casey McGrath|
|"Supersoaker"|2013|W.I.Z.|
|"Beautiful War"|2013|Casey McGrath|
|"Temple"|2014|Casey McGrath|
|"Waste a Moment"|2016|Dimitri Basil|
|"Walls"|2016|Dimitri Basil|
|"Around the World"|2016||

question: What are the names of the directors that have directed movies since 2010, and how many movies have they produced?
 
Output: 
```r
library(dplyr)

data <- read.csv("data.csv", stringsAsFactors = FALSE, check.names = FALSE)

movies_since_2010 <- subset(data, Year >= 2010)

director_counts <- aggregate(Title ~ `Director(s)`, data = movies_since_2010, FUN = function(x) length(x))

# colnames(director_counts) <- c("Director", "MoviesProduced")

write.csv(director_counts, "result.csv", row.names = FALSE)
```

Now, for the input table and question below,  generate its corresponding R code for the natural language question. No explanation, return the code only with the following markdown codeblock format : ```r<R CODE>```.

Input:
Table:
|Ward|Highest point|Elevation (approx.)|
|---|---|---|
|West Carleton-March Ward|2.6 km SSE of Manion Corners|166m|
|Rideau-Goulbourn Ward|Jinkinson Road, 8 km N of Munster|153m|
|Kanata South Ward|Glen Cairn Reservoir|131m|
|College Ward|Khymer Ct, 1 km N of Fallowfield|129m|
|Stittsville Ward|Rockson Cres.|128m|
|Kanata North Ward|Huntsville Dr, Kanata Lakes|126m|
|Osgoode Ward|1 km SE of Bank Street & Rideau Road, South Gloucester|120m+|
|Cumberland Ward|Cumberland Ridge Dr, Quigley Hill|120m+|
|Barrhaven Ward|Cedarview Road, Cedar Hill Estates|120m+|
|Gloucester-Southgate Ward|Tom Roberts Ave, Macdonald-Cartier International Airport|119m|
|Knoxdale-Merivale Ward|Cedarview Road at Cedarhill Drive|115m+|
|River Ward|Carlington Hill, (Carlington Heights Reservoir, Carlington Park) Carlington|115m+|
|Bay Ward|Corkstown Road, Ottawa - Nepean Tent & Trailer Park|114m|
|Gloucester-South Nepean Ward|Osgoode Link Pathway (former CPR) & High Rd, 4 km SW of Leitrim|114m|
|Innes Ward|200m WSW of Forest Ridge Pumping Station|114m|
|Beacon Hill-Cyrville Ward|Quarry Park, Rothwell Heights|113m|
|Rideau-Rockcliffe Ward|Foxview Pleasant, Quarries|106m|
|Alta Vista Ward|Alta Vista (Alta Vista Drive & Randall Ave)|102m|
|Capital Ward|Bank Street & Alta Vista Drive|96m|
|Orleans Ward|Clearcrest Cres, Fallingbrook|93m|
|Kitchissippi Ward|Maitland Avenue Bridge (over the Queensway)|89m|
|Somerset Ward|Parliament Hill|86m|
|Rideau-Vanier Ward|Richelieu Park, Vanier|75m+|


question: What is the highest elevation among all the wards in the table?

Output:

```r
library(dplyr)

data <- read.csv("data.csv", stringsAsFactors = FALSE, check.names = FALSE)

data$Elevation <- as.numeric(gsub("[a-zA-Z+ ]", "", data$`Elevation (approx.)`))

highest_elevation <- data %>%
  summarise(HighestElevation = max(Elevation, na.rm = TRUE))

write.csv(highest_elevation, "result.csv", row.names = FALSE)
```
\end{lstlisting}
\begin{lstlisting}[title=NL2R (Classification)]
Task description:
Natural language to R script, is a problem of generating R script  that can be executed on a input CSV table, for a given natural language question.

You will be given an input table called "table" in the markdown format below, a question in natural language for this table, and a corresponding R code for the natural language question. You need to: (1) determine if the R code is grammarly correct, (2) determine if the R code can truthfully represent the natural language question being asked.  Return your answer in the following JSON format: {"grammarly_correct": <YES or NO>, "R-correct": <YES or NO>} 

The code is generated with the following iunstructions:

When generating R script, you can assume that the input table is always stored in a file named "data.csv", so always start your R script with the boilerplate prefix:
"library(dplyr)

data <- read.csv("data.csv", check.names = FALSE)
"

After you generate R script code to produce results, please always save your final results into a file named "result.csv" with no index column, so your R script should always end with:
"write.csv(final_result, "result.csv", row.names = FALSE)"

Now, given the following input table, question, and R code, please determine if the R code is grammarly correct and if it can truthfully represent the natural language question being asked. Return your answer in the following JSON format: {"grammarly_correct": <YES or NO>, "R-correct": <YES or NO>}.

Table:
|Episode|Date|Challenging team|Mystery car|Winner|
|---|---|---|---|---|
|3|March 9|LDRSHIP Design (Temecula, California)|1986 Chevrolet El Camino|All Stars|
|9|April 27|Reyes Automotive Customs (South Los Angeles, California)|2000 Ford Crown Victoria|All Stars|
|5|March 23|SKJ Customs (St. George, Utah)|1979 Cadillac Coupe de Ville|All Stars|
|10|May 1|A recap of the past nine episodes where the All Stars count down their top 15 moments while the judges name the All-Stars' 1973 Datsun 240Z the Car Warriors Car of the Year.|A recap of the past nine episodes where the All Stars count down their top 15 moments while the judges name the All-Stars' 1973 Datsun 240Z the Car Warriors Car of the Year.|A recap of the past nine episodes where the All Stars count down their top 15 moments while the judges name the All-Stars' 1973 Datsun 240Z the Car Warriors Car of the Year.|
|2|March 2|Danny D. Customs (Chandler, Arizona)|1999 Ford F-150|All Stars|
|7|April 13|Gearhead Garage (Sacramento, California)|1976 Chevrolet Corvette|Gearhead Garage|
|8|April 20|War Machine (Edwards Air Force Base, Edwards, California)|Ford Taurus ARCA stock car|War Machine|
|4|March 16|True Bliss Customs (Burbank, California)|1973 Datsun 240Z|All Stars|
|1|February 23|Southern Fried D'Lite (Lubbock, Texas)|1966 Ford Mustang Coupe|Southern Fried D'Lite|
|6|March 30|Wheels (Palmdale, California)|1969 Volkswagen Beetle|All Stars|


Question:
Which challenging team won the episode on March 30th, 2019 after removing the prefix from the mystery car and counting the total number of episodes so far?

R code:
library(dplyr)

data <- read.csv("data.csv", check.names = FALSE)

data$Date <- as.Date(data$Date, format = "%B %d")

data$Mystery_car <- gsub(".*\\)", "", data$Mystery_car)

episode_count <- nrow(data)

episode_10_winner <- data %>%
  filter(Date == "2019-03-30") %>%
  select(`Challenging team`, Mystery_car)

write.csv(episode_10_winner, "result.csv", row.names = FALSE)

Output:

{"grammarly_correct": "YES", "R-correct": "YES"}
\end{lstlisting}
\begin{lstlisting}[title=NL-to-Scala (Generative)]
# Task description:
Natural language to Spark Scala, is a problem of generating Scala code using the Apache Spark library that can be executed on a input CSV table, for a given natural language question.

You will be given an input table called "table" in the markdown format, as well as a question. Your task is to generate Scala code using the Apache Spark library that can be executed on the input table to answer the question.

Please ensure that the generated Scala code only return relevant columns being specifically asked in the question.

When generating Scala code, you can assume that the input table is always stored in a file named "data.csv", so always start your Scala code with the boilerplate prefix:
"val df = spark.read.option("header", "true").csv("data.csv")"

After you generate Scala code to produce results, please always save your final results into a single CSV file named in the "result" directory, so your Scala code should always end with:
"finalResult.coalesce(1).write.option("header", "true").csv("result")"

No explanation, return the code only with the following markdown codeblock format : ```scala<SPARK SCALA CODE>```.

For example:
 
Input:
Table:
|City|Population (2015)|Area|Density|City class|Income class|Province|
|---|---|---|---|---|---|---|
|Bacolod|561,875|62.81|9,100|Highly urbanized|1st|Negros Occidental|
|Bago|170,981|154.9|1,100|Component|2nd|Negros Occidental|
|Bais|76,291|123.41|620|Component|3rd|Negros Oriental|
|Bayawan|117,900|269.92|440|Component|2nd|Negros Oriental|
|Cadiz|154,723|202.54|750|Component|2nd|Negros Occidental|
|Canlaon|54,509|66.0|830|Component|4th|Negros Oriental|
|Dumaguete|131,377|12.98|10,000|Component|2nd|Negros Oriental|
|Escalante|94,070|74.43|1,300|Component|4th|Negros Occidental|
|Guihulngan|95,969|150.02|650|Component|5th|Negros Oriental|
|Himamaylan|106,880|141.71|750|Component|3rd|Negros Occidental|
|Kabankalan|181,977|269.25|670|Component|1st|Negros Occidental|
|La Carlota|64,469|53.01|1,200|Component|4th|Negros Occidental|
|Sagay|146,264|127.54|1,100|Component|3rd|Negros Occidental|
|San Carlos|132,536|174.33|750|Component|2nd|Negros Occidental|
|Silay|126,930|82.93|1,500|Component|3rd|Negros Occidental|
|Sipalay|70,070|146.63|470|Component|4th|Negros Occidental|
|Talisay|102,214|77.68|1,300|Component|4th|Negros Occidental|
|Tanjay|80,532|106.58|750|Component|4th|Negros Oriental|
|Victorias|87,933|51.71|1,700|Component|4th|Negros Occidental|

question: What is the top-3 most populous city in 2015 in the table?

Output: 
```scala
val df = spark.read.option("header", "true").csv("data.csv")
val processedDf = df.withColumn("Population (2015)", regexp_replace($"Population (2015)", ",", "").cast("int"))
val top3PopulousCities = processedDf.orderBy($"Population (2015)".desc).limit(3).select("City")
top3PopulousCities.coalesce(1).write.option("header", "true").csv("result")
```

Input:
Table:
|Contestant ID|Finalist|Gender|
|---|---|---|
|SSJ01|Monisha|Female|
|SSJ02|Alka Ajith|Female|
|SSJ03|Balasarangan|Male|
|SSJ04|Sahana|Female|
|SSJ05|Manisha|Female|
|SSJ06|Prasanna Sundar|Male|
|SSJ07|Sowmya|Female|
|SSJ08|Oviya|Female|
|SSJ09|Vishnucharan|Male|
|SSJ10|Sivaranjani|Female|
|SSJ11|Nithyashree|Female|
|SSJ12|Harish|Male|
|SSJ13|Shri Prasanna|Male|
|SSJ14|Sanjana|Female|
|SSJ15|Shravan|Male|
|SSJ16|Srinisha|Female|
|SSJ17|Shrihari|Male|
|SSJ18|Sharath|Male|
|SSJ19|Roshan|Male|
|SSJ20|Srikanth|Male|
|SSJ21|Priyanka|Female|
|SSJ22|Soundarya|Female|
|SSJ23|Sathyanarayanan|Male|
|SSJ24|Srinidhi|Female|
|SSJ25|Ponmozhi|Female|

question: How many female contestants are there in total?
 
Output: 
```scala
val df = spark.read.option("header", "true").csv("data.csv")
val finalResult = df.filter(df("Gender") === "Female").select("Gender").groupBy("Gender").count()
finalResult.select("count").coalesce(1).write.option("header", "true").csv("result")
```
 
Input:
Table:
|Club|Coach|Replaced Coach|Home stadium|
|---|---|---|---|
|FC Dynamo Kyiv|Yozhef Sabo|Mykhailo Fomenko|Republican Stadium|
|FC Shakhtar Donetsk|Valery Yaremchenko||Shakhtar Stadium|
|FC Chornomorets Odessa|Viktor Prokopenko||Black Sea Shipping Stadium|
|FC Dnipro Dnipropetrovsk|Mykola Pavlov||Meteor Stadium|
|FC Karpaty Lviv|Myron Markevych||Ukraina Stadium|
|FC Kryvbas Kryvyi Rih|Volodymyr Brukhtiy|Ihor Nadein|Metalurh Stadium|
|FC Nyva Ternopil|Leonid Buriak||City Stadium|
|SC Tavriya Simferopol|Anatoliy Zayaev||Lokomotyv Stadium|
|FC Temp Shepetivka|Leonid Tkachenko||Temp Stadium|
|FC Nyva Vinnytsia|Yukhym Shkolnykov||Central City Stadium|
|FC Volyn Lutsk|Roman Pokora||Avanhard Stadium|
|FC Veres Rivne|Vyacheslav Kobyletskyi|Mykhailo Dunets|Avanhard Stadium|
|FC Torpedo Zaporizhia|Ihor Nadein|Viktor Matviyenko|AvtoZAZ Stadium|
|FC Zorya-MALS|Volodymyr Kobzarev|Anatoliy Shakun|Avanhard Stadium|
|FC Kremin Kremenchuk|Evhen Rudakov|Boris Streltsov 8 games Tiberiy Korponay 9 games|Dnipro Stadium|
|FC Metalurh Zaporizhia|Anatoliy Kuksov|Janis Skredelis Hryhoriy Vul|Metalurh Stadium|
|FC Bukovyna Chernivtsi|Oleksandr Pavlenko||Bukovyna Stadium|
|FC Metalist Kharkiv|Viktor Kamarzayev|Oleksandr Dovbiy Yevhen Lemeshko|Metalist Stadium|

question: What is the name of the home statium of the football club FC Kryvbas Kryvyi Rih?"

Output:
```scala
val df = spark.read.option("header", "true").csv("data.csv")
val finalResult = df.select("Home stadium").where(df("Club") === "FC Kryvbas Kryvyi Rih")
finalResult.coalesce(1).write.option("header", "true").csv("result")
```
 
Input:
Table:
|Title|Year|Director(s)|
|---|---|---|
|"Red Morning Light"|2003|Douglas Biro|
|"Molly's Chambers"|2003|Honey|
|"Wasted Time"|2003|Mark Pellington|
|"The Bucket"|2004|Patrick Daughters|
|"Four Kicks"|2005|Patrick Daughters|
|"King of the Rodeo"|2005|Patrick Daughters|
|"On Call"|2007|Adria Petty|
|"Charmer"|2007|Robert Hales|
|"Crawl"|2008|Nick Wickham|
|"Sex on Fire"|2008|Sophie Muller|
|"Use Somebody"|2008|Sophie Muller|
|"Notion"|2009|Phil Griffin|
|"Radioactive"|2010|Sophie Muller|
|"Pyro"|2010|Max Goldman, Paul Greenhouse, Casey McGrath|
|"Back Down South"|2011|Casey McGrath|
|"Supersoaker"|2013|W.I.Z.|
|"Beautiful War"|2013|Casey McGrath|
|"Temple"|2014|Casey McGrath|
|"Waste a Moment"|2016|Dimitri Basil|
|"Walls"|2016|Dimitri Basil|
|"Around the World"|2016||

question: What are the names of the directors that have directed movies since 2010, and how many movies have they produced?
 
Output:
```scala
val df = spark.read.option("header", "true").csv("data.csv")
val processedDf = df.withColumn("Year", df("Year").cast("int"))
val final_rst = processedDf.filter(processedDf("Year") >= 2010).groupBy("Director(s)").agg(count("Title").alias("Movies Produced"))
final_rst.coalesce(1).write.option("header", "true").csv("result")
```

Now, for the input table and question below, generate its generate its corresponding Scala code using the Apache Spark library. No explanation, return the code only with the following markdown codeblock format : ```scala<SPARK SCALA CODE>```.

Input:
Table:
|Title of Book|Author|Year|
|---|---|---|
|An Overview of 100 years of Norway as an Independent State (1905-2005)|Tasneem Sultana|2006|
|The Vicissitudes of the Palestinian Quest for Statehood and the European Union|Naveed Ahmad Tahir|2005|
|Indo-Russian Relations Since the Collapse of the Soviet Union|Naveed Ahmad Tahir|1999|
|The Expansion of the European Union: Problems & Prospects|Naveed Ahmad Tahir|1995|
|Muslim Rule in Spain|Dr. Affan Seljuq|1991|
|European Press; Tradition and Transition|M. Shamsuddin|1990|
|Sweden in Con-temporary World Politics|Naveed Ahmad Tahir|1990|
|Austria in World Affairs|Naveed Ahmad Tahir|1989|
|Turkey and the Super-powers|Rubab Hasan|1989|
|Pakistan-Italian Relations|Moonis Ahmar|1989|
|Finland: A Study in Neutrality|Naveed Ahmad Tahir|1987|


question: Who is the author of the book 'Muslim Rule in Spain'?

Output:

```scala
val df = spark.read.option("header", "true").csv("data.csv")
val finalResult = df.select("Author").where(df("Title of Book") === "Muslim Rule in Spain")
finalResult.coalesce(1).write.option("header", "true").csv("result")
```
\end{lstlisting}
\begin{lstlisting}[title=NL-to-Scala (Classification)]
Task description:
Natural language to Spark Scala, is a problem of generating Scala code using the Apache Spark library that can be executed on a input CSV table, for a given natural language question.

You will be given an input table called "table" in the markdown format below, a question in natural language for this table, and a corresponding Spark Scala code for the natural language question. You need to: (1) determine if the Spark Scala code is grammarly correct, (2) determine if the Spark Scala code can truthfully represent the natural language question being asked.  Return your answer in the following JSON format: {"grammarly_correct": <YES or NO>, "scala-correct": <YES or NO>} 

The code is generated with the following iunstructions:

When generating Scala code, you can assume that the input table is always stored in a file named "data.csv", so always start your Scala code with the boilerplate prefix:
"val df = spark.read.option("header", "true").csv("data.csv")"

After you generate Scala code to produce results, please always save your final results into a single CSV file named in the "result" directory, so your Scala code should always end with:
"finalResult.coalesce(1).write.option("header", "true").csv("result")"

Now, given the following input table, question, and Spark Scala code, please determine if the Scala code is grammarly correct and if it can truthfully represent the natural language question being asked. Return your answer in the following JSON format: {"grammarly_correct": <YES or NO>, "scala-correct": <YES or NO>}.

Table:
|Name|Weight(Start)|Weight(Ranch)|Weight(Final)|Total(lb)|Total(Percent)|
|---|---|---|---|---|---|
|Red Team Total|1051|964|753|298|28.35%|
|Sue|170|150|129|41|24.12%|
|Al|239|220|200|39|16.32%|
|Amy|238|223|179|59|24.79%|
|Anita|236|214|184|52|22.03%|
|Steve|368|344|258|110|29.89%|
|Tony|275|247|187|88|32%|
|Amelia|210|194|168|42|20%|
|Blue Team Total|907|833|722|185|20.4%|
|Ashley|222|205|170|52|23.42%|


Question:
What is the total weight of the individuals whose weight at the start is greater than 200, weight at the ranch is less than 250, and weight at the final is less than 190, and what is the average total percentage for this group?

Spark Scala Code:
val df = spark.read.option("header", "true").csv("data.csv")
val processedDf = df.withColumn("Weight(Start)", df("Weight(Start)").cast("int"))
    .withColumn("Weight(Ranch)", df("Weight(Ranch)").cast("int"))
    .withColumn("Weight(Final)", df("Weight(Final)").cast("int"))
    .withColumn("Total(Percent)", regexp_replace(df("Total(Percent)"), "%", "").cast("float"))

val totalWeight = processedDf.filter(processedDf("Weight(Start)") > 200 && processedDf("Weight(Ranch)") < 250 && processedDf("Weight(Final)") < 190).select(sum("Weight(Start)").alias("Total Weight")).collect()(0)(0)
val averagePercent = processedDf.filter(processedDf("Weight(Start)") > 200 && processedDf("Weight(Ranch)") < 250 && processedDf("Weight(Final)") < 190).select(avg("Total(Percent)").alias("Average Total Percent")).collect()(0)(0)

val finalResult = sc.parallelize(Seq(("Total Weight", totalWeight), ("Average Total Percent", averagePercent))).toDF("Metric", "Value")
finalResult.coalesce(1).write.option("header", "true").csv("result")

Output:

{"grammarly_correct": "YES", "scala-correct": "YES"}
\end{lstlisting}

% Row-2-Row

\begin{lstlisting}[title=Transformed Column Generation]
Instruction:
You are given a table below with a single column, please produce a derived column that can be generated from the column using a transformation program. In this specific case, your transformation should:

Perform a Numeric transformation that:
Convert into different unit (length, weight, time, etc.)

If the suggested transformation above is not feasible on the table column, please return your transformation plan in a JSON object as {"plan": "NULL"} and then return two empty code blocks for Pandas and R.

If the suggested transformation is feasible on the table column, please (1) first explain your plan in terms of how to transform the values in the column, in a JSON object as {"plan": "<PLAN>"}. Please then (2) return its Python implementation and R implementation, using two *code blocks*. Remember, the new derived column should always be named as "derived_column" in your python Pandas code and R code.

Please produce your Python Pandas code that can run on this table, to generate the derived column. To read this table, you can assume that the input table is always stored in a file named "data.csv", so always start your Python code with the boilerplate prefix:
"import pandas as pd
df = pd.read_csv('data.csv')
"

You should always generate your derived column in a column named "derived_column" in the final dataframe. Please always save your final dataframe into a file named "result.csv".
Return the final result as a markdown code block, as ```python <PYTHON-CODE>```

Similarly, please also produce your R code that can run on this table, to generate the derived column. To read this table, you can assume that the input table is always stored in a file named "data.csv", so always start your R code with the boilerplate prefix:
"library(dplyr)

data <- read.csv("data.csv", check.names = FALSE)
"

You should always generate your derived column in a column named "derived_column" in the final dataframe. Please always save your final dataframe into a file named "result.csv".
Return the final result as a markdown code block, as ```r <R-CODE>```

**Input table:**
|MergedColumn|
|---|
|18,950,15,522|
|25,367,20,645|
|19,538,23,868|
|8,978,15,657|
|5,304,22,110|
|36,773,26,000|
|11,513,16,210|
|21,306,27,550|
|24,864,19,107|
|12,663,28,406|


If the suggested transformation above is not feasible on the table column, please return your transformation plan in a JSON object as {"plan": "NULL"} and then return two empty code blocks for Pandas and R.

If the suggested transformation is feasible on the table column, please (1) first explain your plan in terms of how to transform the values in the column, in a JSON object as {"plan": "<PLAN>"}. Please then (2) return its Python implementation and R implementation, using two *code blocks*. Remember, the new derived column should always be named as "derived_column" in your python Pandas code and R code.

Output:

```json
{"plan": "Convert the values in the MergedColumn from the current format to a new format by replacing the comma with period (for decimal) and removing the other commas."}
```

```python
import pandas as pd

# Read the input table
df = pd.read_csv('data.csv')

# Perform the transformation
df['derived_column'] = df['MergedColumn'].str.replace(',', '').str.replace('.', ',')

# Save the final dataframe into a file named "result.csv"
df.to_csv('result.csv', index=False)
```

```r
library(dplyr)

# Read the input table
data <- read.csv("data.csv", check.names = FALSE)

# Perform the transformation
data <- data %>% mutate(derived_column = gsub(",", "", MergedColumn), 
                        derived_column = gsub("\\.", ",", derived_column))

# Save the final dataframe into a file named "result.csv"
write.csv(data, file = "result.csv", row.names=FALSE)
```
\end{lstlisting}
\begin{lstlisting}[title=Data Transformation (Pandas) (Generative)]
# Task description: The task of row-to-row transformation using examples, is to perform data transformation using table columns, to produce a new "DERIVED_COLUMN" in the table.
 
The task of "program-by-example" or "transform data by-example", is to use a few given example output values in the new "DERIVED_COLUMN", to infer the underlying transformation program that was used to derive the given output values in the "DERIVED_COLUMN". This is also known as program-synthesis or program-by-example in the research literature.
 
Given the **Input table** shown below, we want to perform data transformations on this Input Table, to produce a **Desired Output Table**, with the new "DERIVED_COLUMN" shown as the rightmost column of the **Desired Output Table**.
 
Please inspect the given output examples in the "DERIVED_COLUMN" of the **Desired Output Table**, to identify the transformation program that can be used to produce the "DERIVED_COLUMN" using columns in the **Input table**. Please generate Python Pandas code for the transformation that can execute on the **Input table**, to produce an output column that can exactly match the given output examples in the "DERIVED_COLUMN".
 
Please pay close attention to the format and the details of the given output examples in the "DERIVED_COLUMN", to make sure that your code will generate output values exactly matching the given examples, in terms of style, format, and casing.

To read this table, you can assume that the input table is always stored in a file named "data.csv", and always read everything as string (using "dtype=str"). So always start your Python code with the boilerplate prefix:
"import pandas as pd
df = pd.read_csv('data.csv', dtype=str)
"

For numerical computation, please use Python Decimal class to avoid floating point precision issues, so always include "from decimal import Decimal" at the beginning of your code.

Please do not hard code your transformation program (e.g., if row id == 1, return output-value-1, etc.). Please produce transformation programs that can *generalize* to new data rows from the same table, that are not shown below in the interest of space.
Your Python Pandas code should always generate your target column in a column named "DERIVED_COLUMN" in the final dataframe. Please always save your final dataframe with this  "DERIVED_COLUMN" into a file named "result.csv" with header and no index column.

**Input table:**
|MergedColumn|
|---|
|0.1666|
|0.15|
|0.131|


**Desired Output Table:**
|MergedColumn|DERIVED_COLUMN|
|---|---|
|0.1666|0.17|
|0.15|0.15|
|0.131|0.13|


No explanation, return the final result as a markdown code block, as ```python <PYTHON-CODE>```.
Output:

```python
import pandas as pd
from decimal import Decimal

df = pd.read_csv('data.csv', dtype=str)
df['DERIVED_COLUMN'] = df['MergedColumn'].apply(lambda x: str(Decimal(x).quantize(Decimal('0.01'))))
df.to_csv('result.csv', index=False)
```
\end{lstlisting}
\begin{lstlisting}[title=Data Transformation (R) (Generative)]
# Task description: The task of row-to-row transformation using examples, is to perform data transformation using table columns, to produce a new "DERIVED_COLUMN" in the table.
 
The task of "program-by-example" or "transform data by-example", is to use a few given example output values in the new "DERIVED_COLUMN", to infer the underlying transformation program that was used to derive the given output values in the "DERIVED_COLUMN". This is also known as program-synthesis or program-by-example in the research literature.
 
Given the **Input table** shown below, we want to perform data transformations on this Input Table, to produce a **Desired Output Table**, with the new "DERIVED_COLUMN" shown as the rightmost column of the **Desired Output Table**.
 
Please inspect the given output examples in the "DERIVED_COLUMN" of the **Desired Output Table**, to identify the transformation program that can be used to produce the "DERIVED_COLUMN" using columns in the **Input table**. Please generate R code for the transformation that can execute on the **Input table**, to produce an output column that can exactly match the given output examples in the "DERIVED_COLUMN".
 
Please pay close attention to the format and the details of the given output examples in the "DERIVED_COLUMN", to make sure that your code will generate output values exactly matching the given examples, in terms of style, format, and casing.
When you generate R code to read the input table, you can assume that the input table is always stored in a file named "data.csv", so always start your R code with the boilerplate prefix:
"library(dplyr)
 
data <- read.csv("data.csv", check.names = FALSE)
"
 
Please do not use case_when statements, or anything that is like CASE/IF/ELSE, to hard code your transformation program (e.g., if row id == 1, return output-value-1, etc.). Please produce transformation programs that can *generalize* to new data rows from the same table, that are not shown below in the interest of space.
 
Your R code should always generate your target column in a column named "DERIVED_COLUMN" in the final dataframe. Please always save your final dataframe with this "DERIVED_COLUMN" into a file named "result.csv" with header and no index column. No explanation, Return the final result as a markdown code block, as ```r<R CODE>```.

**Input table:**
|MergedColumn|
|---|
|Gary Payton|
|Alonzo Mourning|
|Shareef Abdur-Rahim|


**Desired Output Table:**
|MergedColumn|DERIVED_COLUMN|
|---|---|
|Gary Payton|GaryPayton|
|Alonzo Mourning|AlonzoMourning|
|Shareef Abdur-Rahim|ShareefAbdur-Rahim|


No explanation, return the final result as a markdown code block, as ```r<R CODE>```.
Output:

```r
library(dplyr)

data <- read.csv("data.csv", check.names = FALSE)

data <- data %>%
  mutate(DERIVED_COLUMN = gsub(" ", "", MergedColumn))

write.csv(data, "result.csv", row.names=FALSE)
```
\end{lstlisting}
\lstinputlisting[title=Data Transformation (SQL) (Generative)]{sections/Prompts/r2r_sql_generation.tex}

% Error Detection

\lstinputlisting[title=Error Detection (Generative)]{sections/Prompts/error_detection_generation.tex}
\lstinputlisting[title=Error Detection (Classification)]{sections/Prompts/error_detection_validation.tex}

% Schema matching
\lstinputlisting[title=Schema Matching (Generative)]{sections/Prompts/schema_matching_generation.tex}
\lstinputlisting[title=Schema Matching (Classification)]{sections/Prompts/schema_matching_validation.tex}

}
{
    % for now
    \clearpage
    %\appendix
    %\input{sections/apx-prompt}
}

%\break
\clearpage
\bibliographystyle{ACM-Reference-Format}
\bibliography{Table-GPT-Specialist}

\end{document}